\pdfoutput=1
\documentclass[11pt]{article}

\usepackage{emnlp2023}

\usepackage{times}
\usepackage{latexsym}
\usepackage{graphicx}
\usepackage{comment}
\usepackage{tabularx}
\usepackage{enumitem}
\usepackage{makecell}
\usepackage{colortbl}
\usepackage{booktabs}
\usepackage{multirow}
\usepackage{subcaption}

\usepackage[T1]{fontenc}
\usepackage[utf8]{inputenc}

\usepackage{microtype}

\title{Evaluating the Rationale Understanding of Critical Reasoning \\ in Logical Reading Comprehension}
\author{Akira Kawabata \\
  The Asahi Shimbun Company \\
  \texttt{kawabata-a@asahi.com} \\\And
  Saku Sugawara \\
  National Institute of Informatics \\
  \texttt{saku@nii.ac.jp} \\\
}

\begin{document}
\maketitle
\begin{abstract}
To precisely evaluate a language model's capability for logical reading comprehension, we present a dataset for testing the understanding of the rationale behind critical reasoning. % why answer choices in a main question are correct or incorrect.
For questions taken from an existing multiple-choice logical reading comprehension dataset, we crowdsource rationale texts that explain why we should select or eliminate answer options, resulting in 3,003 multiple-choice subquestions that are associated with 943 main questions.
Experiments on our dataset show that recent large language models (e.g., InstructGPT) struggle to answer the subquestions even if they are able to answer the main questions correctly.
We find that the models perform particularly poorly in answering subquestions written for the incorrect options of the main questions, implying that the models have a limited capability for explaining why incorrect alternatives should be eliminated.
These results suggest that our dataset encourages further investigation into the critical reasoning ability of language models while focusing on the elimination process of relevant alternatives.
\end{abstract}

\section{Introduction}

Critical reasoning, a type of logical reasoning not tied to formal logic, is a core ability of humans that is required for thoughtful reading of text.
It involves not only understanding what a passage explicitly says but also comprehending its underlying assumptions, argument structure, and supported conclusions.
Developing % natural language understanding 
systems capable of critical reasoning as reliably as humans is one of the ultimate goals of natural language processing.
Recent studies have proposed datasets that evaluate logical reasoning including critical reasoning ability \cite{yu-etal-2020-reclor, Liu-etal-2020-logiqa} in reading comprehension.
Owing to the recent development of large language models \cite[LLMs;][]{Brown-etal-2020-gpt3, he2023debertav}, the performance of the state-of-the-art models is nearing that of humans \cite{jiao-etal-2022-merit,wang-etal-2022-logic}.  %\footnote{\url{https://eval.ai/web/challenges/challenge-page/503/leaderboard/1347}}

%Despite promising strides in Large Language Models (LLMs)—achieving human-level accuracy on real-world test datasets—a critical question persists: Do these models genuinely understand logical reasoning, or are they merely proficient at predicting correct answers?

\begin{figure}
    \centering
    \includegraphics[width=\linewidth]{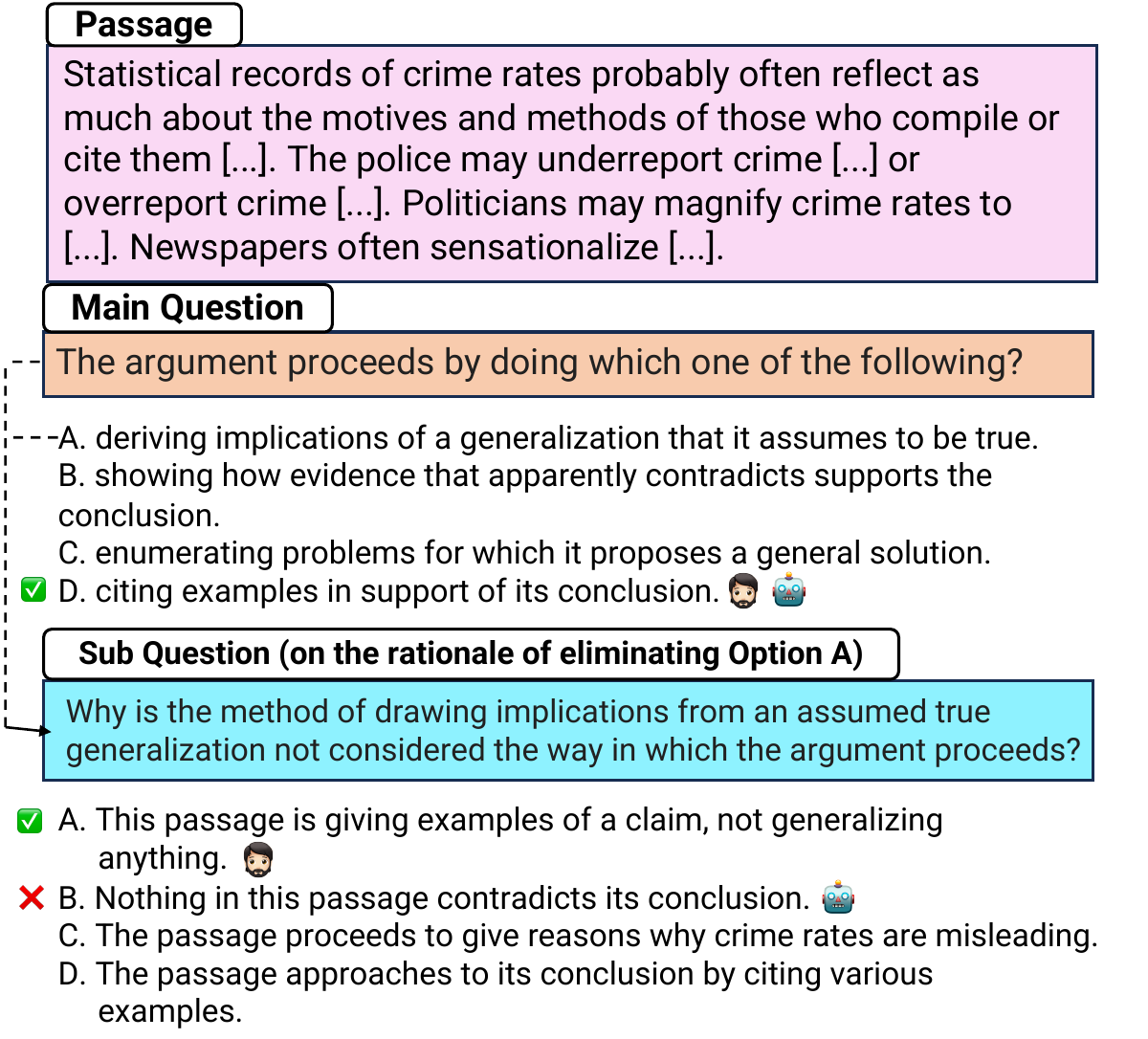}
    \caption{Example of ReClor \cite{yu-etal-2020-reclor} and its subquestion we create to test the understanding of implicit rationale.
    We find that even if the model can answer the original question correctly, it cannot answer subquestions that should be answerable.
    }
    \label{fig:intro-example}
\end{figure}

However, current multiple-choice questions in existing logical reading comprehension datasets may not sufficiently test the ability of critical reasoning.
The example illustrated in Figure~\ref{fig:intro-example} shows that even if a model can answer a question taken from the ReClor dataset \cite{yu-etal-2020-reclor} that has questions for graduate admission examinations, it cannot answer an auxiliary question that queries the implicit rationale for eliminating a relevant alternative.
This behavior might be due to the model's limited generalizability that is exposed by input perturbation \cite{si-etal-2021-benchmarking, lin-etal-2021-using, pmlr-v202-shi23a} or characterized as shortcut reasoning \cite{niven-kao-2019-probing,Geirhos2020-zj}.
Because a single question cannot fully ask the rationale of why we select an option as the correct answer and eliminate the others as the incorrect ones, current datasets may not be sufficient to comprehensively evaluate the process of critical reasoning. % that involves implicit rationales.

Recent studies propose methods for probing the reasoning process using auxiliary generation tasks such as in the form of simple commonsense facts \cite{aggarwal-etal-2021-explanations}, logical graphs \cite{huang-etal-2022-metalogic}, and arithmetic equations \cite{ribeiro2023street}.
However, this line of approach may not be suitable to capture the implicit rationale of critical reasoning.
In particular, it cannot explicitly consider the selection and elimination process of relevant alternatives in logical reasoning.
In addition, the format of such auxiliary tasks is usually not the same as that of the main task, which may fail to evaluate the target abilities consistently.
% However, this line of approach may have the following limitations: (i) they do not necessarily capture the implicit rationale that is necessary for critical reasoning, (ii) the selection and elimination process of relevant alternatives in logical reasoning is not explicitly considered, and (iii) the format of auxiliary tasks is not the same as that of the main task, which may be misleading to evaluate the target abilities consistently.
% However, those attempts may not be suitable for dealing with the critical reasoning necessary that involves implicit rationale behind the selection and elimination process of relevant alternatives.
% In addition, the formats of auxiliary tasks that are inconsistent with those of the main tasks, in particular using the generation task, can be insufficient to concisely evaluate the target abilities.

\begin{figure*}[h]
    \centering
    \includegraphics[width=0.9\linewidth]{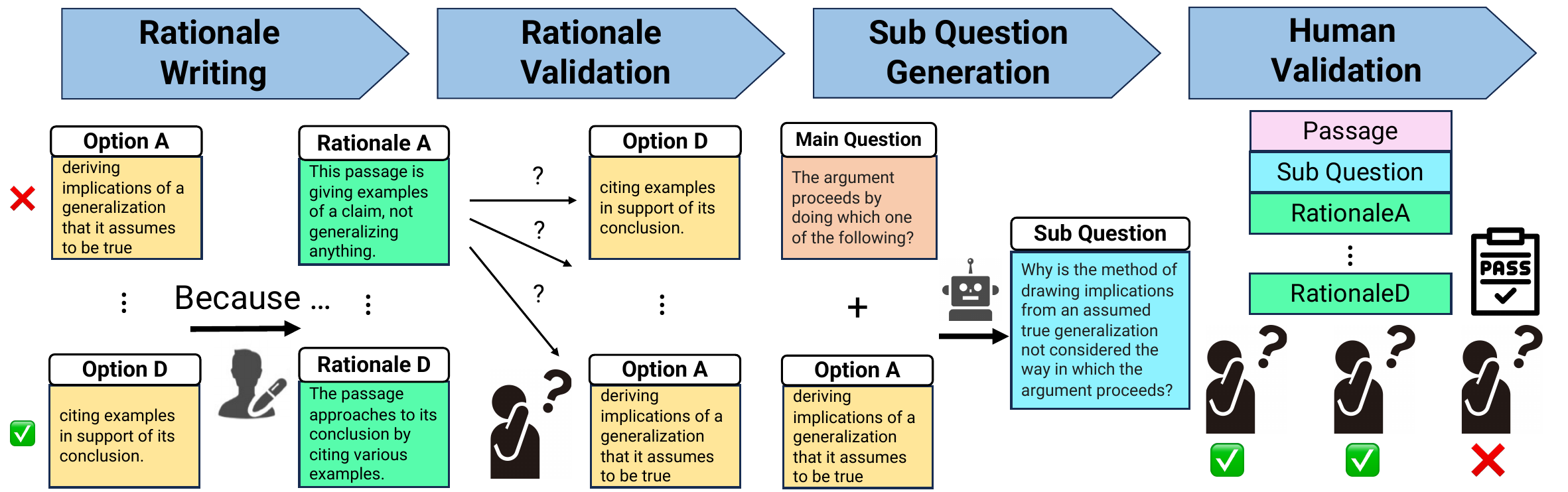}
    \caption{
        Our dataset construction process.
        We first ask crowdworkers to write the rationale for each answer option.
        After validating the collected rationale by aligning them to the source options, we use a large language model to generate subquestion texts.
        We finally ensure the human answerability of the generated subquestions.
    }
    \label{fig:rule-creation}
\end{figure*}

As a first step to address these limitations, we construct a benchmark that comprehensively evaluates language models' ability of critical reasoning in logical reading comprehension.
% Our dataset, RULE (Rationale Understanding for Logical reasoning Evaluation), consists of main questions taken from ReClor and auxiliary subquestions that we collect in this study.
Our dataset, rationale understanding for logical reasoning evaluation (RULE), consists of main questions taken from ReClor and auxiliary subquestions that we newly create for this study.
The process of constructing our dataset is illustrated in Figure~\ref{fig:rule-creation}. 
% As a main question has four answer options, each main option has at most four subquestions.
Our core idea is that for each answer option in a main question, we crowdsource a free-form human-written rationale that explains why that option should be selected or eliminated, and use those rationales to create a set of subquestions that are associated with the main question.  % evaluate the understanding of the rationale behind answering the main questions. % in the form of multiple-choice questions.
% We use crowdsourcing to collect rationales through carefully-designed annotations, and then use an LLM to generate a subquestion text given a pair of the main question and the target option.
% We collect free-form explanations using crowdsourcing as implicit rationales used in critical reasoning.
After manual filtering to ensure human answerability, in addition to 943 main questions, we obtain 3,003 subquestions for the test-only purpose.
The common multiple-choice format of the main questions and subquestions enables us to evaluate the models' capability of critical reasoning concisely and consistently.
% For concise and consistent evaluation, we form our subquestions as a discriminative task in the multiple-choice format as well. % mirroring the main questions' format.
% This common format allows us to easily evaluate the degree to which the models tested on the main questions demonstrate the ability to understand the necessary rationale through answering the subquestions.

% The process of constructing our dataset involves several steps, illustrated in Figure~\ref{fig:rule-creation}. 
% We first sample 1,200 examples in the training set of ReClor as our main questions.
% We then crowdsource a rationale text for each option in the main questions that asks the reason why the option is correct or incorrect.
% To collect such rationales manually, we use crowdsourcing with carefully designed qualification tests.
% We use crowdsourcing to collect rationales for each option in the main questions, and then use a large language model (LLM) to generate subquestion texts given a pair of the main question and the target option.
% After manual filtering to ensure human answerability, on top of 953 main questions, we obtain 3,003 subquestions for the test-only purpose.
% Using a large language model (LLM) to generate an auxiliary question text given a pair of the main question and the target option, we create a pool of multiple-choice subquestions.
% After manual filtering to ensure the human answerability, on top of 953 main questions, we obtain 3,003 subquestions for the test-only purpose.

In our experiments using strong baseline models including LLMs, e.g., Flan-UL2, \cite{tay2023ul}, LLaMA 2 \cite{Touvron2023-zh}, and InstructGPT \cite{Ouyang-etal-2022-instructgpt}, we observe that the models cannot answer the main questions and subquestions consistently, showing a larger than 30\% gap against humans in our strict consistency metric.
In particular, we find that the models struggle to answer eliminative subquestions, which are pertinent to the rationale of eliminating incorrect options, showing a large gap ($\approx$ 20\% accuracy) between humans and the best-performing LLM.
Conversely, the models tend to correctly answer selective subquestions, which are pertinent to the rationale of selecting the correct option.
% Some of the models surpass the human accuracy for those subquestions. % , some of which surpass the human accuracy. % recent in-context learning models surpass fine-tuned bert-like models in the consistent reasoning ability. 
This clear contrast suggests that these models provide the correct answer without fully understanding why the other options are incorrect. 
Our analysis using a follow-up task and manual annotations supports this observation.
We also compare our human-written rationales with model-generated ones using an LLM, finding that our rationales are likely to be more detailed and supportive than the model-generated ones.

% Our subsequent analysis also reveals that 
% \begin{enumerate}
%     \item An eliminative rationale may be difficult for the models to align with its target option, regardless of the complexity of the subquestion texts.
%     \item Rationale understanding involves not only direct, contextual reasoning but also indirect, external reasoning
%     \item The eliminative rationales are not helpful for the current models to answer the main questions, although they are necessary for humans.
%     \item The rationales we collected in this study are more meaningful than those generated by InstructGPT.
% \end{enumerate}

Our contributions are as follows:
(i) Based on an existing logical reading comprehension dataset, we create a dataset including over 3,000 auxiliary questions designed to test a model's consistent ability for critical reasoning.  % in logical reading comprehension.
(ii) We evaluate cutting-edge models, including LLMs, across finetuned, few-shot, and zero-shot settings, showing that even the best model falls short of human performance, particularly lagging in understanding eliminative rationales for incorrect answer options.  % various types of prompting and number of shots.
(iii) Our annotation analysis also highlights the model's deficiency in understanding eliminative rationales and shows that our human-written rationales are of higher quality than model-generated ones.\footnote{Our dataset, evaluation scripts with model hypterparameters, and annotation results are publicly available at \texttt{\href{https://github.com/nii-cl/rule}{github.com/nii-cl/rule}}}
%\footnote{We will make our dataset, evaluation scripts with model hyperparameters, and annotation results publicly available.} % upon acceptance.}
% (iii) We reveal that even the best model falls short of human performance, particularly lagging in answering rationales for incorrect answer options.

\section{Related Works}

\paragraph{Critical and Logical Reasoning}

Critical reasoning is one of the core abilities of logical reasoning that humans perform, along with analytical reasoning \cite{zhong-etal-2022-analytical} and abductive reasoning \cite{Bhagavatula2020Abductive}.  %  ability humans have, such as 
This reasoning is related to understanding the structure of practical arguments that is generally composed of ground (premise), warrant (rationale), and claim (conclusion).
As formulated by \citet{toulmin2003uses}, given facts or data as the ground, we provide the warrant that acts as a bridge between the ground and the claim we are making.
% Natural language understanding 
Recent research includes developing ways to model this behavior in tasks such as argument mining and question answering (QA) (e.g., ReClor).
For example, \citet{habernal-etal-2018-argument} propose a task of identifying implicit rationale (i.e., warrant) in arguments.
% Various aspects of Logical reasoning, such as analytical reasoning, and abductive reasoning, have been utilized to evaluate the capabilities of language models.
% \citet{zhong-etal-2022-analytical} study analytical reasoning, a process where a model analyzes a scenario and deduces conclusions.
% \citet{Bhagavatula2020Abductive} explore abductive reasoning, the process of finding the most plausible explanation for a situation.
% \citet{singh-etal-2022-irac} focus on how to identify implicit knowledge in arguments to better understand their logical connections.
% Similar to ReClor, LogiQA \cite{Liu-etal-2020-logiqa} and AR-LSAT \cite{zhong-etal-2022-analytical} datasets also cover complex logical reasoning in reading comprehension.
However, \citet{niven-kao-2019-probing} find that successful systems on the argument reasoning task exploit superficial input features.  %, i.e., shortcut learning \cite{Geirhos2020-zj}.
Similarly, QA systems have been shown to exhibit shallow understanding % that indicates unintended behaviors 
by input perturbation \cite{si-etal-2021-benchmarking, lin-etal-2021-using, pmlr-v202-shi23a}. % or characterized as shortcut reasoning \cite{niven-kao-2019-probing,Geirhos2020-zj}
For example, \citet{lin-etal-2021-using} demonstrate that QA performance significantly decreases when incorrect options are replaced with irrelevant texts in an adversarial manner.
% However, these datasets may be insufficient to evaluate the generalizable capability of critical reasoning as suffering from superficial input features that are exploitable by shortcut learning of language understanding models \cite{niven-kao-2019-probing,Geirhos2020-zj}.
% A similar issue is observed in multi-hop reasoning such as 
This means that successful models on those datasets do not necessarily exhibit generalizable capabilities in other datasets.
These findings necessitate the explainability of the (informal) logical reasoning process for better evaluation of intended reasoning abilities (e.g., the critical reasoning in this study).
%Furthermore, Tang et al. (2021) reveal that existing models trained on multi-hop QA struggle to generalize to single hop sub-questions. 

\paragraph{Reasoning Explanation}
% Our work is different from studies focusing on generating explanations. 
Although some studies explain the rationale behind commonsense and logical reasoning using graphs \cite{saha-etal-2021-explagraphs, ribeiro2023street}, others explain it as a decomposition \cite{Khot2020-ml, dalvi-etal-2021-explaining, geva-etal-2021-aristotle}, a combination of supporting textual spans in the input \cite{yang-etal-2018-hotpotqa, inoue-etal-2020-r4c}, commonsense rules \cite{saha-etal-2022-hard}, or underlying facts \cite{aggarwal-etal-2021-explanations}.
The work most similar to ours is MetaLogic \cite{huang-etal-2022-metalogic}, which focuses on generating graphs explaining the logical relations between sentences in ReClor examples, aiming to model the valid reasoning process.
In contrast, we employ free-text rationales that explain the process of critical reasoning, enabling us to construct multiple-choice questions about the understanding of rationales.
We also aim to faithfully test the models' performance on the main questions as well as auxiliary subquestions in the multiple-choice discrimination task, instead of the generation of the reasoning process in a different format from the original task. % (e.g., asking to generate the explanation for the multiple-choice commonsense reasoning task).
% The common format enables us to evaluate the intended ability  
% , and for this purpose it is reasonable to create questions in the same format as the main question.

\section{RULE Data Collection}

\subsection{Design Choices}

% In this study, 
We construct a new dataset, RULE (rationale understanding for logical reasoning evaluation), to evaluate the consistent rationale understanding in logical reading comprehension.
% This study aims to evaluate the consistent understanding of logical reasoning through the construction of a dataset, RULE.
The dataset comprises % two types of questions: 
main questions and their auxiliary questions (subquestions).
% A total of 1,200 main questions were randomly curated from ReClor. Correspondingly,
The subquestions are designed to test the understanding of the rationale necessary for answering the main questions correctly. % behind each option presented in the main questions.
%
% In constructing our dataset, we make three decisions in its design choices.
% \begin{itemize}
%     \item Focus on the task of \textit{logical reading comprehension} to assess the capability of complex reasoning skills.
%     \item Use the \textit{multiple-choice question-answering format} to decompose the rationale into smaller pieces and explicate the rationale that is necessary for eliminating relevant alternatives.
%     \item Define the auxiliary questions in the \textit{same task format} (i.e., multiple-choice) as the main questions to test models in a consistent way.
% \end{itemize}
%
In constructing our dataset, we make three decisions in its design choices.

\paragraph{Source Dataset} % or Target Task
% Logical reading comprehension is tested in the multiple-choice format in existing datasets.
Among existing datasets for testing logical reading comprehension, we use ReClor for the following reasons:
% In this study, we use ReClor as an example dataset for the following reasons:
(1) It covers various types of logical reasoning required in the multiple-choice format, % exams of GMAT and LSAT, % by selecting and eliminating relevant alternatives in the multiple-choice format,
(2) its context passages are of sufficient length to compose a meaningful rationale (e.g., the contexts in LogiQA \cite{Liu-etal-2020-logiqa} are shorter), and (3) it contains a sufficient number of examples to create an auxiliary benchmarking dataset.
We cannot find other candidate datasets, but our approach is applicable to similar ones.

\paragraph{Rationale Collection}
The task of writing implicit rationales from scratch for logical reasoning questions is not straightforward because the reasoning process can involve multiple steps with differing granularity.
Therefore, to facilitate rationale writing, % the collection of rationales in a reliable way,
we use answer options in the multiple-choice questions.
To answer a question with four options, the reasoning process should involve the rationale of both identifying the correct option and eliminating the three incorrect options.
By focusing on the correctness of each option, we can decompose the complex task of rationale writing into smaller intuitive tasks.
In addition, we collect human-written free-form rationales to expect benefits over model-generated rationales \cite{sun-etal-2022-investigating}, in particular for covering the implicit process of critical reasoning.

\paragraph{Task Format}
We also aim to design auxiliary questions so that we can easily evaluate models on both main questions and subquestions in the same task format.
To this end, we use four rationales collected for a main question as the four answer options of its subquestion.
A single main question has at most four subquestions that share the same set of answer options, which can be seen as question-wise contrastive evaluation \cite{gardner-etal-2020-evaluating,ashida-sugawara-2022-possible}.
% for consistent evaluation, we use the multiple-choice format both for our main questions and subquestions.

% The following section details the construction process of the subquestions, including the collection of rationales (Section~\ref{sec:rationale-collection}), the generation of subquestions (Section~\ref{sec:subquestion-generation}), and the manual validation of the subquestions (Section~\ref{sec:subquestion-validation}).

\subsection{Collecting Rationales}
\label{sec:rationale-collection}

We use crowdsourcing to collect rationales for creating our subquestions.
Appendix~\ref{app:instructions} shows our crowdsourcing instructions and examples.

\paragraph{Qualification}

We conduct a two-stage qualification test to recruit crowdworkers for our tasks. 
The first stage is a QA task to identify workers who carefully answer logical reading comprehension questions.
The task consists of ten questions taken from ReClor, and workers achieving $\geq 80\%$ accuracy advance to the next test.
In the second stage, workers are presented with a single ReClor question that is randomly sampled from a pool of ten questions. % the authors select. 
The task is to write four implicit rationales (one sentence each) behind each option's (in)correctness.
To guide them, we provide detailed instructions with eight writing examples.

Through preliminary pilot studies, we define two essential criteria for writing rationales: specificity and necessity. % consistency with the question's intent.  % consistency? essentiality?
Specificity requires rationales to be well informed and support the corresponding options exclusively. 
This requirement is crucial because non-specific rationales could support multiple options, rendering them unsuitable for options in subquestions. 
Necessity emphasizes the importance of ensuring that the rationale is essential for validating the option's correctness. 
Even if a detailed rationale is provided, it must be aligned with the main question's point to preserve its validity.

% The importance of these criteria was confirmed through observations from three prior pilot studies of rationale writing tasks. 
Following these criteria, the authors manually assess the rationales provided by the workers. 
%This rigorous qualification process resulted in 57 diverse participants, thereby enhancing the quality of the constructed questions \cite{hoge}. 
We identify 57 workers through this qualification process.
These workers are invited to both the rationale writing and subsequent validation tasks.
% These selected individuals participated in both the rationale writing task and the subsequent evaluation tasks.

\paragraph{Rationale Writing}

% After passing the qualification testa, participants were recruited for the rationale writing task. 
We take 1,200 questions from the training set of ReClor.
As with the second phase of the qualification task, we present workers with a context, question, and four options marked as either correct or incorrect, and then ask workers to write rationale sentences for each option.
Of these qualified individuals, 50 were actively engaged in this task. % They were each presented with a ReClor question and its four options, which were marked as either correct or incorrect. 
% Their task was to write a rationale explaining the correctness of each option, resulting in four rationales per main question. 
We collect 4,800 rationales in total and send them to the rationale validation step. % from 1,200 ReClor questions for this study.

\paragraph{Rationale Validation} % on specificity?

To validate the collected rationales, we first focus on their specificity, % because each rationale must exclusively support its corresponding answer option.
% This requirement 
which is critical for creating a set of reasonable subquestions about a given main question.
% In contrast,
Because assessing the necessity of rationales may not be straightforward, we analyze the reasoning types involved in understanding rationales in Section~\ref{sec:analysis}. % later, as described in Section~\ref{sec:analysis}.
% To ensure the quality of collected rationales, we assess their specificity by the following alignment validation.
% The reason why we focus only on specificity is that the assessment of necessity is not straightforward.
% After completing the rationale collection, we undertook quality assurance. 
% This involved assessing the specificity of each rationale to determine whether it was detailed enough to only support the corresponding option.

For the validation, we conduct an alignment test between a set of rationales and answer options.
In this test, workers are presented with one main question, its four options, and one rationale.
They are then asked to identify which one of the options is supported by the given rationale.
If a rationale is insufficiently detailed and could potentially support other options, it would be difficult for workers to correctly match the rationale to its corresponding option.
We ensure that the worker who validates a rationale is different from the one who wrote it.
% To do this, we conducted a matching test. 
% In this task, participants were presented with a ReClor question and a rationale corresponding to one of the options. 
% They were then asked to identify which of the options was supported by the rationale.
% We made sure the person who wrote the rationale is not the one evaluating it to maintain objectivity in the process.

% We theorized that if the rationale was insufficiently detailed and could potentially support other options, it would be difficult for participants to correctly match the rationale to its corresponding option. 
% Therefore, we concluded that rationales that passed this test had sufficient specificity.

This test enables us to refine our initial pool of 4,800 rationales down to 3,828, ensuring that each rationale is sufficiently specific to support its corresponding option.

\subsection{Subquestion Construction}
\label{sec:subquestion-generation}

\paragraph{Question Generation}

We then generate question texts to construct subquestions using a language model.
Given one main question and one of its options, the model is instructed to generate a subquestion that asks about the reason for the correctness of the option. % (i.e., why the option is correct or incorrect). 
For example, when we input the prompt ``What mistake does the argument make in its reasoning?'' and the incorrect answer option ``It confuses probability and certainty,'' the model generates the question ``What evidence is there that the argument does not make the mistake of confusing probability and certainty?''
We use different prompts for the correct and incorrect options to avoid the problem of the model omitting negatives (e.g., ``not'') when generating eliminative subquestions.  % for the incorrect option.
For the generation, we use InstructGPT (\texttt{text-davinci-003}), % \cite[\texttt{text-davinci-003};][]{Ouyang-etal-2022-instructgpt},
%which is one of the stable versions of strong large language models. 
which is one of the strong large language models.
%\footnote{\url{https://platform.openai.com/docs/models/gpt-3-5}}
Appendix~\ref{app:prompt-qg} shows an example of our prompt.

% \subsection{Rationale Question Construction}

\paragraph{Subquestion Construction}
Coupling the validated rationales with generated question texts, we construct at most four subquestions for a single main question.
Each subquestion corresponds to each of the four answer options in the main question.
The four answer options of the subquestions are identical to the four rationales written for the main question.
The correct answer option of a subquestion is the rationale written for the option that the subquestion is made from.
% The subquestions share the same context as the corresponding main question, and the four options of the subquestion are identical to the four rationales written for the main question.
% The correct answer option is the rationale written for the option on which the subquestion is.
% Formally, given a main question $q$ on its passage $p$ with four answer options $a_i$ ($i=1,...,4$), we have rationales $r_i$ for each answer option, and then construct at most four subquestions $q_k=M(q, a_k)$ on $p$ with four answer options $r_i$ in which the answer is $r_k$.

% Intuitively, 
A subquestion must have four validated rationales to compose the multiple-choice format.
However, when we look at a main question, % it is not realistic that 
all four rationales are not always valid, % (i.e., exclusively specific),
which could largely decrease the number of possible subquestions. % we can create.
To mitigate this issue, we create a subquestion even if three out of the four rationales are valid, by replacing the invalid rationale with the ``None of the above choices'' option.
Through this process, we obtain 3,824 subquestions. 
We discard a main question if it has no valid subquestions. % (i.e., it has two or fewer valid rationales).

% \subsection{Validation by Human Accuracy}
\subsection{Human Validation}
\label{sec:subquestion-validation}

As the final step of our data collection, we validate the answerability of the subquestions by humans. 
Despite the ensured specificity of rationales, the complexity of the subquestion texts could potentially make the subquestions unanswerable. 
To address this issue, we ask three workers to answer each subquestion to evaluate its human answerability. 
A subquestion is considered answerable if at least two workers answer it correctly, or if all workers select ``None of the above choices.''
In the latter scenario, we replace the correct answer in the question with ``None of the above choices.'' 
This process results in 3,003 answerable subquestions with 943 main questions.
We expect the number of questions in our dataset can demonstrate statistical power for meaningful model benchmarking and comparison \cite{card-etal-2020-little}.

We then ask different workers to answer the questions, collecting three additional labels for each question to measure human accuracy. % on our dataset.

\subsection{Dataset Statistics}

\begin{table}
    \centering % \small
    \begin{tabular}{lc} \toprule
        \# Main / Sub Questions & 943 / 3,003 \\
        \# SubQ / MainQ & 3.18 \\
        \# Selective / Eliminative (S/E) & 785 / 2,218 \\
        Avg. context length & 73.8 \\
        Avg. question length & 31.4 (15.5) \\
        Avg. option length & 23.5 (17.7) \\
        Avg. correct option length & 24.0 (18.6) \\
        \# Question vocabulary & 8,843 (1,085) \\
        \# Option vocabulary & 9,849 (9,652) \\
        % \# SubQ w/ ``None'' & 1,102 \\ % (36.3\%) \\
        % \# SubQ w/ ``None'' as the answer & 222 \\ % (7.3\%) \\
        \# SubQ w/ ``None'' (\# answer) & 1,102 (222) \\ 
        %  & MainQ & SubQ \\ \midrule
        % \# Questions & 943 & 3,003 \\
        % \# SubQ / \# MainQ & \multicolumn{2}{c}{3.18} \\
        % \# Correct/Incorrect & & 2,218/785 \\
        % Avg. context length & 73.8 & 73.8 \\
        % Avg. question length & 15.5 & 31.4 \\
        % Avg. correct option length & 18.6 & 24.0 \\
        % Avg. option length & xxx & yyy \\    
        \bottomrule
    \end{tabular}
    \caption{
    Dataset statistics of our RULE dataset.
    \textit{S/E} indicates the numbers of two types of subquestions written about the correct (selective) or incorrect (eliminative) options of their main questions, respectively.
    The question and option lengths of the main questions are separately reported in parentheses for comparison.
    \textit{``None''} denotes ``None of the above choices.'' % and at the bottom row we also report the number of subquestions in which this choice is the correct answer.
    }
    \label{tab:stats}
\end{table}

Table~\ref{tab:stats} shows the dataset statistics. % are shown in Table~\ref{tab:stats}.
% A single main question has 3.18 subquestions on average.
Compared to the main questions (ReClor), our subquestions have longer questions and answer options.
The subquestions that have ``None of the above choices'' as the correct answer comprise 7.4\% (222/3,003) of the dataset, which is comparable to a similar multiple-choice reading comprehension dataset \cite[6.7\% in CosmosQA;][]{huang-etal-2019-cosmos}.
We also report the crowdsourcing details in Appendix~\ref{app:crowdsourcing}.

\section{Baseline Performance on RULE}
\label{sec:experiments}

We measure the baseline performance of recent state-of-the-art models on our dataset.
Because the main purpose of our dataset is to perform an extensive evaluation of the models tested on ReClor, we use all of our main questions and subquestions as a test set.
Our hypothesis is that if the models can effectively generalize to understand the rationale behind the correct answer, they should exhibit a similar degree of performance on both the main questions and subquestions.

\paragraph{Evaluation Metrics}
In addition to the simple accuracy over the main questions (\textit{MainQ Accuracy}) and subquestions (\textit{SubQ Accuracy}), we calculate the accuracy across the subquestions written for the correct and incorrect original options (\textit{Selective} and \textit{Eliminative SubQ Accuracy}), respectively.
We also calculate the \textit{Consistency} score to see how often a model answers both the main question and all of its subquestions correctly and thereby shows the comprehensive capability of critical reasoning.
Because the SubQ accuracy is a micro average, we also report a macro average for reference (\textit{MainQ-wise SubQ Accuracy}).
To compute these scores for humans, we take a majority vote of the three labels for each main question and subquestion.

%In this experiment, we aim to investigate these two questions.
%(1) Can models reason with a comprehension of the rationale?
%(2) Do models comprehend the rationales for correct and incorrect answers equally well?

\subsection{Models and Settings}

% The baseline models are grouped into three categories on the basis of their usage of our provided training data: fully finetuned models that use all data, few-shot models that use only a small fraction, and zero-shot models that use only instruction.
The models we evaluate are either in the fully-finetuned setting on the training set of ReClor (excluding our main questions), few-shot of ReClor, and zero-shot that uses only the task instruction.

\paragraph{Fully-Finetuned Models}
We use DeBERTa-v3 \cite[large;][]{he2023debertav} and UnifiedQA-v2 \cite[base, large, and 3B;][]{khashabi-etal-2020-unifiedqa,khashabi2022unifiedqav2}.
Both models are reported to exhibit strong generalization performance on QA datasets.
% While DeBERTa models are finetuned for the multiple-choice task, UnifiedQA models are finetuned to generate the sentence of an answer option, by which we compute the similarity scores and choose the option that gives the highest similarity score.
% In our full finetuning category, we included three BERT-like models,  DeBERTa-v3 \cite{he2021debertav3}, alongside UnifiedQA-v2 3B (tabun). 
% UnifiedQA is a variant of T5 \cite{} that is trained on 20 QA datasets and demonstrates superior generalization to unknown question types.

\paragraph{Few- and Zero-Shot Models}
We include recent LLMs such as FLAN-T5 \cite[XXL;][]{chung2022scaling}, Flan-UL2 \cite[20B;][]{tay2023ul}, Vicuna \cite[7B and 13B;][]{vicuna2023}, LLaMA 2 \cite[7B to 70B;][]{Touvron2023-zh}, Mistral \cite[7B;][]{jiang2023mistral} and InstructGPT \cite[\texttt{text-davinci-003};][]{Ouyang-etal-2022-instructgpt}.
% For InstructGPT, we use \texttt{text-davinci-003} which is reported to be the best model in this model series.
In the few-shot setting, the input prompt has five ReClor exemplars.
Because some models only accept a limited length of input, we only report one-shot results of those models. % in Appendix~\ref{app:one-shot-result} instead.
For reference, we report few-shot results using RULE examples.
% In the zero-shot settings, we provide the models with only the task instruction.
The zero-shot prompt only has the task instruction.
We also include Chain-of-Thoughts \cite[CoT;][]{wei2022chain} and zero-shot CoT \cite{kojima2022large} of InstructGPT, providing the models with explanatory examples to potentially enhance their performance.
In CoT, the prompt includes ReClor exemplars each of which is followed by the rationale of the correct answer option that is collected in this study.
Appendix~\ref{app:prompt-cot} shows examples of our CoT prompt.

In the few- and zero-shot settings, we follow the test split approach used by \citet{ravichander-etal-2022-condaqa} and split our dataset into five disjoint sets to measure the variability of models' performance.
Appendix~\ref{app:test-split-setting} describes the details.

% \paragraph{Test Set Split for In-Context Learning}
% The in-context learning performance of LLMs may vary depending on the exemplars of the prompt, but it incurs a high computational cost % (or financial cost for proprietary models) 
% if we repeatedly evaluate the models on the entire dataset using various sets of different exemplars to take the average performance.
% Because of this cost limitation, we follow the test split approach used by \citet{ravichander-etal-2022-condaqa}, splitting our dataset into five disjoint sets and testing the models on each set with different exemplars to measure the performance variance across the disjoint sets.
% Note that we do not split the set of the main questions, because it has only 943 examples; hence, in the few-shot setting, we take the average across five runs on all main questions.
% In the few-shot setting using ReClor, we sample questions disjointly from its training set, whereas in using RULE, the exemplars are sampled from the corresponding disjoint set.

\subsection{Results}

\definecolor{light-gray}{gray}{0.94}
\begin{table*}[!h]
\centering \small
% \resizebox{\textwidth}{!}{%
\begin{tabular}{lcrrrrrr} \toprule
\textbf{Model} & \textbf{\# Param} & \makecell[c]{\textbf{MainQ}\\\textbf{Acc.}} & \makecell[c]{\textbf{SubQ}\\\textbf{Acc.}} & \makecell[c]{\textbf{Selective}\\\textbf{SubQ Acc.}} & \makecell[c]{\textbf{Eliminative}\\\textbf{SubQ Acc.}} & \textbf{Consist.} &  \makecell[r]{\textbf{MainQ-wise}\\\textbf{SubQ Acc.}} \\ \midrule
\multicolumn{8}{c}{\cellcolor{light-gray} \emph{Fully Finetuned on ReClor}} \\

\textsc{DeBERTa-v3-Large} & 304M & 66.0\hphantom{$_{\pm0.00}$} & \textbf{33.1}\hphantom{$_{\pm0.00}$} & \textbf{56.1}\hphantom{$_{\pm0.00}$} & \textbf{25.0}\hphantom{$_{\pm0.00}$} & \textbf{2.4} & \textbf{32.8}\\
\textsc{UnifiedQA-v2-Base} & 220M & 40.5\hphantom{$_{\pm0.00}$} & 25.8\hphantom{$_{\pm0.00}$} & 21.3\hphantom{$_{\pm0.00}$} & 27.4\hphantom{$_{\pm0.00}$} & 0.7 & 26.0\\
\textsc{UnifiedQA-v2-Large} & 770M & 57.7\hphantom{$_{\pm0.00}$} & 25.0\hphantom{$_{\pm0.00}$} & 19.9\hphantom{$_{\pm0.00}$} & 26.8\hphantom{$_{\pm0.00}$} & 1.4 & 24.7\\
\textsc{UnifiedQA-v2-3B} & 3B & \textbf{66.8}\hphantom{$_{\pm0.00}$} & 25.3\hphantom{$_{\pm0.00}$} & 21.8\hphantom{$_{\pm0.00}$} & 26.6\hphantom{$_{\pm0.00}$} & 1.4 & 25.2\\
\multicolumn{8}{c}{\cellcolor{light-gray} \emph{Five-Shot on ReClor}} \\
\textsc{Vicuna 13B} & 13B & 46.2$_{\pm0.7\hphantom{0}}$ & 50.0$_{\pm4.4\hphantom{0}}$ & 78.2$_{\pm3.0\hphantom{0}}$ & 40.1$_{\pm4.6\hphantom{0}}$ & 5.6 & 49.4\\
\textsc{Flan-UL2} & 20B & 58.5$_{\pm0.3\hphantom{0}}$ & \textbf{65.5}$_{\pm5.1\hphantom{0}}$ & 88.0$_{\pm4.0\hphantom{0}}$ & \textbf{57.6}$_{\pm5.4\hphantom{0}}$ & 16.9 & \textbf{64.3}\\
% \textsc{LLaMA 7B} & 7B & 25.8$_{\pm1.6\hphantom{0}}$ & 28.6$_{\pm7.1\hphantom{0}}$ & 39.8$_{\pm13.1}$ & 24.6$_{\pm6.3\hphantom{0}}$ & 0.8 & 28.2\\
% \textsc{LLaMA 13B} & 13B & 38.7$_{\pm2.7\hphantom{0}}$ & 36.3$_{\pm3.5\hphantom{0}}$ & 63.6$_{\pm4.0\hphantom{0}}$ & 26.6$_{\pm3.6\hphantom{0}}$ & 2.9 & 36.6\\
%\textsc{LLaMA 33B} & 33B & 58.5$_{\pm1.2\hphantom{0}}$ & 48.1$_{\pm3.5\hphantom{0}}$ & 77.3$_{\pm3.6\hphantom{0}}$ & 37.7$_{\pm3.6\hphantom{0}}$ & 6.2 & 48.0\\
%\textsc{LLaMA 65B} & 65B & 69.1$_{\pm0.9\hphantom{0}}$ & 55.3$_{\pm2.7\hphantom{0}}$ & 85.0$_{\pm1.4\hphantom{0}}$ & 44.8$_{\pm2.5\hphantom{0}}$ & 11.2 & 54.9\\
% \textsc{Vicuna 7B} & 7B & 33.4$_{\pm2.6\hphantom{0}}$ & 38.6$_{\pm3.3\hphantom{0}}$ & 61.4$_{\pm5.8\hphantom{0}}$ & 30.5$_{\pm3.4\hphantom{0}}$ & 2.8 & 38.3\\
\textsc{InstructGPT} & N/A & 71.8$_{\pm1.0\hphantom{0}}$ & 65.3$_{\pm1.8\hphantom{0}}$ & 88.4$_{\pm2.5\hphantom{0}}$ & 57.1$_{\pm1.5\hphantom{0}}$ & \textbf{18.2} & 64.0\\
\textsc{InstructGPT + CoT} & N/A & 67.8$_{\pm0.5\hphantom{0}}$ & 63.2$_{\pm2.1\hphantom{0}}$ & 88.5$_{\pm2.5\hphantom{0}}$ & 54.2$_{\pm2.8\hphantom{0}}$ & 17.2 & 61.8\\
\textsc{LLaMA2 13B} & 13B & 48.5$_{\pm2.5\hphantom{0}}$ & 44.6$_{\pm3.2\hphantom{0}}$ & 75.3$_{\pm3.4\hphantom{0}}$ & 33.8$_{\pm4.0\hphantom{0}}$ & 5.3 & 44.7\\
\textsc{LLaMA2 70B} & 70B & \textbf{80.3}$_{\pm0.4\hphantom{0}}$ & 60.0$_{\pm2.6\hphantom{0}}$ & \textbf{90.0}$_{\pm1.1\hphantom{0}}$ & 49.4$_{\pm2.9\hphantom{0}}$ & 17.7 & 59.3\\
\textsc{Mistral 7B} & 7B & 59.9$_{\pm0.9\hphantom{0}}$ & 55.3$_{\pm3.4\hphantom{0}}$ & 83.6$_{\pm3.4\hphantom{0}}$ & 45.4$_{\pm3.6\hphantom{0}}$ & 9.0 & 54.4\\
\multicolumn{8}{c}{\cellcolor{light-gray} \emph{Five-Shot on RULE (for reference)}} \\
\textsc{Vicuna 13B} & 13B & 43.9$_{\pm1.3\hphantom{0}}$ & 44.2$_{\pm2.7\hphantom{0}}$ & 72.6$_{\pm2.6\hphantom{0}}$ & 34.2$_{\pm2.6\hphantom{0}}$ & 4.1 & 44.0\\
\textsc{Flan-UL2} & 20B & 57.9$_{\pm0.2\hphantom{0}}$ & 66.0$_{\pm4.9\hphantom{0}}$ & 87.7$_{\pm4.6\hphantom{0}}$ & 58.4$_{\pm5.0\hphantom{0}}$ & 17.8 & 64.9\\
% \textsc{LLaMA 7B} & 7B & 29.1$_{\pm2.3\hphantom{0}}$ & 34.9$_{\pm2.5\hphantom{0}}$ & 64.3$_{\pm4.3\hphantom{0}}$ & 24.5$_{\pm2.4\hphantom{0}}$ & 1.5 & 35.5\\
% \textsc{LLaMA 13B} & 13B & 36.8$_{\pm3.5\hphantom{0}}$ & 35.6$_{\pm2.5\hphantom{0}}$ & 68.1$_{\pm3.0\hphantom{0}}$ & 24.2$_{\pm2.8\hphantom{0}}$ & 2.4 & 36.0\\
%\textsc{LLaMA 33B} & 33B & 53.6$_{\pm0.4\hphantom{0}}$ & 47.8$_{\pm4.5\hphantom{0}}$ & 77.8$_{\pm3.9\hphantom{0}}$ & 37.2$_{\pm4.8\hphantom{0}}$ & 5.7 & 47.6\\
%\textsc{LLaMA 65B} & 65B & 66.2$_{\pm0.7\hphantom{0}}$ & 58.9$_{\pm6.3\hphantom{0}}$ & 86.5$_{\pm1.7\hphantom{0}}$ & 49.1$_{\pm7.6\hphantom{0}}$ & 12.0 & 58.0\\
% \textsc{Vicuna 7B} & 7B & 35.0$_{\pm1.1\hphantom{0}}$ & 39.9$_{\pm3.9\hphantom{0}}$ & 60.2$_{\pm8.9\hphantom{0}}$ & 32.7$_{\pm4.4\hphantom{0}}$ & 3.2 & 39.8\\
\textsc{InstructGPT} & N/A & 70.2$_{\pm0.4\hphantom{0}}$ & \textbf{70.1}$_{\pm2.3\hphantom{0}}$ & 90.0$_{\pm3.5\hphantom{0}}$ & \textbf{63.0}$_{\pm2.0\hphantom{0}}$ & \textbf{23.1} & \textbf{69.2}\\
\textsc{LLaMA2 13B} & 13B & 47.7$_{\pm3.0\hphantom{0}}$ & 46.3$_{\pm4.0\hphantom{0}}$ & 80.0$_{\pm2.1\hphantom{0}}$ & 34.4$_{\pm4.7\hphantom{0}}$ & 5.1 & 47.1\\
\textsc{LLaMA2 70B} & 70B & \textbf{78.9}$_{\pm0.6\hphantom{0}}$ & 64.0$_{\pm4.8\hphantom{0}}$ & \textbf{90.6}$_{\pm2.5\hphantom{0}}$ & 54.6$_{\pm5.5\hphantom{0}}$ & 21.1 & 63.5\\
\textsc{Mistral 7B} & 7B & 58.2$_{\pm1.6\hphantom{0}}$ & 57.5$_{\pm5.4\hphantom{0}}$ & 88.1$_{\pm3.0\hphantom{0}}$ & 46.7$_{\pm7.3\hphantom{0}}$ & 9.4 & 57.2\\
\multicolumn{8}{c}{\cellcolor{light-gray} \emph{Zero-Shot}} \\
% \textsc{UnifiedQA-v2-Base} & 220M & 30.4\hphantom{$_{\pm0.00}$} & 42.2$_{\pm1.0\hphantom{0}}$ & 48.5$_{\pm2.9\hphantom{0}}$ & 39.9$_{\pm1.3\hphantom{0}}$ & 2.7 & 41.7\\
% \textsc{UnifiedQA-v2-Large} & 770M & 41.4\hphantom{$_{\pm0.00}$} & 42.9$_{\pm1.8\hphantom{0}}$ & 55.0$_{\pm4.9\hphantom{0}}$ & 38.5$_{\pm1.3\hphantom{0}}$ & 3.3 & 41.9\\
\textsc{UnifiedQA-v2-3B} & 3B & 45.5\hphantom{$_{\pm0.00}$} & 47.9$_{\pm2.1\hphantom{0}}$ & 71.6$_{\pm2.9\hphantom{0}}$ & 39.4$_{\pm2.2\hphantom{0}}$ & 5.7 & 47.8\\
\textsc{UnifiedQA-v2-11B} & 11B & 55.2\hphantom{$_{\pm0.00}$} & 57.3$_{\pm2.7\hphantom{0}}$ & 74.8$_{\pm5.2\hphantom{0}}$ & 51.1$_{\pm2.7\hphantom{0}}$ & 9.7 & 56.5\\
\textsc{Flan-T5-XXL} & 11B & 60.0\hphantom{$_{\pm0.00}$} & 64.3$_{\pm4.0\hphantom{0}}$ & 86.2$_{\pm5.4\hphantom{0}}$ & 56.5$_{\pm3.3\hphantom{0}}$ & 14.7 & 63.4\\
\textsc{Vicuna 13B} & 13B & 44.2\hphantom{$_{\pm0.00}$} & 49.5$_{\pm2.7\hphantom{0}}$ & 77.1$_{\pm1.7\hphantom{0}}$ & 39.7$_{\pm2.7\hphantom{0}}$ & 6.2 & 49.4\\
\textsc{Flan-UL2} & 20B & 56.2\hphantom{$_{\pm0.00}$} & \textbf{65.7}$_{\pm5.2\hphantom{0}}$ & 84.5$_{\pm4.4\hphantom{0}}$ & \textbf{59.1}$_{\pm5.1\hphantom{0}}$ & 14.7 & \textbf{64.2}\\
% \textsc{LLaMA 7B} & 7B & 27.7\hphantom{$_{\pm0.00}$} & 27.4$_{\pm4.1\hphantom{0}}$ & 38.2$_{\pm4.2\hphantom{0}}$ & 23.6$_{\pm4.6\hphantom{0}}$ & 0.8 & 27.1\\
% \textsc{LLaMA 13B} & 13B & 31.7\hphantom{$_{\pm0.00}$} & 36.0$_{\pm3.3\hphantom{0}}$ & 59.3$_{\pm2.1\hphantom{0}}$ & 27.8$_{\pm3.7\hphantom{0}}$ & 1.4 & 36.7\\
%\textsc{LLaMA 33B} & 33B & 54.5\hphantom{$_{\pm0.00}$} & 50.9$_{\pm4.4\hphantom{0}}$ & 81.3$_{\pm2.8\hphantom{0}}$ & 40.1$_{\pm4.5\hphantom{0}}$ & 6.8 & 50.9\\
%\textsc{LLaMA 65B} & 65B & 52.1\hphantom{$_{\pm0.00}$} & 47.9$_{\pm3.1\hphantom{0}}$ & 78.9$_{\pm1.8\hphantom{0}}$ & 36.9$_{\pm3.5\hphantom{0}}$ & 5.4 & 47.2\\
% \textsc{Vicuna 7B} & 7B & 36.7\hphantom{$_{\pm0.00}$} & 40.5$_{\pm2.3\hphantom{0}}$ & 73.1$_{\pm3.2\hphantom{0}}$ & 28.9$_{\pm2.9\hphantom{0}}$ & 2.4 & 40.1\\
\textsc{InstructGPT} & N/A & 64.1\hphantom{$_{\pm0.00}$} & 62.8$_{\pm2.2\hphantom{0}}$ & \textbf{89.9}$_{\pm2.0\hphantom{0}}$ & 53.2$_{\pm2.1\hphantom{0}}$ & \textbf{15.5} & 61.8\\
\textsc{InstructGPT+ CoT} & N/A & 63.8\hphantom{$_{\pm0.00}$} & 62.3$_{\pm1.0\hphantom{0}}$ & 89.6$_{\pm1.5\hphantom{0}}$ & 52.6$_{\pm1.5\hphantom{0}}$ & 14.2 & 61.2\\
\textsc{LLaMA2 13B} & 13B & 43.8\hphantom{$_{\pm0.00}$} & 44.4$_{\pm3.0\hphantom{0}}$ & 75.3$_{\pm3.1\hphantom{0}}$ & 33.5$_{\pm2.5\hphantom{0}}$ & 4.7 & 44.5\\
\textsc{LLaMA2 70B} & 70B & \textbf{70.8}\hphantom{$_{\pm0.00}$} & 58.0$_{\pm3.7\hphantom{0}}$ & 88.1$_{\pm3.4\hphantom{0}}$ & 47.3$_{\pm4.0\hphantom{0}}$ & 14.1 & 57.3\\
\textsc{Mistral 7B} & 7B & 54.0\hphantom{$_{\pm0.00}$} & 55.9$_{\pm3.2\hphantom{0}}$ & 
85.9$_{\pm2.0\hphantom{0}}$ & 45.3$_{\pm3.6\hphantom{0}}$ & 8.6 & 55.0\\
% \midrule \textsc{Human} & N/A & 86.5\hphantom{$_{\pm0.00}$} & 77.7\hphantom{$_{\pm0.00}$} & 87.9\hphantom{$_{\pm0.00}$} & 74.1\hphantom{$_{\pm0.00}$} & 53.0 & 81.5 \\   % average accuracy-based
\midrule \textsc{Human} & - & \textbf{91.5}\hphantom{$_{\pm0.00}$} & \textbf{82.6}\hphantom{$_{\pm0.00}$} & \textbf{93.0}\hphantom{$_{\pm0.00}$} & \textbf{78.9}\hphantom{$_{\pm0.00}$} & \textbf{52.9} & \textbf{81.5} \\   % majority vote-based

\bottomrule
\end{tabular}
%}
\caption{
Model performance on our RULE dataset consisting of the main questions (\textit{MainQ}) and subquestions (\textit{SubQ}).
We report the accuracy for the subquestions written about the correct option (\textit{Selective SubQ Acc.}) and incorrect options (\textit{Eliminative SubQ Acc.}) of the main questions.
Consistency holds only when the model answers both the main question and its subquestions correctly.
InstructGPT is \texttt{text-davinci-003}.
}
\label{tab:main-result}
\end{table*}

Table~\ref{tab:main-result} presents our main results.
In the fully-finetuned setting, we observe that % an improvement of the MainQ accuracy (e.g., 45.5\% to 66.0\% by UnifiedQA-v2 3B).
% In contrast, 
the SubQ accuracy % the performance of the models on the subquestions
does not significantly exceed the chance rate (25.0\%), which is far below the zero-shot performance of UnifiedQA-v2 as well as the human performance.
% We speculate that
This degradation may be due to overfitting to ReClor examples, by which the models rely heavily on superficial features of answer options that are not useful in answering the subquestions.
In our dataset, a group of subquestions shares the same set of four rationales, which requires that the models closely examine the question texts.

In the few- and zero-shot settings, we observe that the highest accuracy is 80.3\% on the main questions by LLaMA 2 70B with five-shot exemplars of ReClor and 65.7\% on the subquestions by Flan-UL2 in the zero-shot setting.
Both the MainQ and the SubQ accuracies are lower than the human accuracy by large margins ($\Delta=$ 11.2\%, 16.9\%), highlighting a severe limitation in the models' rationale understanding; in most cases, the models may only understand part of the necessary rationales for the comprehension process.

Although it is not our intended task setting, when we use a part of the subquestions for in-context learning, the highest SubQ accuracy is 70.1\% by InstructGPT in the five-shot setting.
This result is still below the human accuracy by a noticeable margin.
Interestingly, the in-context learning on subquestions is not helpful for smaller models such as %LLaMA 33B and Vicuna 7B and 13B.
Vicuna 7B and 13B.

Looking at the best Selective and Eliminative SubQ Accuracies, we find that although the former accuracy (five-shot LLaMA 2 70B, 90.0\%) is close to the human performance, % (89.9\% by the zero-shot InstructGPT vs. 87.9\% by humans), 
the latter accuracy (zero-shot Flan-UL2, 59.1\%) is significantly below the human performance (78.9\%). % (59.1\% by Flan-UL2 vs. 74.1\%).
This contrast shows that answering the eliminative subquestions is difficult for the models, highlighting the limited capacity of % recent models including strong
LLMs: Even if the models can choose the correct answer option, they may not understand why incorrect answer options should be refuted.

Consistency and MainQ-wise SubQ Accuracy also conform to this trend.
Although the consistency by humans is not high (52.9\%), probably owing to the difficulty of the subquestions, a large margin still exists between the human consistency and the best consistency by InstructGPT (18.2\%).
MainQ-wise SubQ Accuracy provides a bit more intuitive observation: The best model answers only 64.3\% of the subquestions per one main question, although humans get them wrong less often (81.5\%).
We report the detailed number of MainQ-wise SubQ Accuracy in Appendix~\ref{app:subq-result-distribution}.

Contrary to our expectations, CoT does not improve the performance of InstructGPT.
Rather, it leads to a decline in the MainQ and SubQ accuracies. % overall performance in both main questions and subquestions.
% In fact, it led to a decline in overall performance. 
This result is consistent with findings on the unreliable nature of CoT \cite{wang-etal-2023-towards,turpin2023language}, which may be exposed by the complexity of critical reasoning.  % logical reasoning questions, such as those in ReClor. 
% In contrast to arithmetic reasoning \cite{patel-etal-2021-nlp, Cobbe2021training} that has a limited variety of possible explanations, logical reasoning questions can have varied reasoning processes that lead to the correct answer.
% This diversity might make it more difficult for InstructGPT to find valid reasoning pathways.

\begin{table}[t]
    \centering
    \begin{tabular}{lccc} \toprule
        Batch & Acc. & \# \textit{None} in shot & \textit{None} Acc. \\ \midrule
        \#1 & 70.1 & 0 & 10.3 \\
        \#2 & 69.7 & 0 & 25.9 \\
        \#3 & 72.9 & 0 & 0.0 \\
        \#4 & 71.3 & 1 & 43.8 \\
        \#5 & 66.0 & 1 & 40.6 \\ \midrule
       Avg. & 70.1 & 0.4 & 32.0 \\ 
        \bottomrule
    \end{tabular}
    \caption{
        Accuracy of the subquestions that have ``None of the above choices'' as the correct answer (\textit{None Acc}), compared to that of all subquestions (\textit{Acc}).
        \textit{None in shot} indicates how many ``None'' examples are included in the few-shot prompt for each test split.
    }
    \label{tab:none-of-the-above}
\end{table}

\paragraph{Does the Model Answer ``None of the above choices'' Questions Correctly?}
Some of our subquestions contain ``None of the above choices,'' which might make the questions challenging. % and possibly ambiguous.
In particular, the model performance on this type of question might be strongly affected by the in-context learning of exemplars.
% To investigate this hypothesis, we compare the accuracy of the subquestions with that of the subquestions that include it as the correct answer.
To investigate this hypothesis, we calculate the accuracy of the subquestions that include the ``None'' option. % as the correct answer.
In the five-shot InstructGPT using RULE examples, we find that although the model achieves 62.7\% accuracy for the subquestions that have the ``None'' option, % which is a 7.4-point reduction from the overall performance (70.1\%),
it shows 32.0\% when %for those that have 
``None'' is the correct answer.
This low accuracy is decomposed into 40.9\% accuracy if the prompt includes the ``None'' option as the correct answer and 13.7\% accuracy otherwise.
These results demonstrate that using exemplars helps to answer those questions to some extent but not significantly.
% Appendix~\ref{app:none-of-the-above} reports detailed results.
Table~\ref{tab:none-of-the-above} reports the accuracy of five-shot InstructGPT across the five batches.

% Because of the length limit for this paper,
We report the complementary results of the main experiment in Appendix~\ref{app:one-shot-result}, in which the one-shot setting does not improve the model performance consistently.
Appendix~\ref{app:correct-only-result} shows the SubQ accuracy only for the main questions the models answer correctly. 
% We also report the performance plot across the question and option length in Appendix~\ref{app:length-and-performance}. %; we do not observe any strong trends for it.
Appendix~\ref{app:length-and-performance} shows the performance plot across the question and option length. % in Appendix~\ref{app:length-and-performance}. %; 

\section{Analysis}
\label{sec:analysis}

To qualitatively investigate the models' behavior observed in Section~\ref{sec:experiments}, we aim to answer the following research questions.

% \subsection{Rationale Choice for Probing Question Understanding}

\paragraph{Why Are the Eliminative Subquestions Difficult?}

As discussed in the previous section, we find a performance discrepancy between the selective and eliminative subquestions. % written for correct options of main questions and the ones for incorrect options.
We attribute this discrepancy to two potential reasons.
First, the eliminative subquestions are inherently complex because of the negation included in their question text, which the models may find difficult to handle \cite{ravichander-etal-2022-condaqa}.
Second, the model may lack the ability to comprehend why certain options are incorrect, which is partially supported by studies that highlight the susceptibility for distractors in the multiple-choice QA \cite{si-etal-2021-benchmarking}.

%For instance, the study referenced as \hoge revealed a notable lack of robustness for distractors. 
%They achieved this by developing an adversarial dataset from RACE, termed AdvRACE. 
%The dataset was constructed by substituting original distractors in RACE with irrelevant texts.
%Surprisingly, a simple replacement that humans can easily detect resulted in a marked decrease in performance compared to the original RACE dataset. 
%This underpins the claim that the models may be prone to distractor vulnerability, supporting the second hypothesis.

%This hypothesis, if validated, questions the efficacy of multiple-choice Q&A in assessing the capabilities of models. 
%The format inherently requires participants to discern not only the correct answer but also reject incorrect options.

To distinguish between the difficulty of comprehending complex questions and that of refuting relevant alternatives in the eliminative subquestions, we develop a follow-up task, \textit{rationale alignment}.
%To investigate this, we designed an experiment focused on the model's understanding of the rationale behind incorrect answer options, excluding the influence of the question. 
In this task, given a context, the main question, one of the main options, and four rationales, the model selects one out of the four rationales that validates the correctness of the given option.
% The model is given a context, a question, and one of the options of a main question, and four rationales for the main question.
% Then the model is asked to select which rationale validates the correctness of the given option.
%The experiment required the model to match an option to its corresponding rationale.
%In this task, the model is given main-question and rationale, and the model have to answer which option corresponds to the rationale.
% Similar to the main experiment, we will report the average results from five prompts. 
% Each prompt is composed of five exemplars, with two exemplars presenting the correct option and three exemplars presenting the incorrect option.
We use InstructGPT in the five-shot setting and report the average results from five different prompts.
Appendix~\ref{app:rationale-alignment} provides the input prompt.

Because the subquestion text is not used in this task, we expect that the results are not affected by the complexity of subquestion texts.
% The result presented in Table~\ref{tab:rationale-alignment} shows a distinct difference in the model performance between the correct and incorrect options. %  when given the correct option rationale versus an incorrect one. 
The result is 89.7\% and 31.5\% accuracy for the correct and incorrect answer options, respectively, showing a distinct difference between them.
% This result shows a distinct difference in the model performance between the correct and incorrect options.
This discrepancy suggests the model's serious deficiency in comprehending eliminative rationales.  % the rationales that are written for incorrect options.
%Since this experiment eliminates the impact of complex sub-question text, it provides support for the second hypothesis, suggesting a potential vulnerability in the model's understanding of distractors.

\paragraph{Is the Model Better at Writing Rationales than Humans?}
Given that CoT does not improve the model performance, we are interested in the quality and potential usefulness of model-generated rationales compared to our human-written rationales.
% We compare our human-written rationales with InstructGPT-generated rationales.
We use a similar prompt to that used in our CoT setting, instructing InstructGPT to generate rationales for 50 options.
We then randomly shuffle the order of human-written and model-generated rationales, and manually annotate which rationale is better in terms of necessity and specificity.
The result is 35 wins by humans and 15 wins by the model among the 50 comparisons, showing that the human-written rationales are likely to be more detailed and supportive than the model-generated ones.
In particular, we find that the model rationales struggle to capture the \textit{implicit} rationale necessary for certifying the validity of the target option.
When the rationale is explicit and described well in the context, the model rationale looks convincing and close to the human rationale.
Among the 15 examples where humans lose, we find five examples unsatisfactory to validate the target option, implying that approximately 10\% of unreasonable rationales are potentially included in our dataset.

%%%%%%%%%%%%%%%%%%%%%%%%%%

\begin{table}[t]
    \centering % \small
    \begin{tabular}{lccc} \toprule
        & Direct & Indirect & Total\\ \midrule
        Contextual & 37 / 47 & 28 / 22 & 65 / 69\\
        External & 22 / 20 & 13 / 11 & 35 / 31 \\ \midrule
        Total & 59 / 67 & 41 / 33 & 100 \\
        \bottomrule
    \end{tabular}
    \caption{
        Annotation results of rationale types on 100 examples randomly sampled from all subquestions (left) and from the error examples by InstructGPT (right).
    }
    \label{tab:rationale-type}
\end{table}

\paragraph{What Types of Reasoning are Required in the Rationale Understanding?}
% \label{sec:rationale-type}
% \paragraph{What types of rationales does the dataset have and on what types does the model make errors?}

To qualitatively analyze the collected rationales, we first sample 100 subquestions to annotate reasoning types.
We define two dichotomies: \textit{direct/indirect} and \textit{contextual/external}.
Direct reasoning occurs if a rationale involves an explicit description for the certification of a target option's (in)validity, whereas indirect reasoning only provides relevant facts for the validity.
Context reasoning includes facts (or their interpretation and summarization) described in the context, while external reasoning is pertinent to commonsense and norms that are not described in the context.
% The detailed definitions are described in Appendix~\ref{app:rationale-type}.
For comparative error analysis, we also sample 100 subquestions among those that InstructGPT answers incorrectly.  % to analyze what types of rationales the model may not be good at.

We report our annotation results %the distributions of reasoning types 
in Table~\ref{tab:rationale-type}.
The number of the direct and contextual rationales is the largest among the other types, which further increases when we look at the error cases of InstructGPT.
We find that our dataset covers a sufficient number of indirect and external reasoning, i.e., various modes of rationale understanding.
Error examples for the four reasoning types are reported in Appendix~\ref{app:reasoning-type}.
% Although this analysis is not supported by statistically significant evidence, it may help to grasp the trends of dataset examples and error cases in terms of reasoning types.
%
Although we also examine the reasoning types originally labeled in the ReClor dataset, we do not observe any remarkable trends in the subquestion accuracy (Appendix~\ref{app:original-reasoning-type}).

%%%%%%%%%%%%%%%%%%%%%%%%%%

\begin{table}[t]
    \centering % \small
    \begin{tabular}{lc}
        \toprule
        Input & Accuracy \\ \midrule
        Context & 72.2 \\
        + Selective Rationale & 91.4 \\
        + Eliminative Rationale & 66.0 \\
        + Both & 89.6 \\
        \bottomrule
    \end{tabular}
    \caption{
        % MainQ accuracy of InstructGPT that uses the rationales written for the correct options (selective) or the incorrect options (eliminative).
        MainQ accuracy of InstructGPT that uses the selective or eliminative rationales in the input.
        % The selective rationales improve the accuracy, whereas the eliminative rationales do not.
    }
    \label{tab:context-addition}
\end{table}

\paragraph{Do the Rationales Help the Model to Answer the Main Questions?}
Because the collected rationales are expected to support the decision of selecting and eliminating answer options, we investigate whether adding the rationales to the main questions improves the performance in the five-shot InstructGPT.
We append the rationale to the context, main question, and four options with the \texttt{Rationale:} label.
% We use the main questions that have at least one valid selective rationale and one valid eliminative rationale. % written for the correct option and one incorrect option. %  (excluding ``None of the above choices'') 
% Table~\ref{tab:context-addition} shows the MainQ accuracy when we add the selective rationale, eliminative rationale, and both.
The results are shown in Table~\ref{tab:context-addition}.
We observe an improvement when the selective rationale is added; however, degradation occurs when we add the eliminative rationale, even if it is provided with the selective rationale.
This result adds insight to the observation by \citet{sun-etal-2022-investigating}, showing that the model cannot use eliminative rationales for answering main questions and becomes confused by those rationales.
We also investigate the context-ablated setting in Appendix~\ref{app:context-ablation}.

%%%%%%%%%%%%%%%%%%%%%%%%%%

% \begin{table}[t]
%     \centering \small
%     \begin{tabular}{lcccc} \toprule
%     Setting & MainQ & SubQ & \makecell[c]{Selective \\ SubQ} & \makecell[c]{Eliminat.\\ SubQ} \\ \midrule
%     0-shot & 41.0$_{-23.0}$ & 58.8$_{-4.2}$ & 86.4$_{-3.7}$ & 53.4$_{-4.3}$ \\
%     5-shot & 42.5$_{-29.7}$ & 70.5$_{+0.0}$ & 86.9$_{-3.1}$ & 64.7$_{+1.7}$ \\
%     \bottomrule
%     \end{tabular}
%     \caption{
%         Context-ablated accuracy.
%         The subscript values indicate the accuracy gap against the full-input setting.
%         % Accuracy with the context ablated.
%         % The subscript values indicate the performance difference against the full-input setting.
%     }
%     \label{tab:context-ablation}
% \end{table}

% \paragraph{Does the Context Help in Answering Subquestions?}
% We also conduct input-ablation analysis.
% % As the subquestions of a main question share the same set of answer options, we cannot ablate the subquestion texts (chance rate is 31.4\%).
% By removing the context, we analyze the model performance on the subquestions (and the main questions for reference) to see the dependency between question texts and answer options.
% The results in Table~\ref{tab:context-ablation} show the performance reduction by approximately 4 points in the zero-shot setting and no reduction in the five-shot setting.
% This result implies question texts depend on answer options to some extent, which potentially makes the subquestions difficult for the models, given the first analysis in this section.

\section{Conclusion}

% In this study,  % new benchmarking
We construct a dataset to evaluate the models' ability of critical reasoning in logical reading comprehension.
% Using crowdsourcing,
We crowdsource free-form rationale for main questions taken from an existing dataset and use an LLM to generate subquestion texts.
Resulting questions ask about the underlying rationales for why a certain answer option should be selected and the others should be eliminated.
We find that LLMs are particularly bad at answering eliminative subquestions, highlighting that those models do not necessarily have the comprehensive ability of critical reasoning.
For future work, we will % conduct a detailed analysis of rationales, 
develop a more efficient pipeline for data collection and facilitate better rationale generation by LLMs.

\section*{Ethical Consideration}

We use crowdsourcing in our data collection.
We make sure to be responsible to the crowdworkers and to make fair compensation for their work.
We do not collect any personal information other than worker IDs on the platform, which are removed in our data release.
Before the workers accept our tasks, we inform them of our purpose for the data collection.
This study is approved by the internal review board of the authors' institutes.

\section*{Limitations}

We recognize the following limitations in this study.

% \paragraph{Limited Evaluation Methodology}
% Our finding that "the language model is poor at understanding rationales" is based on the evaluation of multiple-choice questions and the author's manual assessment. 
% The inherent constraints of the multiple-choice question methodology limit the flexibility of measurable responses. 
% This means that the results of such questions may not fully represent the language model's capabilities.

\paragraph{Task Format}
In this study, we focus on the multiple-choice QA task.
This task format allows us to flexibly ask about various linguistic phenomena and human reasoning by selecting and eliminating alternatives, and we consider solving such a discriminative task would be a minimal requirement for human-like linguistic behaviors.
However, it has an inherent limitation in assessing the ability of natural language understanding.
For example, we cannot evaluate the models' ability to produce an intended output.

\paragraph{Annotation Analysis}
We conduct the annotation analysis in Section~\ref{sec:analysis}, in which we define the reasoning types and manually review the sampled examples.
Although we make our annotation data and guideline publicly available for ensuring the reproducibility of annotation results, the results of our annotation analysis inevitably involve our subjective judgments.
% The author's manual assessments could introduce a degree of subjectivity into the results.

\paragraph{Source Dataset}
We create our auxiliary questions on top of an existing English logical reading comprehension dataset, ReClor.
Although our methodology of the data collection (i.e., writing the rationale for selecting and eliminating alternatives) is widely applicable to other datasets and languages, using the single dataset in the single language would limit the generalizability of our findings.

% \paragraph{Limited Generalizability across Datasets}
% Our study is conducted solely on the basis of ReClor, and we have not tested whether similar results would be observed with other datasets that assess similar capabilities. 
% This could potentially limit the external validity and generalizability of our findings across different datasets.

% \section*{Acknowledgements}

\section*{Acknowledgments}
We would like to thank the anonymous reviewers for their helpful comments.
This work was supported by JST PRESTO Grant Number JPMJPR20C4 and JSPS KAKENHI Grant Number 22K17954.

\bibliography{anthology,custom}

\begin{thebibliography}{46}
\expandafter\ifx\csname natexlab\endcsname\relax\def\natexlab#1{#1}\fi

\bibitem[{Aggarwal et~al.(2021)Aggarwal, Mandowara, Agrawal, Khandelwal, Singla, and Garg}]{aggarwal-etal-2021-explanations}
Shourya Aggarwal, Divyanshu Mandowara, Vishwajeet Agrawal, Dinesh Khandelwal, Parag Singla, and Dinesh Garg. 2021.
\newblock \href {https://doi.org/10.18653/v1/2021.acl-long.238} {{E}xplanations for {C}ommonsense{QA}: {N}ew {D}ataset and {M}odels}.
\newblock In \emph{Proceedings of the 59th Annual Meeting of the Association for Computational Linguistics and the 11th International Joint Conference on Natural Language Processing (Volume 1: Long Papers)}, pages 3050--3065, Online. Association for Computational Linguistics.

\bibitem[{Ashida and Sugawara(2022)}]{ashida-sugawara-2022-possible}
Mana Ashida and Saku Sugawara. 2022.
\newblock \href {https://aclanthology.org/2022.coling-1.319} {Possible stories: Evaluating situated commonsense reasoning under multiple possible scenarios}.
\newblock In \emph{Proceedings of the 29th International Conference on Computational Linguistics}, pages 3606--3630, Gyeongju, Republic of Korea. International Committee on Computational Linguistics.

\bibitem[{Bhagavatula et~al.(2020)Bhagavatula, Bras, Malaviya, Sakaguchi, Holtzman, Rashkin, Downey, tau Yih, and Choi}]{Bhagavatula2020Abductive}
Chandra Bhagavatula, Ronan~Le Bras, Chaitanya Malaviya, Keisuke Sakaguchi, Ari Holtzman, Hannah Rashkin, Doug Downey, Wen tau Yih, and Yejin Choi. 2020.
\newblock \href {https://openreview.net/forum?id=Byg1v1HKDB} {Abductive commonsense reasoning}.
\newblock In \emph{International Conference on Learning Representations}.

\bibitem[{Brown et~al.(2020)Brown, Mann, Ryder, Subbiah, Kaplan, Dhariwal, Neelakantan, Shyam, Sastry, Askell, Agarwal, Herbert-Voss, Krueger, Henighan, Child, Ramesh, Ziegler, Wu, Winter, Hesse, Chen, Sigler, Litwin, Gray, Chess, Clark, Berner, McCandlish, Radford, Sutskever, and Amodei}]{Brown-etal-2020-gpt3}
Tom Brown, Benjamin Mann, Nick Ryder, Melanie Subbiah, Jared~D Kaplan, Prafulla Dhariwal, Arvind Neelakantan, Pranav Shyam, Girish Sastry, Amanda Askell, Sandhini Agarwal, Ariel Herbert-Voss, Gretchen Krueger, Tom Henighan, Rewon Child, Aditya Ramesh, Daniel Ziegler, Jeffrey Wu, Clemens Winter, Chris Hesse, Mark Chen, Eric Sigler, Mateusz Litwin, Scott Gray, Benjamin Chess, Jack Clark, Christopher Berner, Sam McCandlish, Alec Radford, Ilya Sutskever, and Dario Amodei. 2020.
\newblock \href {https://proceedings.neurips.cc/paper_files/paper/2020/file/1457c0d6bfcb4967418bfb8ac142f64a-Paper.pdf} {Language models are few-shot learners}.
\newblock In \emph{Advances in Neural Information Processing Systems}, volume~33, pages 1877--1901. Curran Associates, Inc.

\bibitem[{Card et~al.(2020)Card, Henderson, Khandelwal, Jia, Mahowald, and Jurafsky}]{card-etal-2020-little}
Dallas Card, Peter Henderson, Urvashi Khandelwal, Robin Jia, Kyle Mahowald, and Dan Jurafsky. 2020.
\newblock \href {https://doi.org/10.18653/v1/2020.emnlp-main.745} {With little power comes great responsibility}.
\newblock In \emph{Proceedings of the 2020 Conference on Empirical Methods in Natural Language Processing (EMNLP)}, pages 9263--9274, Online. Association for Computational Linguistics.

\bibitem[{Chiang et~al.(2023)Chiang, Li, Lin, Sheng, Wu, Zhang, Zheng, Zhuang, Zhuang, Gonzalez, Stoica, and Xing}]{vicuna2023}
Wei-Lin Chiang, Zhuohan Li, Zi~Lin, Ying Sheng, Zhanghao Wu, Hao Zhang, Lianmin Zheng, Siyuan Zhuang, Yonghao Zhuang, Joseph~E. Gonzalez, Ion Stoica, and Eric~P. Xing. 2023.
\newblock \href {https://lmsys.org/blog/2023-03-30-vicuna/} {Vicuna: An open-source chatbot impressing {GPT}-4 with 90\%* {ChatGPT} quality}.

\bibitem[{Chung et~al.(2022)Chung, Hou, Longpre, Zoph, Tay, Fedus, Li, Wang, Dehghani, Brahma, Webson, Gu, Dai, Suzgun, Chen, Chowdhery, Castro-Ros, Pellat, Robinson, Valter, Narang, Mishra, Yu, Zhao, Huang, Dai, Yu, Petrov, Chi, Dean, Devlin, Roberts, Zhou, Le, and Wei}]{chung2022scaling}
Hyung~Won Chung, Le~Hou, Shayne Longpre, Barret Zoph, Yi~Tay, William Fedus, Yunxuan Li, Xuezhi Wang, Mostafa Dehghani, Siddhartha Brahma, Albert Webson, Shixiang~Shane Gu, Zhuyun Dai, Mirac Suzgun, Xinyun Chen, Aakanksha Chowdhery, Alex Castro-Ros, Marie Pellat, Kevin Robinson, Dasha Valter, Sharan Narang, Gaurav Mishra, Adams Yu, Vincent Zhao, Yanping Huang, Andrew Dai, Hongkun Yu, Slav Petrov, Ed~H. Chi, Jeff Dean, Jacob Devlin, Adam Roberts, Denny Zhou, Quoc~V. Le, and Jason Wei. 2022.
\newblock \href {https://arxiv.org/abs/2210.11416} {Scaling instruction-finetuned language models}.
\newblock {a}rXiv preprint 2210.11416.

\bibitem[{Dalvi et~al.(2021)Dalvi, Jansen, Tafjord, Xie, Smith, Pipatanangkura, and Clark}]{dalvi-etal-2021-explaining}
Bhavana Dalvi, Peter Jansen, Oyvind Tafjord, Zhengnan Xie, Hannah Smith, Leighanna Pipatanangkura, and Peter Clark. 2021.
\newblock \href {https://doi.org/10.18653/v1/2021.emnlp-main.585} {Explaining answers with entailment trees}.
\newblock In \emph{Proceedings of the 2021 Conference on Empirical Methods in Natural Language Processing}, pages 7358--7370, Online and Punta Cana, Dominican Republic. Association for Computational Linguistics.

\bibitem[{Gardner et~al.(2020)Gardner, Artzi, Basmov, Berant, Bogin, Chen, Dasigi, Dua, Elazar, Gottumukkala, Gupta, Hajishirzi, Ilharco, Khashabi, Lin, Liu, Liu, Mulcaire, Ning, Singh, Smith, Subramanian, Tsarfaty, Wallace, Zhang, and Zhou}]{gardner-etal-2020-evaluating}
Matt Gardner, Yoav Artzi, Victoria Basmov, Jonathan Berant, Ben Bogin, Sihao Chen, Pradeep Dasigi, Dheeru Dua, Yanai Elazar, Ananth Gottumukkala, Nitish Gupta, Hannaneh Hajishirzi, Gabriel Ilharco, Daniel Khashabi, Kevin Lin, Jiangming Liu, Nelson~F. Liu, Phoebe Mulcaire, Qiang Ning, Sameer Singh, Noah~A. Smith, Sanjay Subramanian, Reut Tsarfaty, Eric Wallace, Ally Zhang, and Ben Zhou. 2020.
\newblock \href {https://doi.org/10.18653/v1/2020.findings-emnlp.117} {Evaluating models{'} local decision boundaries via contrast sets}.
\newblock In \emph{Findings of the Association for Computational Linguistics: EMNLP 2020}, pages 1307--1323, Online. Association for Computational Linguistics.

\bibitem[{Geirhos et~al.(2020)Geirhos, Jacobsen, Michaelis, Zemel, Brendel, Bethge, and Wichmann}]{Geirhos2020-zj}
Robert Geirhos, J{\"o}rn-Henrik Jacobsen, Claudio Michaelis, Richard Zemel, Wieland Brendel, Matthias Bethge, and Felix~A Wichmann. 2020.
\newblock \href {https://doi.org/10.1038/s42256-020-00257-z} {Shortcut learning in deep neural networks}.
\newblock \emph{Nature Machine Intelligence}, 2(11):665--673.

\bibitem[{Geva et~al.(2021)Geva, Khashabi, Segal, Khot, Roth, and Berant}]{geva-etal-2021-aristotle}
Mor Geva, Daniel Khashabi, Elad Segal, Tushar Khot, Dan Roth, and Jonathan Berant. 2021.
\newblock \href {https://doi.org/10.1162/tacl_a_00370} {Did aristotle use a laptop? a question answering benchmark with implicit reasoning strategies}.
\newblock \emph{Transactions of the Association for Computational Linguistics}, 9:346--361.

\bibitem[{Gierl et~al.(2017)Gierl, Bulut, Guo, and Zhang}]{Girel-et-al-distractor}
Mark Gierl, Okan Bulut, Qi~Guo, and Xinxin Zhang. 2017.
\newblock \href {https://doi.org/10.3102/0034654317726529} {Developing, analyzing, and using distractors for multiple-choice tests in education: A comprehensive review}.
\newblock \emph{Review of Educational Research}, 87:0034654317726529.

\bibitem[{Habernal et~al.(2018)Habernal, Wachsmuth, Gurevych, and Stein}]{habernal-etal-2018-argument}
Ivan Habernal, Henning Wachsmuth, Iryna Gurevych, and Benno Stein. 2018.
\newblock \href {https://doi.org/10.18653/v1/N18-1175} {The argument reasoning comprehension task: Identification and reconstruction of implicit warrants}.
\newblock In \emph{Proceedings of the 2018 Conference of the North {A}merican Chapter of the Association for Computational Linguistics: Human Language Technologies, Volume 1 (Long Papers)}, pages 1930--1940, New Orleans, Louisiana. Association for Computational Linguistics.

\bibitem[{He et~al.(2023)He, Gao, and Chen}]{he2023debertav}
Pengcheng He, Jianfeng Gao, and Weizhu Chen. 2023.
\newblock \href {https://openreview.net/forum?id=sE7-XhLxHA} {De{BERT}av3: Improving de{BERT}a using {ELECTRA}-style pre-training with gradient-disentangled embedding sharing}.
\newblock In \emph{The Eleventh International Conference on Learning Representations}.

\bibitem[{Huang et~al.(2019)Huang, Le~Bras, Bhagavatula, and Choi}]{huang-etal-2019-cosmos}
Lifu Huang, Ronan Le~Bras, Chandra Bhagavatula, and Yejin Choi. 2019.
\newblock \href {https://doi.org/10.18653/v1/D19-1243} {Cosmos {QA}: Machine reading comprehension with contextual commonsense reasoning}.
\newblock In \emph{Proceedings of the 2019 Conference on Empirical Methods in Natural Language Processing and the 9th International Joint Conference on Natural Language Processing (EMNLP-IJCNLP)}, pages 2391--2401, Hong Kong, China. Association for Computational Linguistics.

\bibitem[{Huang et~al.(2022)Huang, Zhang, Hong, Liang, Zhang, and Yu}]{huang-etal-2022-metalogic}
Yinya Huang, Hongming Zhang, Ruixin Hong, Xiaodan Liang, Changshui Zhang, and Dong Yu. 2022.
\newblock \href {https://doi.org/10.18653/v1/2022.emnlp-main.310} {{M}eta{L}ogic: Logical reasoning explanations with fine-grained structure}.
\newblock In \emph{Proceedings of the 2022 Conference on Empirical Methods in Natural Language Processing}, pages 4698--4724, Abu Dhabi, United Arab Emirates. Association for Computational Linguistics.

\bibitem[{Inoue et~al.(2020)Inoue, Stenetorp, and Inui}]{inoue-etal-2020-r4c}
Naoya Inoue, Pontus Stenetorp, and Kentaro Inui. 2020.
\newblock \href {https://doi.org/10.18653/v1/2020.acl-main.602} {{R}4{C}: A benchmark for evaluating {RC} systems to get the right answer for the right reason}.
\newblock In \emph{Proceedings of the 58th Annual Meeting of the Association for Computational Linguistics}, pages 6740--6750, Online. Association for Computational Linguistics.

\bibitem[{Jiang et~al.(2023)Jiang, Sablayrolles, Mensch, Bamford, Chaplot, de~las Casas, Bressand, Lengyel, Lample, Saulnier, Lavaud, Lachaux, Stock, Scao, Lavril, Wang, Lacroix, and Sayed}]{jiang2023mistral}
Albert~Q. Jiang, Alexandre Sablayrolles, Arthur Mensch, Chris Bamford, Devendra~Singh Chaplot, Diego de~las Casas, Florian Bressand, Gianna Lengyel, Guillaume Lample, Lucile Saulnier, Lélio~Renard Lavaud, Marie-Anne Lachaux, Pierre Stock, Teven~Le Scao, Thibaut Lavril, Thomas Wang, Timothée Lacroix, and William~El Sayed. 2023.
\newblock \href {https://arxiv.org/abs/2310.06825} {Mistral 7{B}}.
\newblock {a}rXiv preprint 2310.06825.

\bibitem[{Jiao et~al.(2022)Jiao, Guo, Song, and Nie}]{jiao-etal-2022-merit}
Fangkai Jiao, Yangyang Guo, Xuemeng Song, and Liqiang Nie. 2022.
\newblock \href {https://doi.org/10.18653/v1/2022.findings-acl.276} {{MERI}t: {M}eta-{P}ath {G}uided {C}ontrastive {L}earning for {L}ogical {R}easoning}.
\newblock In \emph{Findings of the Association for Computational Linguistics: ACL 2022}, pages 3496--3509, Dublin, Ireland. Association for Computational Linguistics.

\bibitem[{Khashabi et~al.(2022)Khashabi, Kordi, and Hajishirzi}]{khashabi2022unifiedqav2}
Daniel Khashabi, Yeganeh Kordi, and Hannaneh Hajishirzi. 2022.
\newblock \href {https://arxiv.org/abs/2202.12359} {{UnifiedQA}-v2: Stronger generalization via broader cross-format training}.
\newblock {a}rXiv preprint 2202.12359.

\bibitem[{Khashabi et~al.(2020)Khashabi, Min, Khot, Sabharwal, Tafjord, Clark, and Hajishirzi}]{khashabi-etal-2020-unifiedqa}
Daniel Khashabi, Sewon Min, Tushar Khot, Ashish Sabharwal, Oyvind Tafjord, Peter Clark, and Hannaneh Hajishirzi. 2020.
\newblock \href {https://doi.org/10.18653/v1/2020.findings-emnlp.171} {{UNIFIEDQA}: Crossing format boundaries with a single {QA} system}.
\newblock In \emph{Findings of the Association for Computational Linguistics: EMNLP 2020}, pages 1896--1907, Online. Association for Computational Linguistics.

\bibitem[{Khot et~al.(2020)Khot, Clark, Guerquin, Jansen, and Sabharwal}]{Khot2020-ml}
Tushar Khot, Peter Clark, Michal Guerquin, Peter Jansen, and Ashish Sabharwal. 2020.
\newblock \href {https://ojs.aaai.org/index.php/AAAI/article/view/6319} {{QASC:} {A} dataset for question answering via sentence composition}.
\newblock In \emph{The Thirty-Fourth {AAAI} Conference on Artificial Intelligence}, pages 8082--8090. {AAAI} Press.

\bibitem[{Kojima et~al.(2022)Kojima, Gu, Reid, Matsuo, and Iwasawa}]{kojima2022large}
Takeshi Kojima, Shixiang~(Shane) Gu, Machel Reid, Yutaka Matsuo, and Yusuke Iwasawa. 2022.
\newblock \href {https://proceedings.neurips.cc/paper_files/paper/2022/file/8bb0d291acd4acf06ef112099c16f326-Paper-Conference.pdf} {Large language models are zero-shot reasoners}.
\newblock In \emph{Advances in Neural Information Processing Systems}, volume~35, pages 22199--22213. Curran Associates, Inc.

\bibitem[{Lin et~al.(2021)Lin, Zou, and Ding}]{lin-etal-2021-using}
Jieyu Lin, Jiajie Zou, and Nai Ding. 2021.
\newblock \href {https://doi.org/10.18653/v1/2021.acl-short.43} {Using adversarial attacks to reveal the statistical bias in machine reading comprehension models}.
\newblock In \emph{Proceedings of the 59th Annual Meeting of the Association for Computational Linguistics and the 11th International Joint Conference on Natural Language Processing (Volume 2: Short Papers)}, pages 333--342, Online. Association for Computational Linguistics.

\bibitem[{Liu et~al.(2020)Liu, Cui, Liu, Huang, Wang, and Zhang}]{Liu-etal-2020-logiqa}
Jian Liu, Leyang Cui, Hanmeng Liu, Dandan Huang, Yile Wang, and Yue Zhang. 2020.
\newblock \href {https://doi.org/10.24963/ijcai.2020/501} {Logi{QA}: A challenge dataset for machine reading comprehension with logical reasoning}.
\newblock In \emph{Proceedings of the Twenty-Ninth International Joint Conference on Artificial Intelligence, {IJCAI-20}}, pages 3622--3628. International Joint Conferences on Artificial Intelligence Organization.
\newblock Main track.

\bibitem[{Nangia et~al.(2021)Nangia, Sugawara, Trivedi, Warstadt, Vania, and Bowman}]{nangia-etal-2021-ingredients}
Nikita Nangia, Saku Sugawara, Harsh Trivedi, Alex Warstadt, Clara Vania, and Samuel~R. Bowman. 2021.
\newblock \href {https://doi.org/10.18653/v1/2021.acl-long.98} {What ingredients make for an effective crowdsourcing protocol for difficult {NLU} data collection tasks?}
\newblock In \emph{Proceedings of the 59th Annual Meeting of the Association for Computational Linguistics and the 11th International Joint Conference on Natural Language Processing (Volume 1: Long Papers)}, pages 1221--1235, Online. Association for Computational Linguistics.

\bibitem[{Niven and Kao(2019)}]{niven-kao-2019-probing}
Timothy Niven and Hung-Yu Kao. 2019.
\newblock \href {https://doi.org/10.18653/v1/P19-1459} {Probing neural network comprehension of natural language arguments}.
\newblock In \emph{Proceedings of the 57th Annual Meeting of the Association for Computational Linguistics}, pages 4658--4664, Florence, Italy. Association for Computational Linguistics.

\bibitem[{Ouyang et~al.(2022)Ouyang, Wu, Jiang, Almeida, Wainwright, Mishkin, Zhang, Agarwal, Slama, Ray, Schulman, Hilton, Kelton, Miller, Simens, Askell, Welinder, Christiano, Leike, and Lowe}]{Ouyang-etal-2022-instructgpt}
Long Ouyang, Jeffrey Wu, Xu~Jiang, Diogo Almeida, Carroll Wainwright, Pamela Mishkin, Chong Zhang, Sandhini Agarwal, Katarina Slama, Alex Ray, John Schulman, Jacob Hilton, Fraser Kelton, Luke Miller, Maddie Simens, Amanda Askell, Peter Welinder, Paul~F Christiano, Jan Leike, and Ryan Lowe. 2022.
\newblock \href {https://proceedings.neurips.cc/paper_files/paper/2022/file/b1efde53be364a73914f58805a001731-Paper-Conference.pdf} {Training language models to follow instructions with human feedback}.
\newblock In \emph{Advances in Neural Information Processing Systems}, volume~35, pages 27730--27744. Curran Associates, Inc.

\bibitem[{Ravichander et~al.(2022)Ravichander, Gardner, and Marasovic}]{ravichander-etal-2022-condaqa}
Abhilasha Ravichander, Matt Gardner, and Ana Marasovic. 2022.
\newblock \href {https://doi.org/10.18653/v1/2022.emnlp-main.598} {{CONDAQA}: A contrastive reading comprehension dataset for reasoning about negation}.
\newblock In \emph{Proceedings of the 2022 Conference on Empirical Methods in Natural Language Processing}, pages 8729--8755, Abu Dhabi, United Arab Emirates. Association for Computational Linguistics.

\bibitem[{Ribeiro et~al.(2023)Ribeiro, Wang, Ma, Zhu, Dong, Kong, Burger, Ramos, zhiheng huang, Wang, Karypis, Xiang, and Roth}]{ribeiro2023street}
Danilo~Neves Ribeiro, Shen Wang, Xiaofei Ma, Henghui Zhu, Rui Dong, Deguang Kong, Juliette Burger, Anjelica Ramos, zhiheng huang, William~Yang Wang, George Karypis, Bing Xiang, and Dan Roth. 2023.
\newblock \href {https://openreview.net/forum?id=1C_kSW1-k0} {{STREET}: A multi-task structured reasoning and explanation benchmark}.
\newblock In \emph{The Eleventh International Conference on Learning Representations}.

\bibitem[{Saha et~al.(2022)Saha, Hase, Rajani, and Bansal}]{saha-etal-2022-hard}
Swarnadeep Saha, Peter Hase, Nazneen Rajani, and Mohit Bansal. 2022.
\newblock \href {https://doi.org/10.18653/v1/2022.emnlp-main.137} {Are hard examples also harder to explain? a study with human and model-generated explanations}.
\newblock In \emph{Proceedings of the 2022 Conference on Empirical Methods in Natural Language Processing}, pages 2121--2131, Abu Dhabi, United Arab Emirates. Association for Computational Linguistics.

\bibitem[{Saha et~al.(2021)Saha, Yadav, Bauer, and Bansal}]{saha-etal-2021-explagraphs}
Swarnadeep Saha, Prateek Yadav, Lisa Bauer, and Mohit Bansal. 2021.
\newblock \href {https://doi.org/10.18653/v1/2021.emnlp-main.609} {{E}xpla{G}raphs: An explanation graph generation task for structured commonsense reasoning}.
\newblock In \emph{Proceedings of the 2021 Conference on Empirical Methods in Natural Language Processing}, pages 7716--7740, Online and Punta Cana, Dominican Republic. Association for Computational Linguistics.

\bibitem[{Shi et~al.(2023)Shi, Chen, Misra, Scales, Dohan, Chi, Sch\"{a}rli, and Zhou}]{pmlr-v202-shi23a}
Freda Shi, Xinyun Chen, Kanishka Misra, Nathan Scales, David Dohan, Ed~H. Chi, Nathanael Sch\"{a}rli, and Denny Zhou. 2023.
\newblock \href {https://proceedings.mlr.press/v202/shi23a.html} {Large language models can be easily distracted by irrelevant context}.
\newblock In \emph{Proceedings of the 40th International Conference on Machine Learning}, volume 202 of \emph{Proceedings of Machine Learning Research}, pages 31210--31227. PMLR.

\bibitem[{Si et~al.(2021)Si, Yang, Cui, Ma, Liu, and Wang}]{si-etal-2021-benchmarking}
Chenglei Si, Ziqing Yang, Yiming Cui, Wentao Ma, Ting Liu, and Shijin Wang. 2021.
\newblock \href {https://doi.org/10.18653/v1/2021.findings-acl.56} {Benchmarking robustness of machine reading comprehension models}.
\newblock In \emph{Findings of the Association for Computational Linguistics: ACL-IJCNLP 2021}, pages 634--644, Online. Association for Computational Linguistics.

\bibitem[{Sun et~al.(2022)Sun, Swayamdipta, May, and Ma}]{sun-etal-2022-investigating}
Jiao Sun, Swabha Swayamdipta, Jonathan May, and Xuezhe Ma. 2022.
\newblock \href {https://doi.org/10.18653/v1/2022.findings-emnlp.432} {Investigating the benefits of free-form rationales}.
\newblock In \emph{Findings of the Association for Computational Linguistics: EMNLP 2022}, pages 5867--5882, Abu Dhabi, United Arab Emirates. Association for Computational Linguistics.

\bibitem[{Tay et~al.(2023)Tay, Dehghani, Tran, Garcia, Wei, Wang, Chung, Bahri, Schuster, Zheng, Zhou, Houlsby, and Metzler}]{tay2023ul}
Yi~Tay, Mostafa Dehghani, Vinh~Q. Tran, Xavier Garcia, Jason Wei, Xuezhi Wang, Hyung~Won Chung, Dara Bahri, Tal Schuster, Steven Zheng, Denny Zhou, Neil Houlsby, and Donald Metzler. 2023.
\newblock \href {https://openreview.net/forum?id=6ruVLB727MC} {{UL}2: Unifying language learning paradigms}.
\newblock In \emph{The Eleventh International Conference on Learning Representations}.

\bibitem[{Toulmin(2003)}]{toulmin2003uses}
Stephen~E. Toulmin. 2003.
\newblock \emph{The uses of argument}.
\newblock Cambridge University Press.

\bibitem[{Touvron et~al.(2023{\natexlab{a}})Touvron, Lavril, Izacard, Martinet, Lachaux, Lacroix, Rozière, Goyal, Hambro, Azhar, Rodriguez, Joulin, Grave, and Lample}]{touvron2023llama}
Hugo Touvron, Thibaut Lavril, Gautier Izacard, Xavier Martinet, Marie-Anne Lachaux, Timothée Lacroix, Baptiste Rozière, Naman Goyal, Eric Hambro, Faisal Azhar, Aurelien Rodriguez, Armand Joulin, Edouard Grave, and Guillaume Lample. 2023{\natexlab{a}}.
\newblock \href {https://arxiv.org/abs/2302.13971} {{LLaMA}: Open and efficient foundation language models}.
\newblock {a}rXiv preprint 2302.13971.

\bibitem[{Touvron et~al.(2023{\natexlab{b}})Touvron, Martin, Stone, Albert, Almahairi, Babaei, Bashlykov, Batra, Bhargava, Bhosale, Bikel, Blecher, Ferrer, Chen, Cucurull, Esiobu, Fernandes, Fu, Fu, Fuller, Gao, Goswami, Goyal, Hartshorn, Hosseini, Hou, Inan, Kardas, Kerkez, Khabsa, Kloumann, Korenev, Koura, Lachaux, Lavril, Lee, Liskovich, Lu, Mao, Martinet, Mihaylov, Mishra, Molybog, Nie, Poulton, Reizenstein, Rungta, Saladi, Schelten, Silva, Smith, Subramanian, Tan, Tang, Taylor, Williams, Kuan, Xu, Yan, Zarov, Zhang, Fan, Kambadur, Narang, Rodriguez, Stojnic, Edunov, and Scialom}]{Touvron2023-zh}
Hugo Touvron, Louis Martin, Kevin Stone, Peter Albert, Amjad Almahairi, Yasmine Babaei, Nikolay Bashlykov, Soumya Batra, Prajjwal Bhargava, Shruti Bhosale, Dan Bikel, Lukas Blecher, Cristian~Canton Ferrer, Moya Chen, Guillem Cucurull, David Esiobu, Jude Fernandes, Jeremy Fu, Wenyin Fu, Brian Fuller, Cynthia Gao, Vedanuj Goswami, Naman Goyal, Anthony Hartshorn, Saghar Hosseini, Rui Hou, Hakan Inan, Marcin Kardas, Viktor Kerkez, Madian Khabsa, Isabel Kloumann, Artem Korenev, Punit~Singh Koura, Marie-Anne Lachaux, Thibaut Lavril, Jenya Lee, Diana Liskovich, Yinghai Lu, Yuning Mao, Xavier Martinet, Todor Mihaylov, Pushkar Mishra, Igor Molybog, Yixin Nie, Andrew Poulton, Jeremy Reizenstein, Rashi Rungta, Kalyan Saladi, Alan Schelten, Ruan Silva, Eric~Michael Smith, Ranjan Subramanian, Xiaoqing~Ellen Tan, Binh Tang, Ross Taylor, Adina Williams, Jian~Xiang Kuan, Puxin Xu, Zheng Yan, Iliyan Zarov, Yuchen Zhang, Angela Fan, Melanie Kambadur, Sharan Narang, Aurelien Rodriguez, Robert Stojnic, Sergey Edunov, and Thomas
  Scialom. 2023{\natexlab{b}}.
\newblock \href {https://arxiv.org/abs/2307.09288} {Llama 2: Open foundation and fine-tuned chat models}.
\newblock {a}rXiv preprint 2307.09288.

\bibitem[{Turpin et~al.(2023)Turpin, Michael, Perez, and Bowman}]{turpin2023language}
Miles Turpin, Julian Michael, Ethan Perez, and Samuel~R. Bowman. 2023.
\newblock \href {https://arxiv.org/abs/2305.04388} {Language models don't always say what they think: Unfaithful explanations in chain-of-thought prompting}.
\newblock {a}rXiv preprint 2305.04388.

\bibitem[{Wang et~al.(2023)Wang, Min, Deng, Shen, Wu, Zettlemoyer, and Sun}]{wang-etal-2023-towards}
Boshi Wang, Sewon Min, Xiang Deng, Jiaming Shen, You Wu, Luke Zettlemoyer, and Huan Sun. 2023.
\newblock \href {https://doi.org/10.18653/v1/2023.acl-long.153} {Towards understanding chain-of-thought prompting: An empirical study of what matters}.
\newblock In \emph{Proceedings of the 61st Annual Meeting of the Association for Computational Linguistics (Volume 1: Long Papers)}, pages 2717--2739, Toronto, Canada. Association for Computational Linguistics.

\bibitem[{Wang et~al.(2022)Wang, Zhong, Tang, Wei, Fan, Jiang, Zhou, and Duan}]{wang-etal-2022-logic}
Siyuan Wang, Wanjun Zhong, Duyu Tang, Zhongyu Wei, Zhihao Fan, Daxin Jiang, Ming Zhou, and Nan Duan. 2022.
\newblock \href {https://doi.org/10.18653/v1/2022.findings-acl.127} {Logic-driven context extension and data augmentation for logical reasoning of text}.
\newblock In \emph{Findings of the Association for Computational Linguistics: ACL 2022}, pages 1619--1629, Dublin, Ireland. Association for Computational Linguistics.

\bibitem[{Wei et~al.(2022)Wei, Wang, Schuurmans, Bosma, ichter, Xia, Chi, Le, and Zhou}]{wei2022chain}
Jason Wei, Xuezhi Wang, Dale Schuurmans, Maarten Bosma, brian ichter, Fei Xia, Ed~Chi, Quoc~V Le, and Denny Zhou. 2022.
\newblock \href {https://proceedings.neurips.cc/paper_files/paper/2022/file/9d5609613524ecf4f15af0f7b31abca4-Paper-Conference.pdf} {Chain-of-thought prompting elicits reasoning in large language models}.
\newblock In \emph{Advances in Neural Information Processing Systems}, volume~35, pages 24824--24837. Curran Associates, Inc.

\bibitem[{Yang et~al.(2018)Yang, Qi, Zhang, Bengio, Cohen, Salakhutdinov, and Manning}]{yang-etal-2018-hotpotqa}
Zhilin Yang, Peng Qi, Saizheng Zhang, Yoshua Bengio, William Cohen, Ruslan Salakhutdinov, and Christopher~D. Manning. 2018.
\newblock \href {https://doi.org/10.18653/v1/D18-1259} {{H}otpot{QA}: A dataset for diverse, explainable multi-hop question answering}.
\newblock In \emph{Proceedings of the 2018 Conference on Empirical Methods in Natural Language Processing}, pages 2369--2380, Brussels, Belgium. Association for Computational Linguistics.

\bibitem[{Yu et~al.(2020)Yu, Jiang, Dong, and Feng}]{yu-etal-2020-reclor}
Weihao Yu, Zihang Jiang, Yanfei Dong, and Jiashi Feng. 2020.
\newblock \href {https://openreview.net/forum?id=HJgJtT4tvB} {{R}e{C}lor: A reading comprehension dataset requiring logical reasoning}.
\newblock In \emph{International Conference on Learning Representations}.

\bibitem[{Zhong et~al.(2022)Zhong, Wang, Tang, Xu, Guo, Chen, Wang, Yin, Zhou, and Duan}]{zhong-etal-2022-analytical}
Wanjun Zhong, Siyuan Wang, Duyu Tang, Zenan Xu, Daya Guo, Yining Chen, Jiahai Wang, Jian Yin, Ming Zhou, and Nan Duan. 2022.
\newblock \href {https://doi.org/10.18653/v1/2022.findings-naacl.177} {Analytical reasoning of text}.
\newblock In \emph{Findings of the Association for Computational Linguistics: NAACL 2022}, pages 2306--2319, Seattle, United States. Association for Computational Linguistics.

\end{thebibliography}
\bibliographystyle{acl_natbib}

\appendix
\section{Crowdsourcing Instructions and Examples}
\label{app:instructions}

\begin{figure*}[!t]
    \centering
    \includegraphics[width=\textwidth]{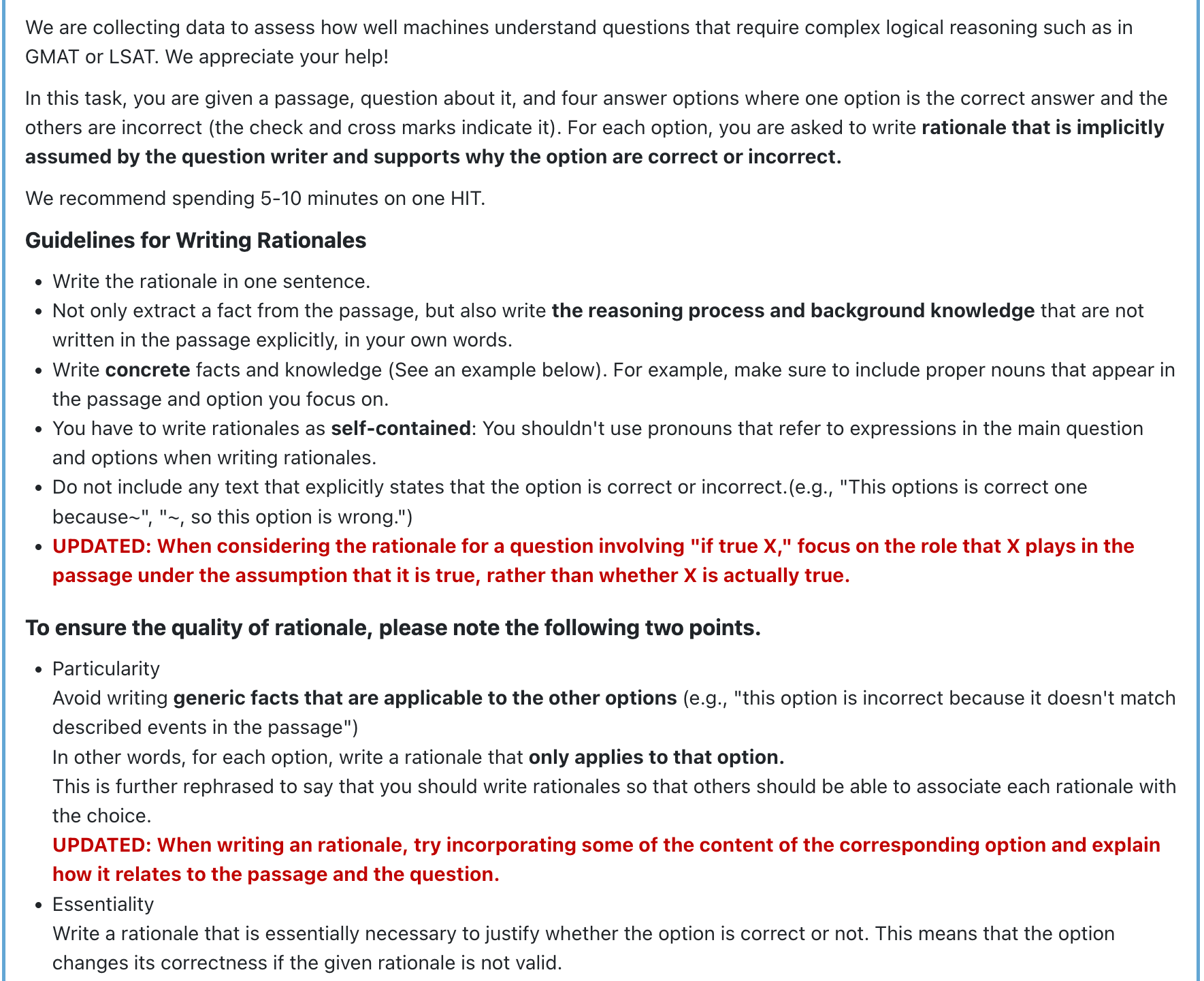}
    \caption{Instructions for the rationale writing task (1/4).}
    \label{fig:rationale-writing-interface1}
\end{figure*}

\begin{figure*}[!t]
    \centering
    \includegraphics[width=\textwidth]{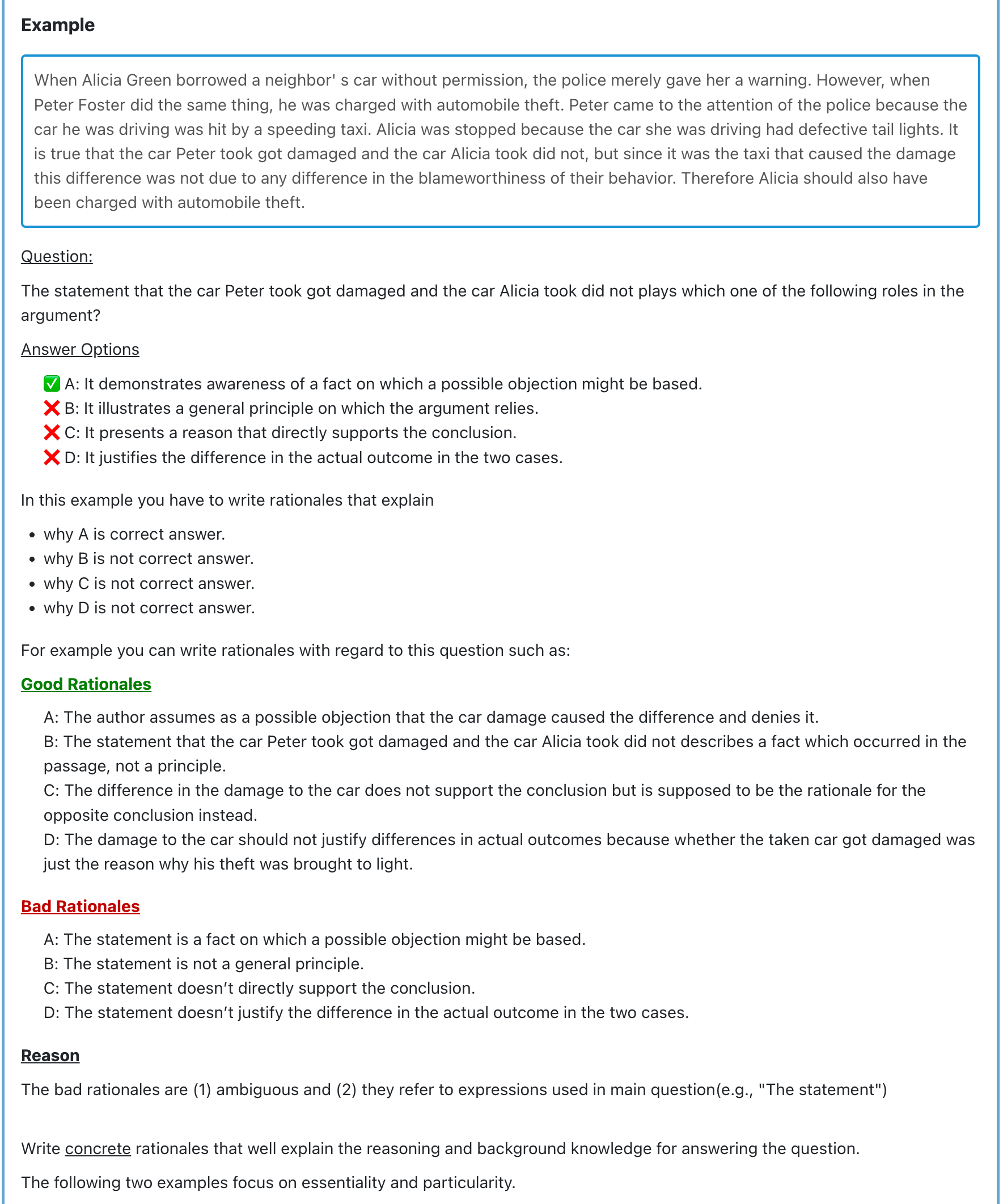}
    \caption{Instructions for the rationale writing task (2/4).}
    \label{fig:rationale-writing-interface2}
\end{figure*}

\begin{figure*}[!t]
    \centering
    \includegraphics[width=\textwidth]{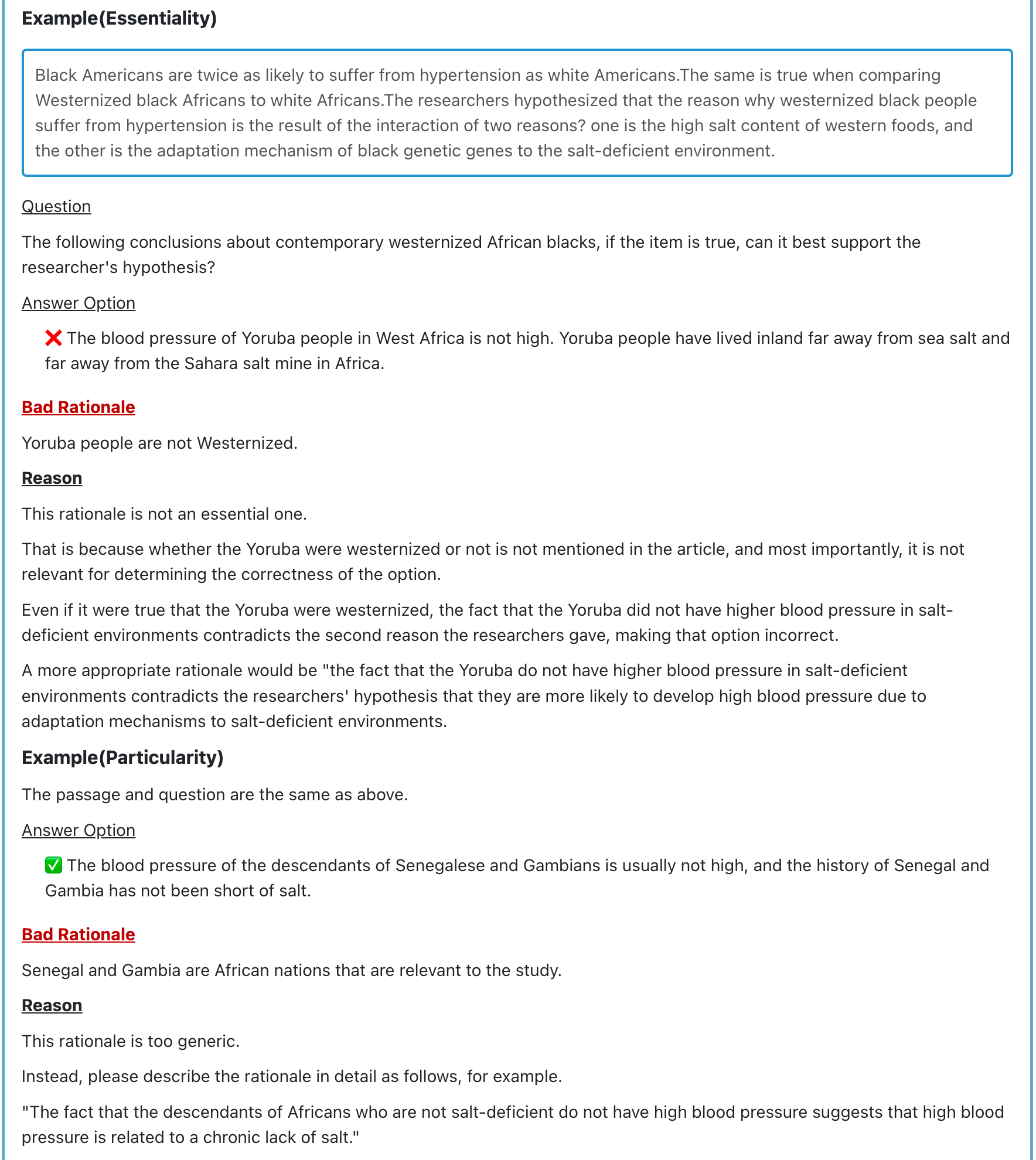}
    \caption{Instructions for the rationale writing task (3/4).}
    \label{fig:rationale-writing-interface3}
\end{figure*}

\begin{figure*}[!t]
    \centering
    \includegraphics[width=\textwidth]{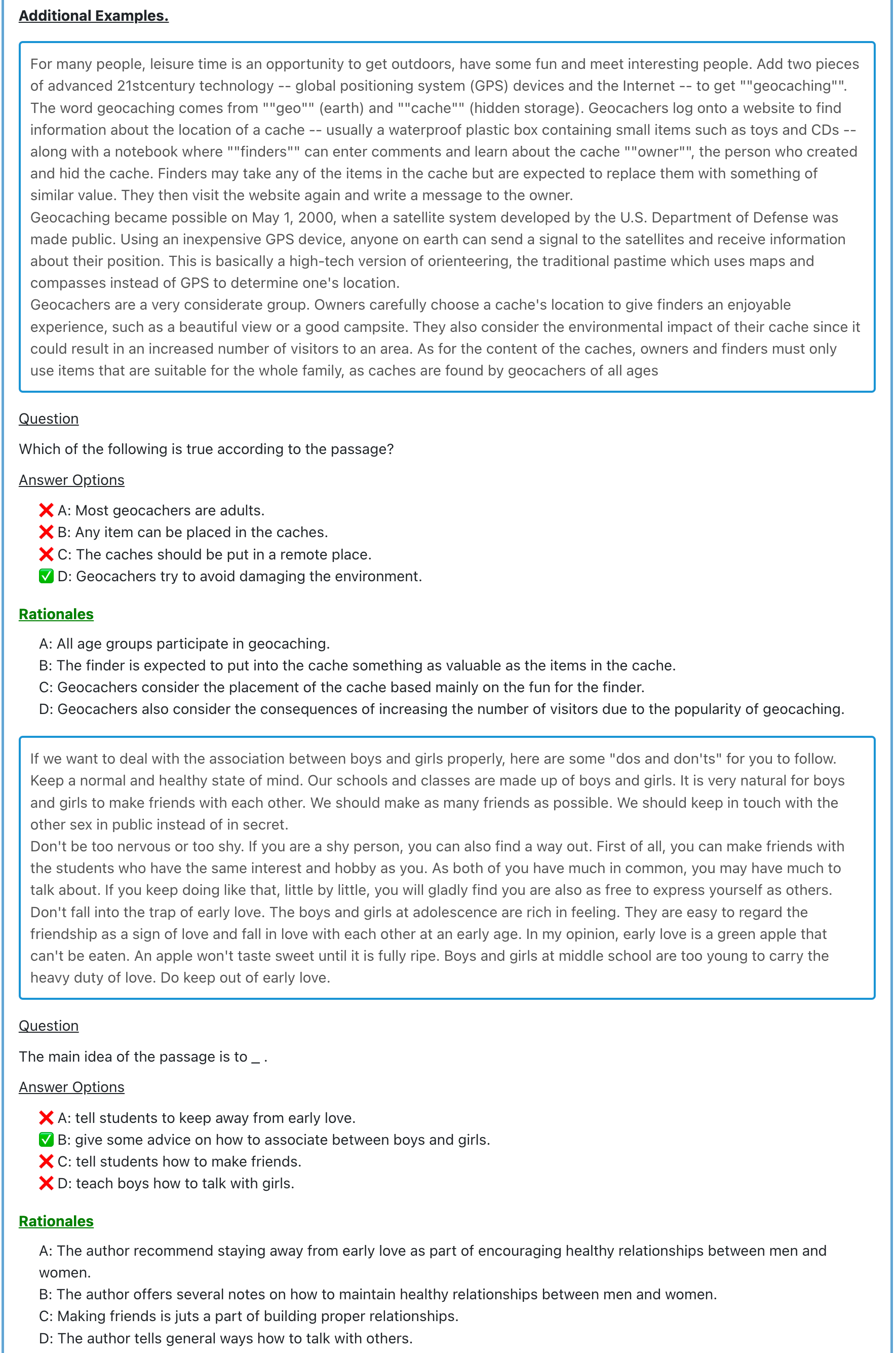}
    \caption{Instructions for the rationale writing task (4/4).}
    \label{fig:rationale-writing-interface4}
\end{figure*}

\begin{figure*}[!t]
    \centering
    \includegraphics[scale=0.22]{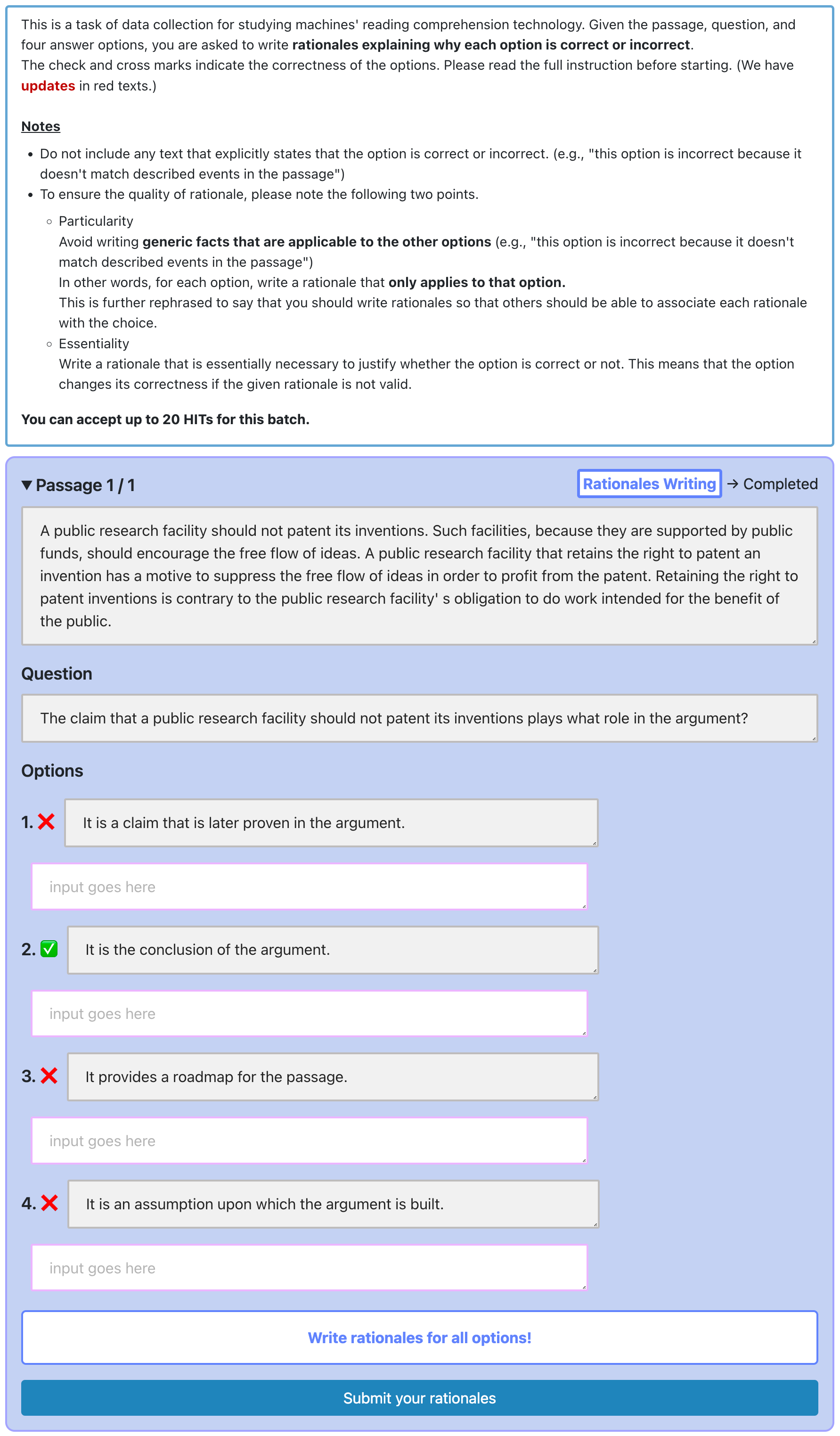}
    \caption{Rationale writing task interface. The workers are given a context, question, and four options along with their correctness, and are asked to provide a rationale for each choice.}
    \label{fig:rationale-writing-interface5}
\end{figure*}

\begin{figure*}[!t]
    \centering
    \includegraphics[width=\textwidth]{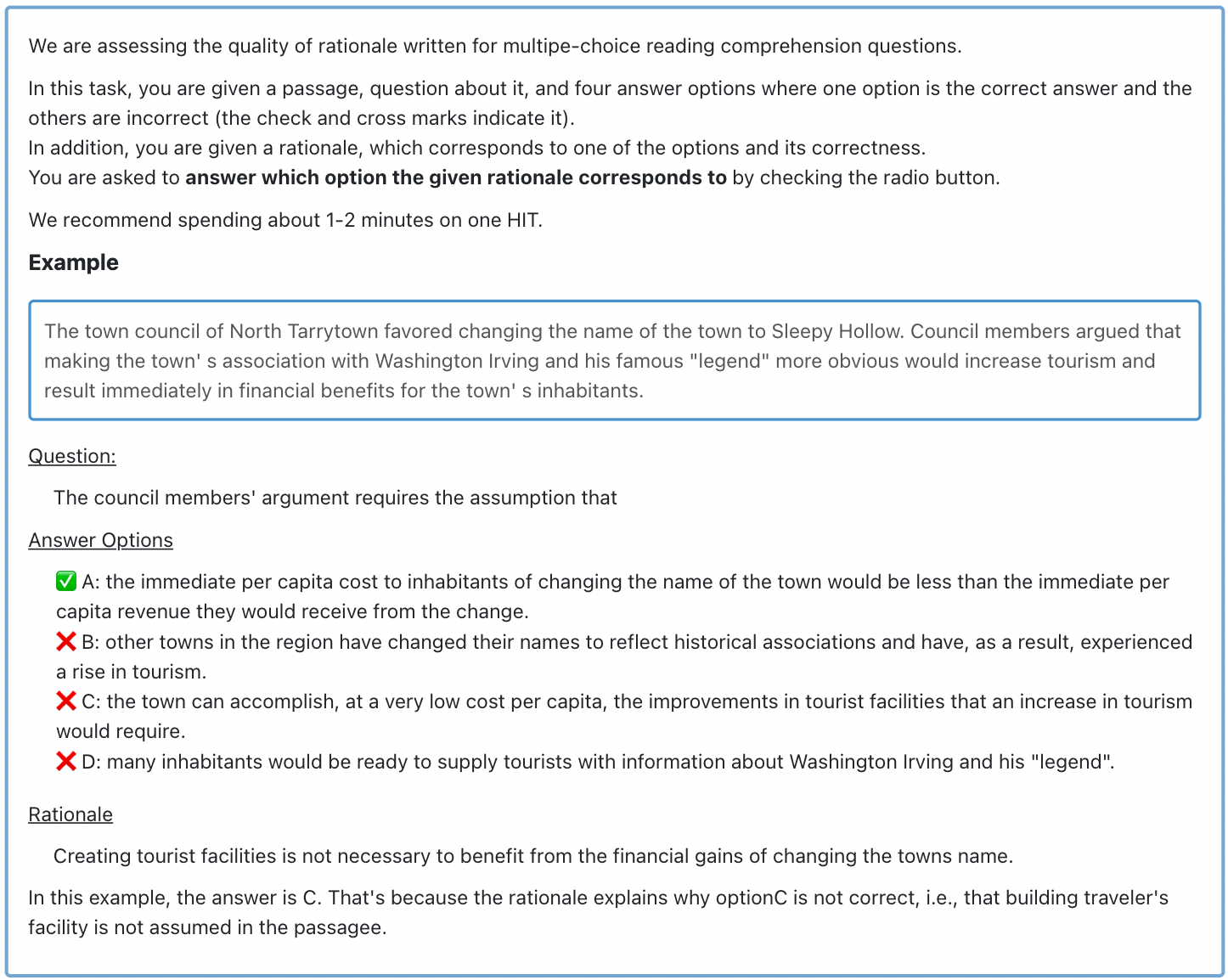}
    \caption{Instructions for the rationale validation task.}
    \label{fig:rationale-validation-interface1}
\end{figure*}

\begin{figure*}[!t]
    \centering
    \includegraphics[width=\textwidth]{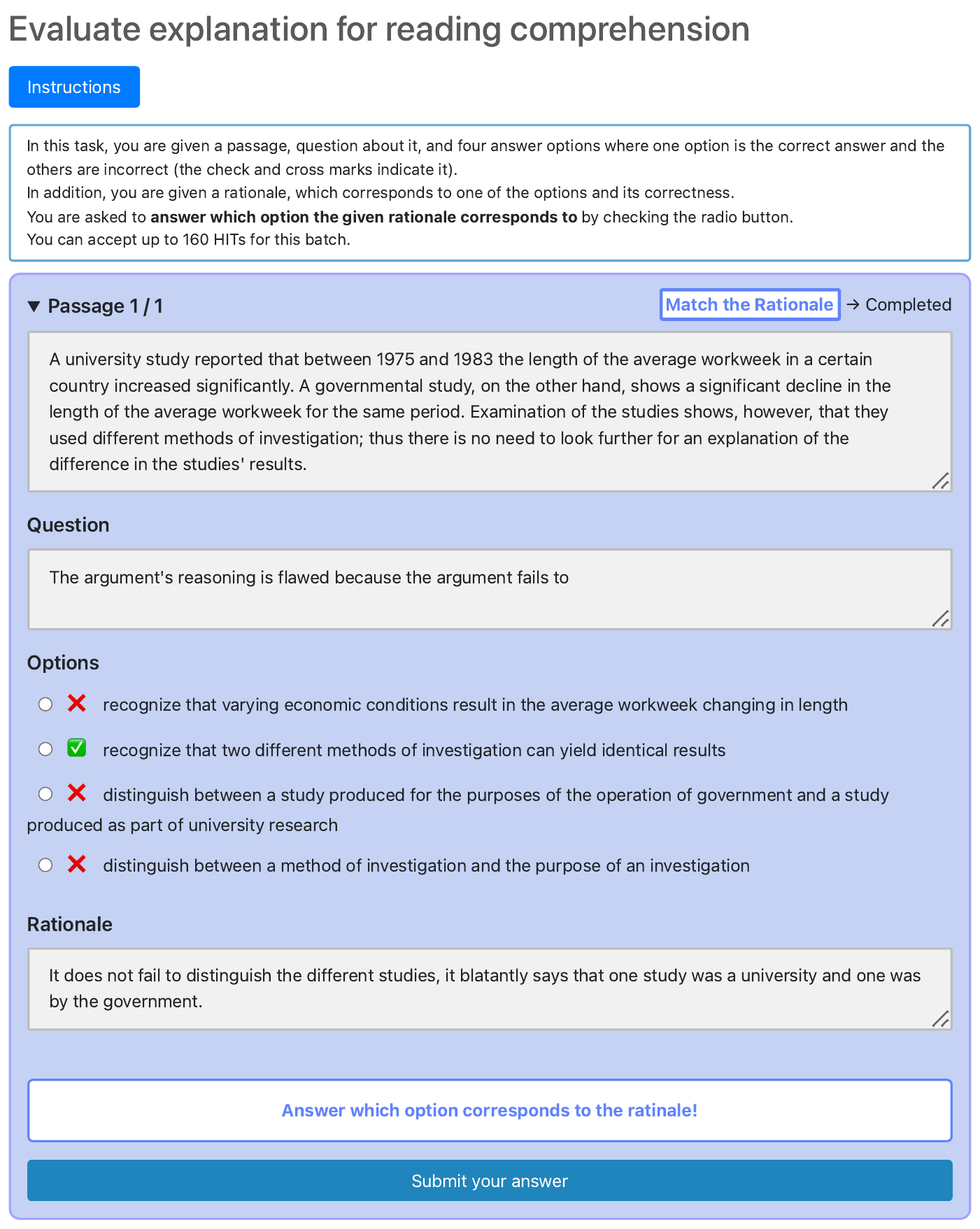}
    \caption{Rationale validation task interface. The workers select the option best supported by the provided rationale.}
    \label{fig:rationale-validation-interface2}
\end{figure*}

\begin{figure*}[!t]
    \centering
    \includegraphics[width=\textwidth]{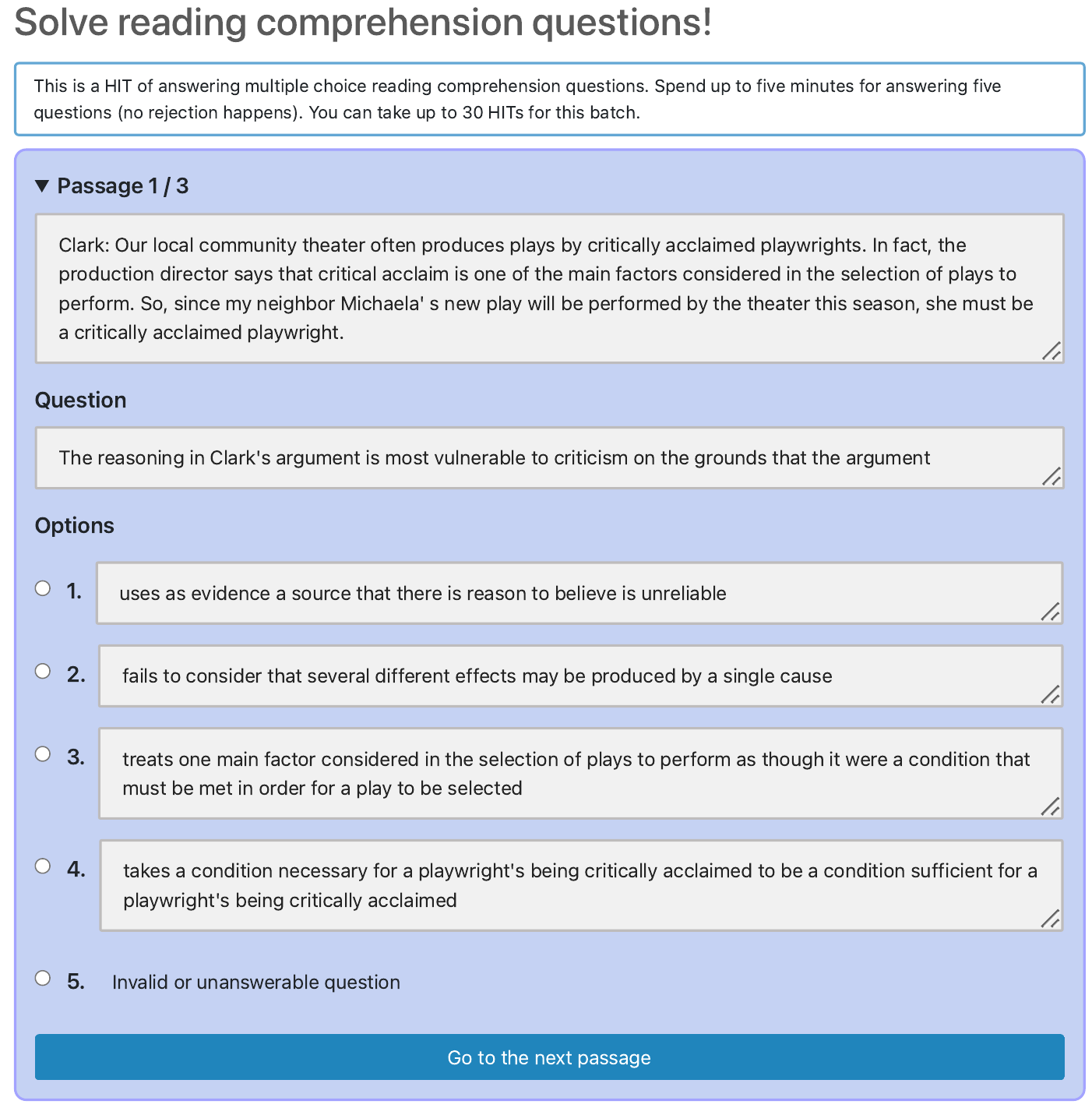}
    \caption{Human validation task interface. The workers are asked to answer subquestions.}
    \label{fig:human-validation-interface}
\end{figure*}

Figures~\ref{fig:rationale-writing-interface1} to ~\ref{fig:human-validation-interface} show the instructions and examples we present to the crowdworkers.
Figures~\ref{fig:rationale-writing-interface1} to ~\ref{fig:rationale-writing-interface5} illustrate the rationale writing task, Figures~\ref{fig:rationale-validation-interface1} and~\ref{fig:rationale-validation-interface2} illustrate the rationale validation task, and Figure~\ref{fig:human-validation-interface} illustrates the human validation task.   %in our rationale writing tasks.

\section{Question Generation Prompt}
\label{app:prompt-qg}

\begin{figure*}[!t]
    \centering
    \includegraphics[width=\textwidth]{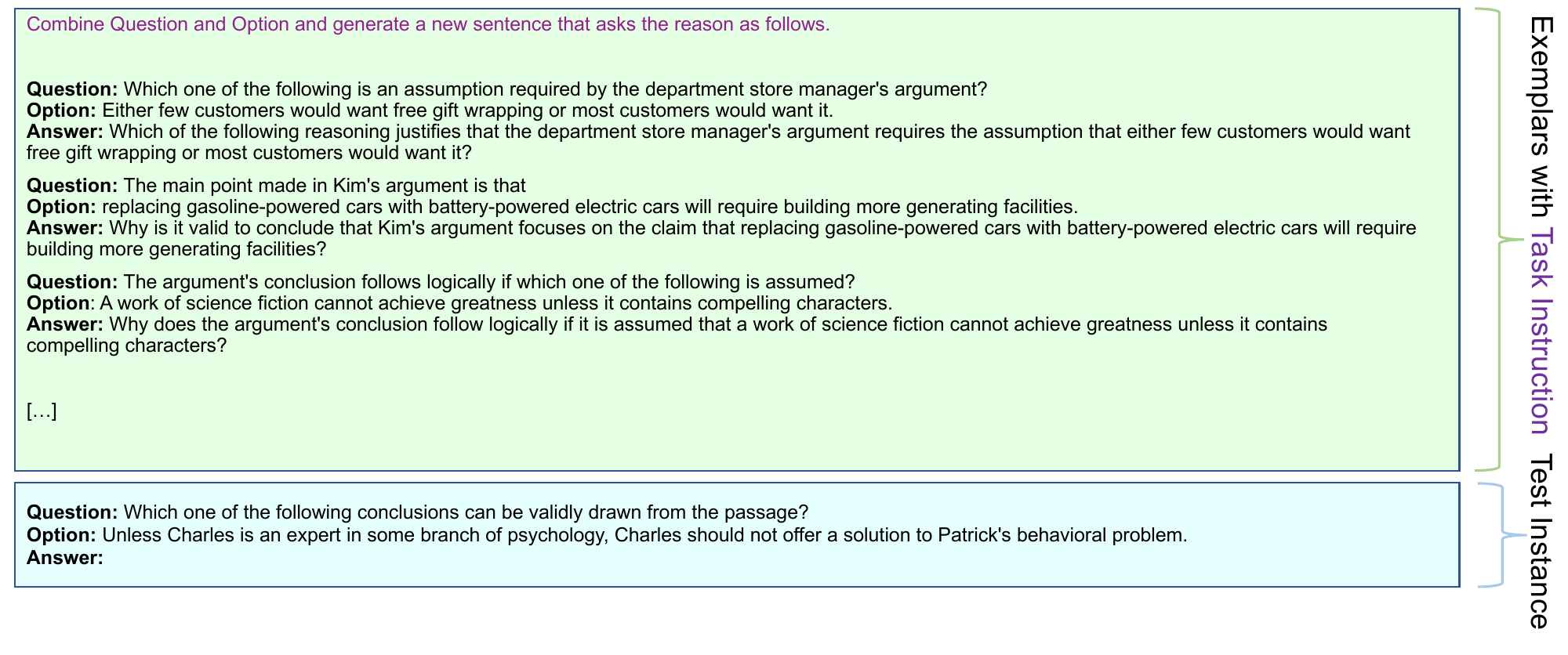}
    \caption{Example of the prompt used to generate subquestions.}
    \label{fig:prompt-qg}
\end{figure*}

Figure~\ref{fig:prompt-qg} shows an example of our prompt used for generating subquestions in Section~\ref{sec:subquestion-generation}.

\section{Crowdsourcing Details}
\label{app:crowdsourcing}

To access a pool of crowdworkers, we used Amazon Mechanical Turk.
The crowdworkers who took the qualification test are based in the United States, United Kingdom, or Canada, have an approval rate of at least 98\%, and have at least 1,000 approved tasks.
We ensure that the average payments exceed \$12.00 USD per hour for each task. The rationale writing task costs \$2.00 per main question (estimating that it takes seven to ten minutes to write the rationales), the rationale validation task costs \$0.30 per rationale (one minute), and the human validation task \$1.50 per five questions (five minutes).
The rationale writing tasks, rationale validation tasks, QA validation tasks, and human performance tasks are taken by 48, 39, 52, and 24 workers, respectively.
We use the crowdsourcing tool used in \citet{nangia-etal-2021-ingredients}.
%For the rationale writing tasks, 48 workers write rationales for 100.0 main questions on average. % (min: 4.0, max: 316.0).
% For the rationale validation tasks, 39 workers annotate 123.5 rationales on average.
% For the human validation task, 52 workers took tasks.
% For the question-answering task to measure human performance, 24 workers took tasks.

\section{Chain-of-Thought Prompt}
\label{app:prompt-cot}

\begin{figure*}[!t]
    \centering
    \includegraphics[width=\textwidth]{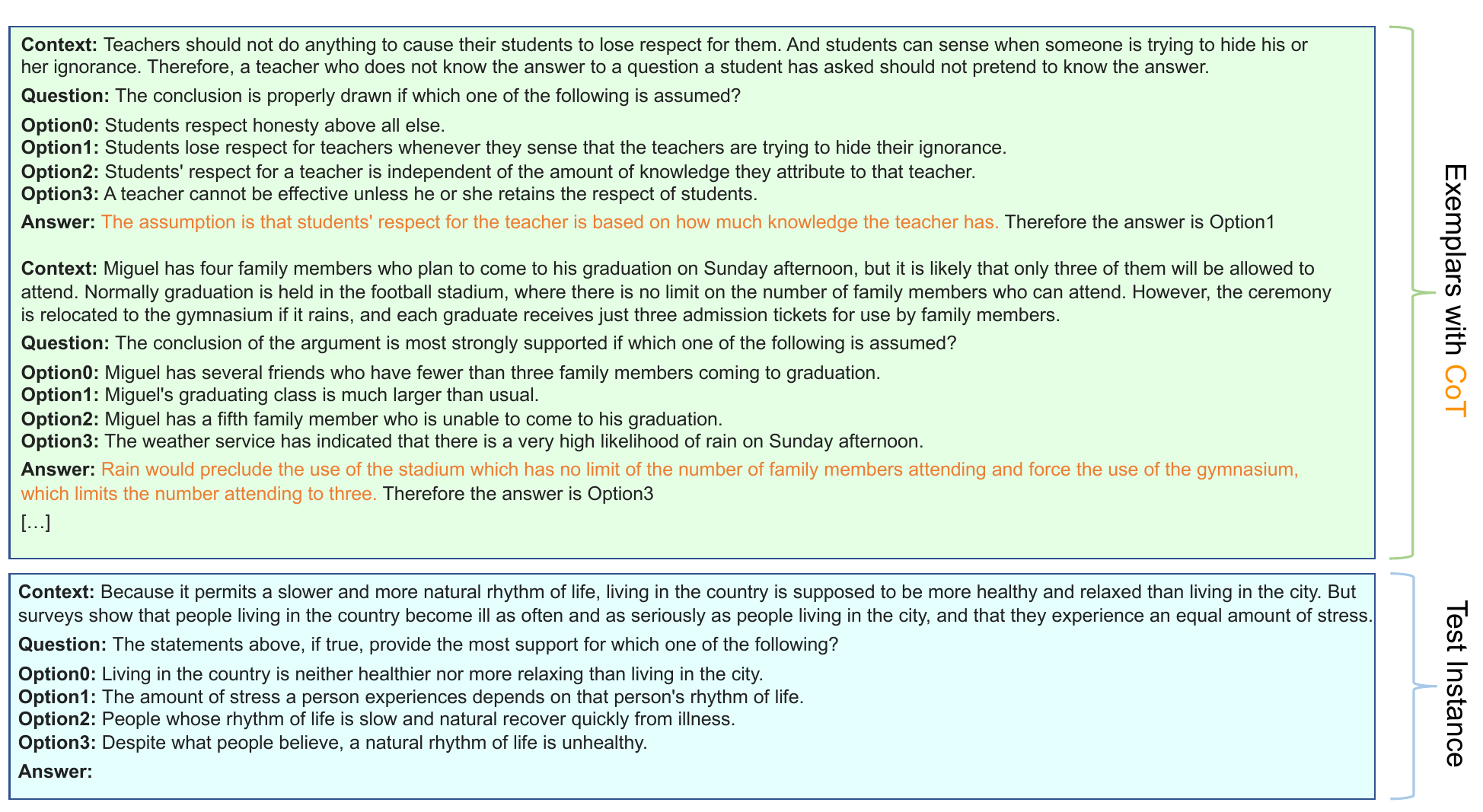}
    \caption{Example of the prompt using the chain-of-thought approach.}
    \label{fig:prompt-cot}
\end{figure*}

Figure~\ref{fig:prompt-cot} shows an example of the prompt used in our chain-of-thought experiment.
We insert the rationale between the \texttt{Answer:} label and the correct option label, with an expectation that it would help the model (InstructGPT) select the correct option.

\section{Test Split Setting}
\label{app:test-split-setting}

% \paragraph{Test Set Split for In-Context Learning}
The in-context learning performance of LLMs may vary depending on the exemplars of the prompt, but it incurs a high computational cost (or financial cost for proprietary models) if we repeatedly evaluate the models on the entire dataset using various sets of different exemplars to take the average performance.
Because of this cost limitation, we follow the test split approach used by \citet{ravichander-etal-2022-condaqa}, splitting our dataset into five disjoint sets and testing the models on each set with different exemplars to measure the performance variance across the disjoint sets.
Note that we do not split the set of the main questions, because it has only 943 examples; hence, in the few-shot setting, we take the average across five runs on all main questions.
In the few-shot setting using ReClor, we sample questions disjointly from its training set, whereas in using RULE, the exemplars are sampled from the corresponding disjoint set.

\section{MainQ-wise SubQ Results of InstructGPT}
\label{app:subq-result-distribution}

\begin{figure}[!t]
    \centering
    \begin{minipage}{0.45\textwidth}
    \includegraphics[width=\linewidth]{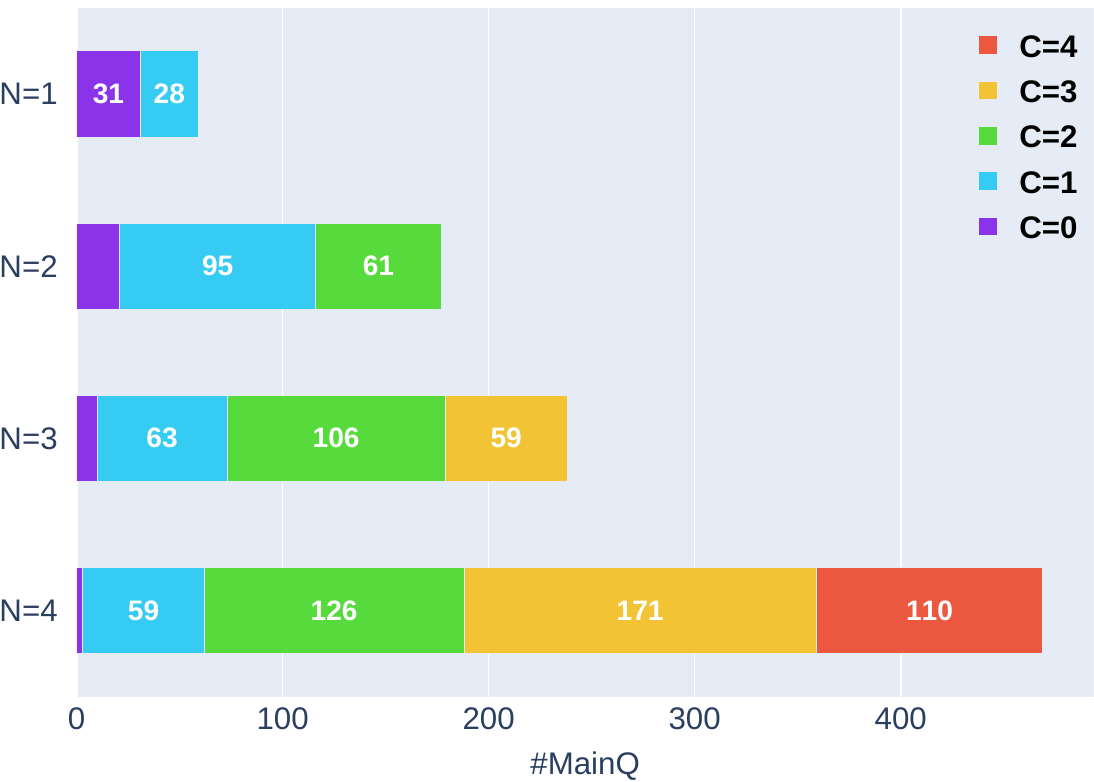}
    \end{minipage}\hfill
    \begin{minipage}{0.45\textwidth}
    \includegraphics[width=\linewidth]{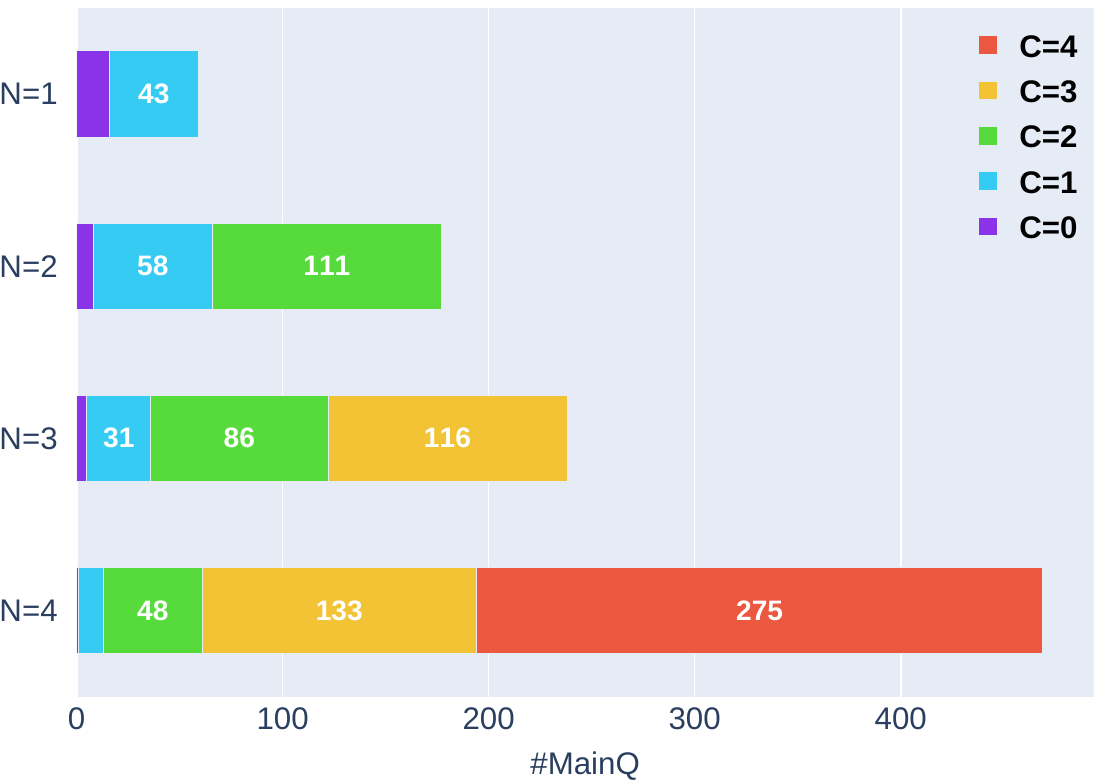}
    \end{minipage}
    \caption{Distribution of correctly answered subquestions ($C$) out of the total number of subquestions ($N$), for both InstructGPT (top) and humans (bottom).}
    \label{fig:subq-result-distribution-gpt3}
\end{figure}

Because a single main question has multiple subquestions in our dataset, we report the detailed numbers of correctly-answered SubQ by InstructGPT in Figure~\ref{fig:subq-result-distribution-gpt3}.

\begin{table*}[!h]
\centering \small
\begin{tabular}{lcrrrrrr} \toprule
\textbf{Model} & \textbf{\# Param} & \makecell[c]{\textbf{MainQ}\\\textbf{Acc.}} & \makecell[c]{\textbf{SubQ}\\\textbf{Acc.}} & \makecell[c]{\textbf{Selective}\\\textbf{SubQ Acc.}} & \makecell[c]{\textbf{Eliminative}\\\textbf{SubQ Acc.}} & \textbf{Consist.} &  \makecell[r]{\textbf{MainQ-wise}\\\textbf{Acc.}} \\ \midrule
% \textsc{DeBERTa-v3-Large} & 304M & 66.0\hphantom{$_{\pm0.00}$} & 33.1\hphantom{$_{\pm0.00}$} & 56.1\hphantom{$_{\pm0.00}$} & 25.0\hphantom{$_{\pm0.00}$} & 2.4 & 32.8\\
% \textsc{UnifiedQA-v2-Base} & 220M & 40.5\hphantom{$_{\pm0.00}$} & 25.8\hphantom{$_{\pm0.00}$} & 21.3\hphantom{$_{\pm0.00}$} & 27.4\hphantom{$_{\pm0.00}$} & 0.7 & 26.0\\
% \textsc{UnifiedQA-v2-Large} & 770M & 57.7\hphantom{$_{\pm0.00}$} & 25.0\hphantom{$_{\pm0.00}$} & 19.9\hphantom{$_{\pm0.00}$} & 26.8\hphantom{$_{\pm0.00}$} & 1.4 & 24.7\\
% \textsc{UnifiedQA-v2-3B} & 3B & 66.8\hphantom{$_{\pm0.00}$} & 25.3\hphantom{$_{\pm0.00}$} & 21.8\hphantom{$_{\pm0.00}$} & 26.6\hphantom{$_{\pm0.00}$} & 1.4 & 25.2\\
\multicolumn{8}{c}{\cellcolor{light-gray} \emph{Five-Shot on ReClor}} \\
% \textsc{Flan-UL2} & 20B & 58.5$_{\pm0.3\hphantom{0}}$ & 65.5$_{\pm5.1\hphantom{0}}$ & 88.0$_{\pm4.0\hphantom{0}}$ & 57.6$_{\pm5.4\hphantom{0}}$ & 16.9 & 64.3\\
\textsc{LLaMA 7B} & 7B & 25.8$_{\pm1.6\hphantom{0}}$ & 28.6$_{\pm7.1\hphantom{0}}$ & 39.8$_{\pm13.1}$ & 24.6$_{\pm6.3\hphantom{0}}$ & 0.8 & 28.2\\
\textsc{LLaMA 13B} & 13B & 38.7$_{\pm2.7\hphantom{0}}$ & 36.3$_{\pm3.5\hphantom{0}}$ & 63.6$_{\pm4.0\hphantom{0}}$ & 26.6$_{\pm3.6\hphantom{0}}$ & 2.9 & 36.6\\
% \textsc{LLaMA 33B} & 33B & 58.5$_{\pm1.2\hphantom{0}}$ & 48.1$_{\pm3.5\hphantom{0}}$ & 77.3$_{\pm3.6\hphantom{0}}$ & 37.7$_{\pm3.6\hphantom{0}}$ & 6.2 & 48.0\\
% \textsc{LLaMA 65B} & 65B & 69.1$_{\pm0.9\hphantom{0}}$ & 55.3$_{\pm2.7\hphantom{0}}$ & 85.0$_{\pm1.4\hphantom{0}}$ & 44.8$_{\pm2.5\hphantom{0}}$ & 11.2 & 54.9\\
\textsc{Vicuna 7B} & 7B & 33.4$_{\pm2.6\hphantom{0}}$ & 38.6$_{\pm3.3\hphantom{0}}$ & 61.4$_{\pm5.8\hphantom{0}}$ & 30.5$_{\pm3.4\hphantom{0}}$ & 2.8 & 38.3\\
\textsc{LLaMA2 7B} & 7B & 36.4$_{\pm2.1\hphantom{0}}$ & 35.2$_{\pm3.9\hphantom{0}}$ & 63.6$_{\pm6.5\hphantom{0}}$ & 25.1$_{\pm3.2\hphantom{0}}$ & 1.7 & 35.6\\
% \textsc{Vicuna 13B} & 13B & 46.2$_{\pm0.7\hphantom{0}}$ & 50.0$_{\pm4.4\hphantom{0}}$ & 78.2$_{\pm3.0\hphantom{0}}$ & 40.1$_{\pm4.6\hphantom{0}}$ & 5.6 & 49.4\\
% \textsc{InstructGPT} & N/A & 71.8$_{\pm1.0\hphantom{0}}$ & 65.3$_{\pm1.8\hphantom{0}}$ & 88.4$_{\pm2.5\hphantom{0}}$ & 57.1$_{\pm1.5\hphantom{0}}$ & 18.2 & 64.0\\
% \textsc{InstructGPT + CoT} & N/A & 67.8$_{\pm0.5\hphantom{0}}$ & 63.2$_{\pm2.1\hphantom{0}}$ & 88.5$_{\pm2.5\hphantom{0}}$ & 54.2$_{\pm2.8\hphantom{0}}$ & 17.2 & 61.8\\
\multicolumn{8}{c}{\cellcolor{light-gray} \emph{One-Shot on ReClor}} \\
\textsc{UnifiedQA-v2-Base} & 220M & 27.4$_{\pm1.4\hphantom{0}}$ & 34.7$_{\pm6.4\hphantom{0}}$ & 42.8$_{\pm7.7\hphantom{0}}$ & 31.9$_{\pm6.2\hphantom{0}}$ & 0.7 & 34.6\\
\textsc{UnifiedQA-v2-Large} & 770M & 27.1$_{\pm1.7\hphantom{0}}$ & 28.0$_{\pm4.7\hphantom{0}}$ & 29.2$_{\pm5.3\hphantom{0}}$ & 27.6$_{\pm4.9\hphantom{0}}$ & 0.0 & 27.6\\
\textsc{UnifiedQA-v2-3B} & 3B & 31.3$_{\pm3.2\hphantom{0}}$ & 26.2$_{\pm2.8\hphantom{0}}$ & 26.0$_{\pm9.8\hphantom{0}}$ & 26.2$_{\pm1.4\hphantom{0}}$ & 0.3 & 26.1\\
\textsc{UnifiedQA-v2-11B} & 11B & 44.1$_{\pm5.9\hphantom{0}}$ & 37.2$_{\pm7.4\hphantom{0}}$ & 53.8$_{\pm16.6}$ & 31.4$_{\pm4.4\hphantom{0}}$ & 2.0 & 36.5\\
\textsc{Flan-T5-XXL} & 11B & 61.3$_{\pm0.3\hphantom{0}}$ & 63.7$_{\pm3.7\hphantom{0}}$ & 85.9$_{\pm2.6\hphantom{0}}$ & 55.9$_{\pm3.9\hphantom{0}}$ & 14.0 & 62.5\\
\textsc{Flan-UL2} & 20B & 58.0$_{\pm0.5\hphantom{0}}$ & 66.0$_{\pm5.3\hphantom{0}}$ & 87.4$_{\pm3.9\hphantom{0}}$ & 58.5$_{\pm5.7\hphantom{0}}$ & 17.7 & 65.1\\
\textsc{LLaMA 7B} & 7B & 26.6$_{\pm0.9\hphantom{0}}$ & 32.4$_{\pm4.2\hphantom{0}}$ & 50.7$_{\pm11.1}$ & 26.1$_{\pm3.7\hphantom{0}}$ & 1.6 & 32.5\\
\textsc{LLaMA 13B} & 13B & 32.7$_{\pm2.4\hphantom{0}}$ & 33.8$_{\pm1.9\hphantom{0}}$ & 56.0$_{\pm5.8\hphantom{0}}$ & 26.0$_{\pm2.5\hphantom{0}}$ & 1.4 & 34.1\\
\textsc{LLaMA 33B} & 33B & 56.1$_{\pm1.0\hphantom{0}}$ & 49.3$_{\pm4.3\hphantom{0}}$ & 80.1$_{\pm3.8\hphantom{0}}$ & 38.5$_{\pm4.3\hphantom{0}}$ & 7.4 & 49.6\\
\textsc{LLaMA 65B} & 65B & 65.2$_{\pm1.4\hphantom{0}}$ & 52.6$_{\pm3.4\hphantom{0}}$ & 85.2$_{\pm1.3\hphantom{0}}$ & 41.1$_{\pm4.3\hphantom{0}}$ & 9.4 & 51.9\\
\textsc{Vicuna 7B} & 7B & 35.8$_{\pm2.0\hphantom{0}}$ & 38.9$_{\pm2.2\hphantom{0}}$ & 65.3$_{\pm3.1\hphantom{0}}$ & 29.5$_{\pm2.3\hphantom{0}}$ & 1.8 & 38.1\\
\textsc{Vicuna 13B} & 13B & 42.5$_{\pm0.8\hphantom{0}}$ & 45.2$_{\pm3.1\hphantom{0}}$ & 72.1$_{\pm3.3\hphantom{0}}$ & 35.8$_{\pm4.0\hphantom{0}}$ & 4.2 & 45.1\\
\textsc{InstructGPT} & N/A & 67.8$_{\pm0.5\hphantom{0}}$ & 64.6$_{\pm1.9\hphantom{0}}$ & 87.8$_{\pm2.0\hphantom{0}}$ & 56.3$_{\pm1.8\hphantom{0}}$ & 17.5 & 63.6\\
\textsc{InstructGPT + CoT} & N/A & 64.3$_{\pm2.5\hphantom{0}}$ & 64.3$_{\pm2.4\hphantom{0}}$ & 88.8$_{\pm1.3\hphantom{0}}$ & 55.7$_{\pm2.5\hphantom{0}}$ & 15.4 & 62.5\\
\textsc{LLaMA2 7B} & 7B & 35.0$_{\pm1.1\hphantom{0}}$ & 34.8$_{\pm2.4\hphantom{0}}$ & 61.7$_{\pm4.2\hphantom{0}}$ & 25.2$_{\pm2.6\hphantom{0}}$ & 2.2 & 35.2\\
\textsc{LLaMA2 13B} & 13B & 46.4$_{\pm2.4\hphantom{0}}$ & 43.7$_{\pm2.9\hphantom{0}}$ & 72.6$_{\pm3.6\hphantom{0}}$ & 33.4$_{\pm2.8\hphantom{0}}$ & 4.7 & 43.6\\
\textsc{LLaMA2 70B} & 70B & 77.2$_{\pm0.2\hphantom{0}}$ & 61.3$_{\pm0.7\hphantom{0}}$ & 90.0$_{\pm1.4\hphantom{0}}$ & 51.2$_{\pm1.2\hphantom{0}}$ & 20.0 & 61.0\\
\textsc{Mistral 7B} & 7B & 52.6$_{\pm1.4\hphantom{0}}$ & 53.4$_{\pm2.3\hphantom{0}}$ & 81.9$_{\pm2.8\hphantom{0}}$ & 43.4$_{\pm2.7\hphantom{0}}$ & 7.4 & 52.7\\
\multicolumn{8}{c}{\cellcolor{light-gray} \emph{Five-Shot on RULE (for reference)}} \\
% \textsc{Flan-UL2} & 20B & 57.9$_{\pm0.2\hphantom{0}}$ & 66.0$_{\pm4.9\hphantom{0}}$ & 87.7$_{\pm4.6\hphantom{0}}$ & 58.4$_{\pm5.0\hphantom{0}}$ & 17.8 & 64.9\\
\textsc{LLaMA 7B} & 7B & 29.1$_{\pm2.3\hphantom{0}}$ & 34.9$_{\pm2.5\hphantom{0}}$ & 64.3$_{\pm4.3\hphantom{0}}$ & 24.5$_{\pm2.4\hphantom{0}}$ & 1.5 & 35.5\\
\textsc{LLaMA 13B} & 13B & 36.8$_{\pm3.5\hphantom{0}}$ & 35.6$_{\pm2.5\hphantom{0}}$ & 68.1$_{\pm3.0\hphantom{0}}$ & 24.2$_{\pm2.8\hphantom{0}}$ & 2.4 & 36.0\\
% \textsc{LLaMA 33B} & 33B & 53.6$_{\pm0.4\hphantom{0}}$ & 47.8$_{\pm4.5\hphantom{0}}$ & 77.8$_{\pm3.9\hphantom{0}}$ & 37.2$_{\pm4.8\hphantom{0}}$ & 5.7 & 47.6\\
% \textsc{LLaMA 65B} & 65B & 66.2$_{\pm0.7\hphantom{0}}$ & 58.9$_{\pm6.3\hphantom{0}}$ & 86.5$_{\pm1.7\hphantom{0}}$ & 49.1$_{\pm7.6\hphantom{0}}$ & 12.0 & 58.0\\
\textsc{Vicuna 7B} & 7B & 35.0$_{\pm1.1\hphantom{0}}$ & 39.9$_{\pm3.9\hphantom{0}}$ & 60.2$_{\pm8.9\hphantom{0}}$ & 32.7$_{\pm4.4\hphantom{0}}$ & 3.2 & 39.8\\
\textsc{LLaMA2 7B} & 7B & 37.8$_{\pm1.1\hphantom{0}}$ & 32.0$_{\pm4.4\hphantom{0}}$ & 62.3$_{\pm6.5\hphantom{0}}$ & 21.1$_{\pm3.4\hphantom{0}}$ & 1.5 & 32.4\\
% \textsc{Vicuna 13B} & 13B & 43.9$_{\pm1.3\hphantom{0}}$ & 44.2$_{\pm2.7\hphantom{0}}$ & 72.6$_{\pm2.6\hphantom{0}}$ & 34.2$_{\pm2.6\hphantom{0}}$ & 4.1 & 44.0\\
% \textsc{InstructGPT} & N/A & 70.2$_{\pm0.4\hphantom{0}}$ & 70.1$_{\pm2.3\hphantom{0}}$ & 90.0$_{\pm3.5\hphantom{0}}$ & 63.0$_{\pm2.0\hphantom{0}}$ & 23.1 & 69.2\\
% \textsc{InstructGPT + CoT} & N/A & 67.8$_{\pm0.5\hphantom{0}}$ & 63.2$_{\pm2.1\hphantom{0}}$ & 88.5$_{\pm2.5\hphantom{0}}$ & 54.2$_{\pm2.8\hphantom{0}}$ & 17.2 & 61.8\\
\multicolumn{8}{c}{\cellcolor{light-gray} \emph{One-Shot on RULE (for reference)}} \\
\textsc{UnifiedQA-v2-Base} & 220M & 27.7$_{\pm2.7\hphantom{0}}$ & 36.5$_{\pm2.5\hphantom{0}}$ & 38.5$_{\pm4.0\hphantom{0}}$ & 35.8$_{\pm2.5\hphantom{0}}$ & 1.6 & 36.9\\
\textsc{UnifiedQA-v2-Large} & 770M & 28.3$_{\pm1.4\hphantom{0}}$ & 27.4$_{\pm1.2\hphantom{0}}$ & 27.4$_{\pm10.2}$ & 27.4$_{\pm3.0\hphantom{0}}$ & 1.0 & 27.7\\
\textsc{UnifiedQA-v2-3B} & 3B & 35.0$_{\pm1.4\hphantom{0}}$ & 30.1$_{\pm4.8\hphantom{0}}$ & 35.8$_{\pm10.7}$ & 28.2$_{\pm4.0\hphantom{0}}$ & 2.0 & 30.6\\
\textsc{UnifiedQA-v2-11B} & 11B & 42.6$_{\pm7.1\hphantom{0}}$ & 37.4$_{\pm11.7}$ & 47.5$_{\pm17.7}$ & 33.9$_{\pm9.7\hphantom{0}}$ & 3.1 & 38.0\\
\textsc{Flan-T5-XXL} & 11B & 60.6$_{\pm0.5\hphantom{0}}$ & 64.1$_{\pm3.6\hphantom{0}}$ & 85.8$_{\pm3.1\hphantom{0}}$ & 56.5$_{\pm3.9\hphantom{0}}$ & 14.0 & 63.2\\
\textsc{Flan-UL2} & 20B & 57.6$_{\pm0.4\hphantom{0}}$ & 66.0$_{\pm4.5\hphantom{0}}$ & 87.4$_{\pm3.8\hphantom{0}}$ & 58.5$_{\pm4.7\hphantom{0}}$ & 17.2 & 64.9\\
\textsc{LLaMA 7B} & 7B & 28.2$_{\pm2.4\hphantom{0}}$ & 32.9$_{\pm2.5\hphantom{0}}$ & 48.3$_{\pm8.1\hphantom{0}}$ & 27.4$_{\pm4.1\hphantom{0}}$ & 1.6 & 33.6\\
\textsc{LLaMA 13B} & 13B & 30.0$_{\pm3.2\hphantom{0}}$ & 32.9$_{\pm1.3\hphantom{0}}$ & 50.9$_{\pm4.3\hphantom{0}}$ & 26.6$_{\pm1.8\hphantom{0}}$ & 1.7 & 33.4\\
\textsc{LLaMA 33B} & 33B & 53.3$_{\pm3.2\hphantom{0}}$ & 48.4$_{\pm4.5\hphantom{0}}$ & 79.7$_{\pm3.4\hphantom{0}}$ & 37.3$_{\pm4.7\hphantom{0}}$ & 6.0 & 48.4\\
\textsc{LLaMA 65B} & 65B & 62.8$_{\pm3.4\hphantom{0}}$ & 52.8$_{\pm6.0\hphantom{0}}$ & 84.9$_{\pm4.4\hphantom{0}}$ & 41.5$_{\pm6.5\hphantom{0}}$ & 7.5 & 52.0\\
\textsc{Vicuna 7B} & 7B & 34.7$_{\pm1.8\hphantom{0}}$ & 41.2$_{\pm2.4\hphantom{0}}$ & 65.3$_{\pm6.4\hphantom{0}}$ & 32.6$_{\pm3.3\hphantom{0}}$ & 3.4 & 41.2\\
\textsc{Vicuna 13B} & 13B & 40.7$_{\pm2.3\hphantom{0}}$ & 41.7$_{\pm1.6\hphantom{0}}$ & 67.3$_{\pm4.8\hphantom{0}}$ & 32.7$_{\pm1.8\hphantom{0}}$ & 3.5 & 41.8\\
\textsc{InstructGPT} & N/A & 65.4$_{\pm1.9\hphantom{0}}$ & 66.5$_{\pm1.3\hphantom{0}}$ & 89.0$_{\pm0.8\hphantom{0}}$ & 58.5$_{\pm1.2\hphantom{0}}$ & 19.5 & 65.5\\
\textsc{InstructGPT + CoT} & N/A & 64.3$_{\pm2.5\hphantom{0}}$ & 64.3$_{\pm2.4\hphantom{0}}$ & 88.8$_{\pm1.3\hphantom{0}}$ & 55.7$_{\pm2.5\hphantom{0}}$ & 15.4 & 62.5\\
\textsc{LLaMA2 7B} & 7B & 32.8$_{\pm3.3\hphantom{0}}$ & 33.4$_{\pm1.7\hphantom{0}}$ & 56.1$_{\pm1.8\hphantom{0}}$ & 25.4$_{\pm1.9\hphantom{0}}$ & 1.4 & 34.0\\
\textsc{LLaMA2 13B} & 13B & 45.0$_{\pm5.5\hphantom{0}}$ & 40.9$_{\pm2.9\hphantom{0}}$ & 70.6$_{\pm4.4\hphantom{0}}$ & 30.3$_{\pm2.6\hphantom{0}}$ & 2.8 & 41.1\\
\textsc{LLaMA2 70B} & 70B & 76.0$_{\pm0.2\hphantom{0}}$ & 61.5$_{\pm2.8\hphantom{0}}$ & 90.4$_{\pm3.1\hphantom{0}}$ & 51.4$_{\pm2.7\hphantom{0}}$ & 19.2 & 61.1\\
\textsc{Mistral 7B} & 7B & 50.8$_{\pm1.1\hphantom{0}}$ & 53.8$_{\pm3.3\hphantom{0}}$ & 84.0$_{\pm3.7\hphantom{0}}$ & 43.2$_{\pm3.6\hphantom{0}}$ & 6.5 & 54.0\\
\multicolumn{8}{c}{\cellcolor{light-gray} \emph{Zero-Shot}} \\
\textsc{UnifiedQA-v2-Base} & 220M & 30.4\hphantom{$_{\pm0.00}$} & 42.2$_{\pm1.0\hphantom{0}}$ & 48.5$_{\pm2.9\hphantom{0}}$ & 39.9$_{\pm1.3\hphantom{0}}$ & 2.7 & 41.7\\
\textsc{UnifiedQA-v2-Large} & 770M & 41.4\hphantom{$_{\pm0.00}$} & 42.9$_{\pm1.8\hphantom{0}}$ & 55.0$_{\pm4.9\hphantom{0}}$ & 38.5$_{\pm1.3\hphantom{0}}$ & 3.3 & 41.9\\
% \textsc{UnifiedQA-v2-3B} & 3B & 45.5\hphantom{$_{\pm0.00}$} & 47.9$_{\pm2.1\hphantom{0}}$ & 71.6$_{\pm2.9\hphantom{0}}$ & 39.4$_{\pm2.2\hphantom{0}}$ & 5.7 & 47.8\\
% \textsc{UnifiedQA-v2-11B} & 11B & 55.2\hphantom{$_{\pm0.00}$} & 57.3$_{\pm2.7\hphantom{0}}$ & 74.8$_{\pm5.2\hphantom{0}}$ & 51.1$_{\pm2.7\hphantom{0}}$ & 9.7 & 56.5\\
% \textsc{Flan-T5-XXL} & 11B & 60.0\hphantom{$_{\pm0.00}$} & 64.3$_{\pm4.0\hphantom{0}}$ & 86.2$_{\pm5.4\hphantom{0}}$ & 56.5$_{\pm3.3\hphantom{0}}$ & 14.7 & 63.4\\
% \textsc{Flan-UL2} & 20B & 56.2\hphantom{$_{\pm0.00}$} & 65.7$_{\pm5.2\hphantom{0}}$ & 84.5$_{\pm4.4\hphantom{0}}$ & 59.1$_{\pm5.1\hphantom{0}}$ & 14.7 & 64.2\\
\textsc{LLaMA 7B} & 7B & 27.7\hphantom{$_{\pm0.00}$} & 27.4$_{\pm4.1\hphantom{0}}$ & 38.2$_{\pm4.2\hphantom{0}}$ & 23.6$_{\pm4.6\hphantom{0}}$ & 0.8 & 27.1\\
\textsc{LLaMA 13B} & 13B & 31.7\hphantom{$_{\pm0.00}$} & 36.0$_{\pm3.3\hphantom{0}}$ & 59.3$_{\pm2.1\hphantom{0}}$ & 27.8$_{\pm3.7\hphantom{0}}$ & 1.4 & 36.7\\
% \textsc{LLaMA 33B} & 33B & 54.5\hphantom{$_{\pm0.00}$} & 50.9$_{\pm4.4\hphantom{0}}$ & 81.3$_{\pm2.8\hphantom{0}}$ & 40.1$_{\pm4.5\hphantom{0}}$ & 6.8 & 50.9\\
% \textsc{LLaMA 65B} & 65B & 52.1\hphantom{$_{\pm0.00}$} & 47.9$_{\pm3.1\hphantom{0}}$ & 78.9$_{\pm1.8\hphantom{0}}$ & 36.9$_{\pm3.5\hphantom{0}}$ & 5.4 & 47.2\\
\textsc{Vicuna 7B} & 7B & 36.7\hphantom{$_{\pm0.00}$} & 40.5$_{\pm2.3\hphantom{0}}$ & 73.1$_{\pm3.2\hphantom{0}}$ & 28.9$_{\pm2.9\hphantom{0}}$ & 2.4 & 40.1\\
\textsc{LLaMA2 7B} & 7B & 32.3\hphantom{$_{\pm0.00}$} & 35.9$_{\pm3.9\hphantom{0}}$ & 66.4$_{\pm4.7\hphantom{0}}$ & 25.1$_{\pm3.4\hphantom{0}}$ & 2.5 & 36.3\\
% \textsc{Vicuna 13B} & 13B & 44.2\hphantom{$_{\pm0.00}$} & 49.5$_{\pm2.7\hphantom{0}}$ & 77.1$_{\pm1.7\hphantom{0}}$ & 39.7$_{\pm2.7\hphantom{0}}$ & 6.2 & 49.4\\
% \textsc{InstructGPT} & N/A & 64.1\hphantom{$_{\pm0.00}$} & 62.8$_{\pm2.2\hphantom{0}}$ & 89.9$_{\pm2.0\hphantom{0}}$ & 53.2$_{\pm2.1\hphantom{0}}$ & 15.5 & 61.8\\
% \textsc{InstructGPT + CoT} & N/A & 63.8\hphantom{$_{\pm0.00}$} & 62.3$_{\pm1.0\hphantom{0}}$ & 89.6$_{\pm1.5\hphantom{0}}$ & 52.6$_{\pm1.5\hphantom{0}}$ & 14.2 & 61.2\\
% \midrule \textsc{Human} & N/A & 86.5\hphantom{$_{\pm0.00}$} & 77.7\hphantom{$_{\pm0.00}$} & 87.9\hphantom{$_{\pm0.00}$} & 74.1\hphantom{$_{\pm0.00}$} & 53.0 & 81.5 \\
\midrule \textsc{Human} & - & 91.5\hphantom{$_{\pm0.00}$} & 82.6\hphantom{$_{\pm0.00}$} & 93.0\hphantom{$_{\pm0.00}$} & 78.9\hphantom{$_{\pm0.00}$} & 52.9 & 81.5 \\   % majority vote-based
\bottomrule
\end{tabular}
\caption{
    Complementary results of the model performance on our dataset including the models in the one-shot setting and omitting those in the five-shot and zero-shot settings.
    % One-shot model performance on our RULE dataset, as well as the fully supervised and human performance for reference.
}
\label{tab:one-shot-result}
\end{table*}

% \section{Results on the ``None of the Above Choices'' Questions}
% \label{app:none-of-the-above}

\begin{figure}[!t]
    \centering
    \includegraphics[width=\linewidth]{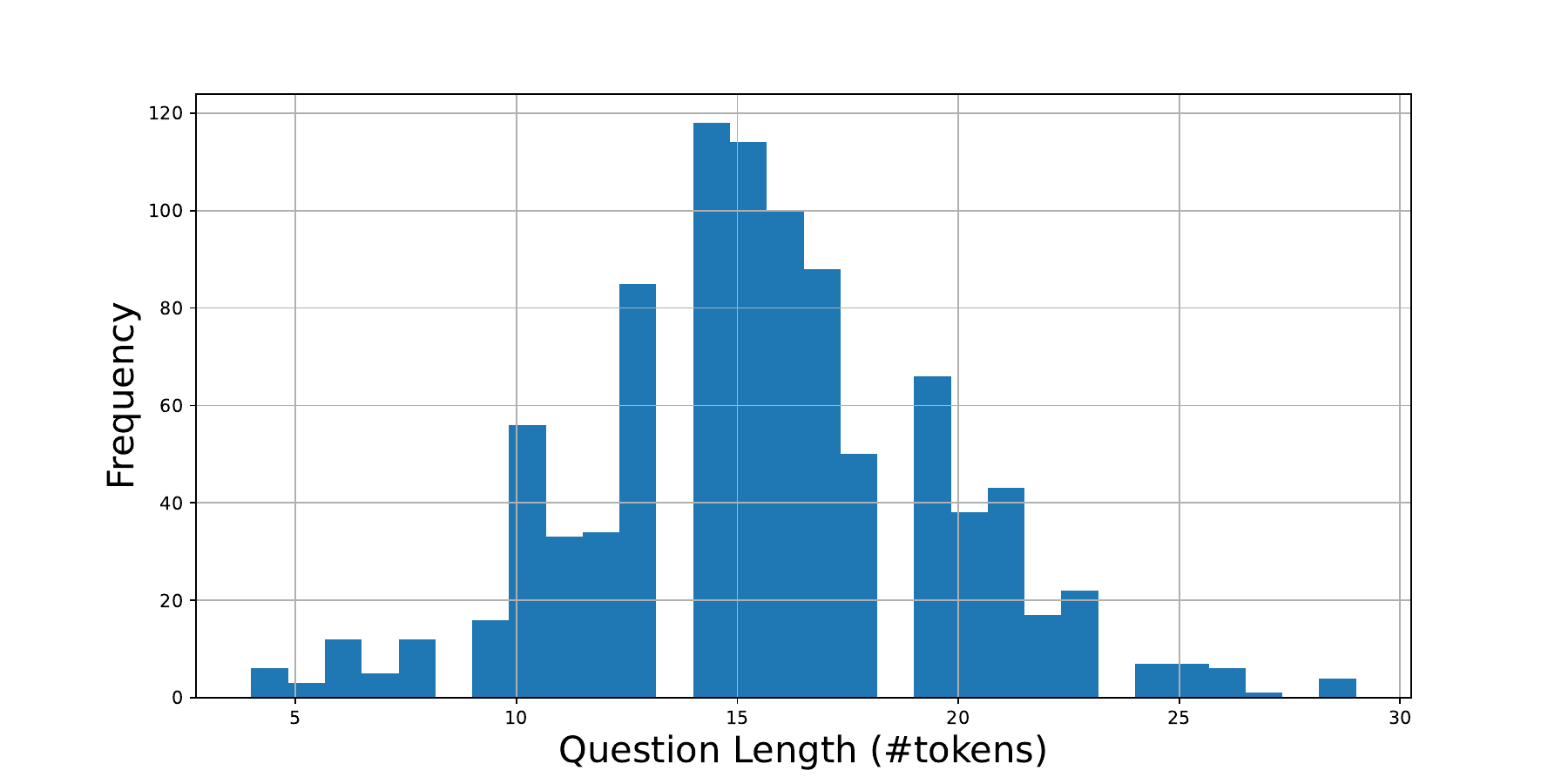}
    \caption{Distribution of the question length (\#tokens) of the main questions.}
    \label{fig:qlen-dist-mainq}
\end{figure}

\begin{figure}[!t]
    \centering
    \includegraphics[width=\linewidth]{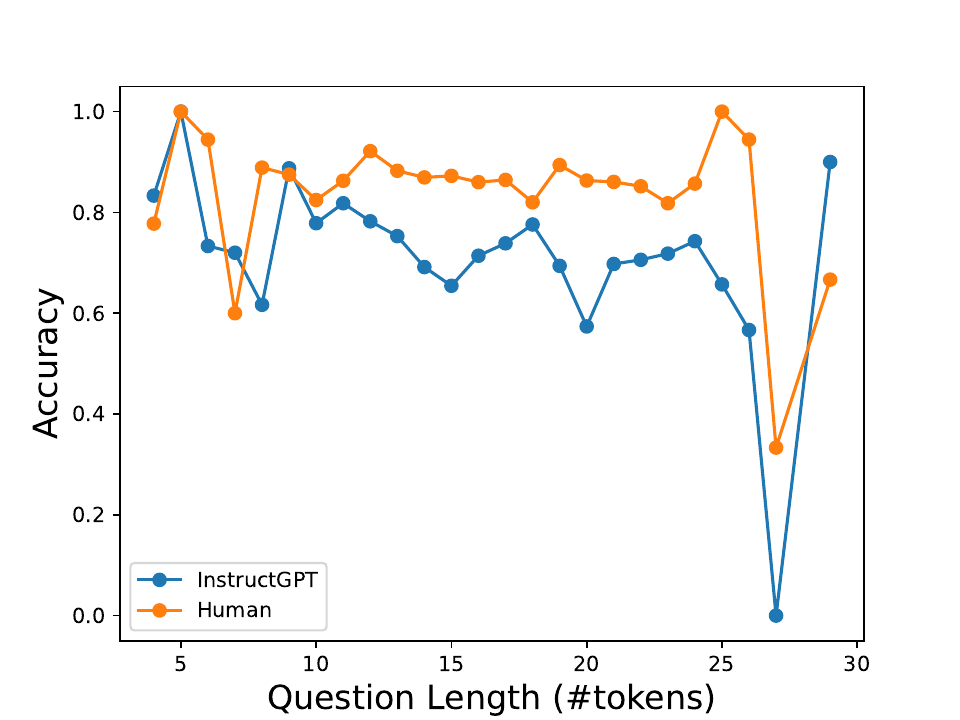}
    \caption{Comparison between the model and human accuracy and question length for the main questions.}
    \label{fig:comp-qlen-mainq}
\end{figure}%

\begin{figure}[!t]
    \centering
    \includegraphics[width=\linewidth]{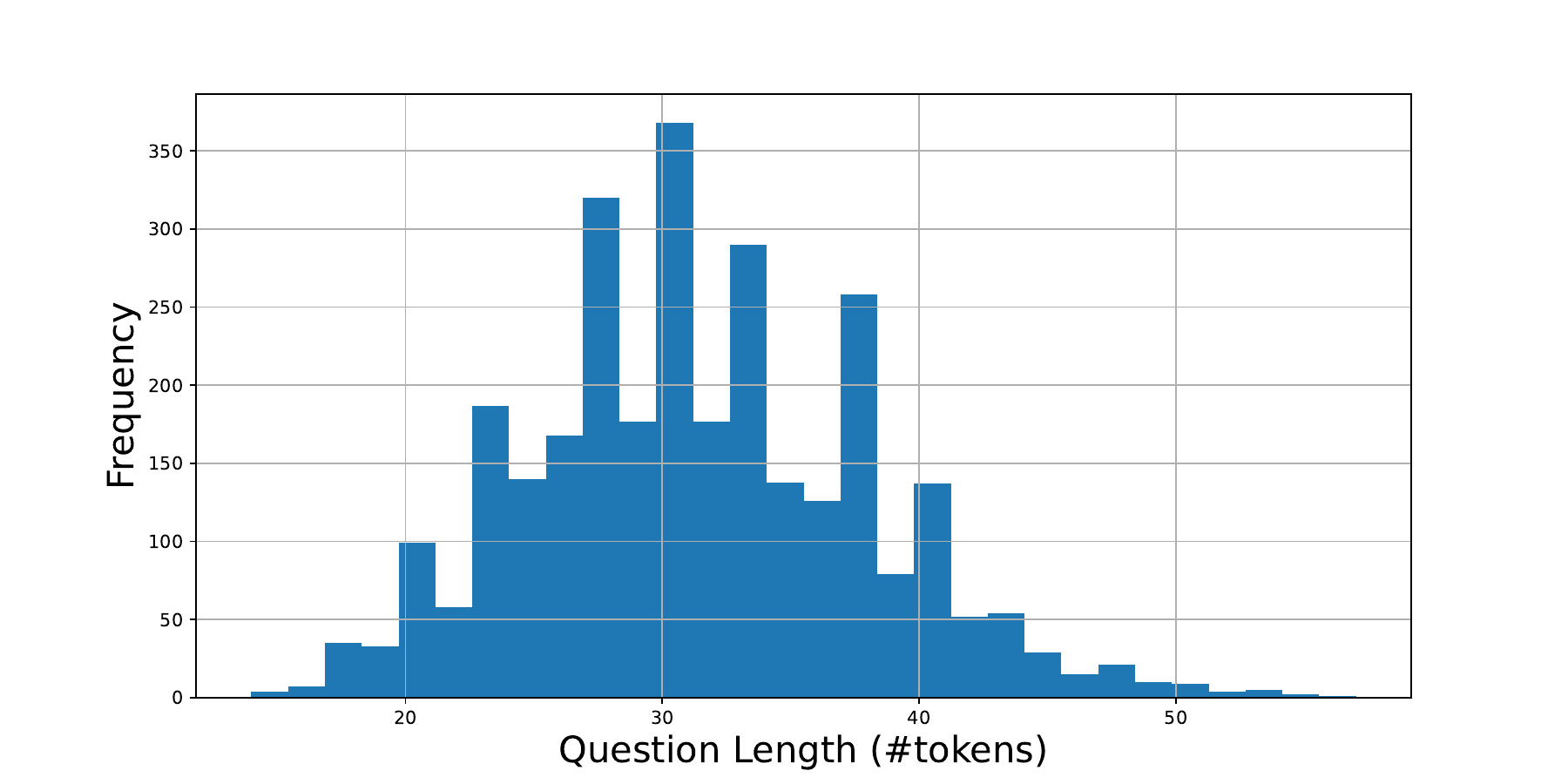}
    \caption{Distribution of the question length (\#tokens) of the subquestions.}
    \label{fig:qlen-dist-subq}
\end{figure}%

\begin{figure}[!t]
    \centering
    \includegraphics[width=\linewidth]{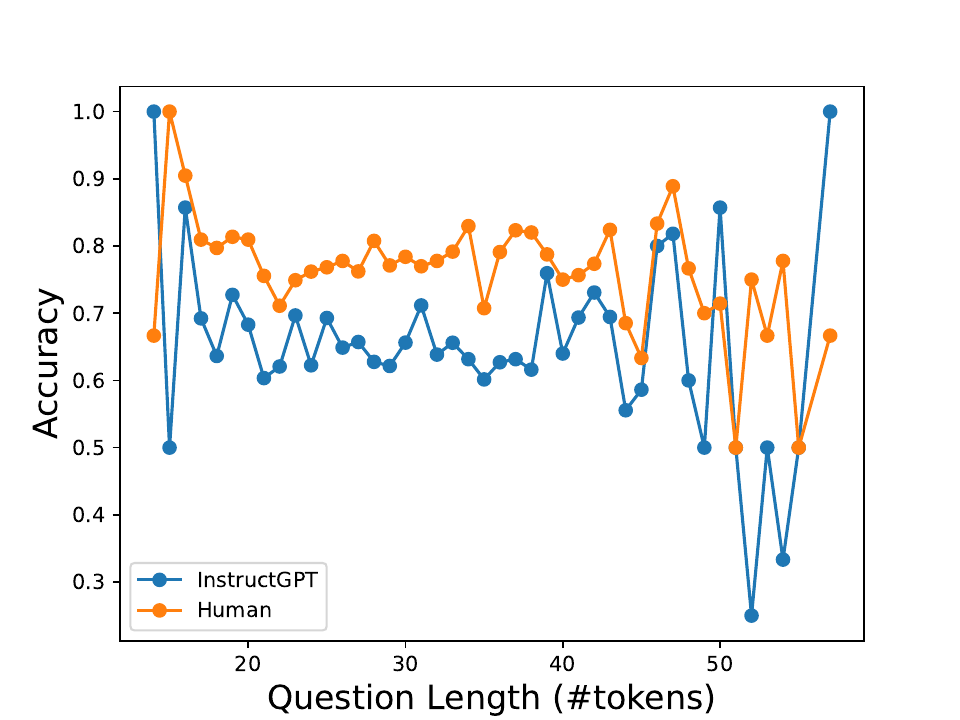}
    \caption{Comparison between the model and human accuracy and question length for the subquestions.}
    \label{fig:comp-qlen-subq}
\end{figure}%

%%%%%%%%%%%%%%%%%%%%%%%%%%%%%%%%

\begin{figure}[!t]
    \centering
    \includegraphics[width=\linewidth]{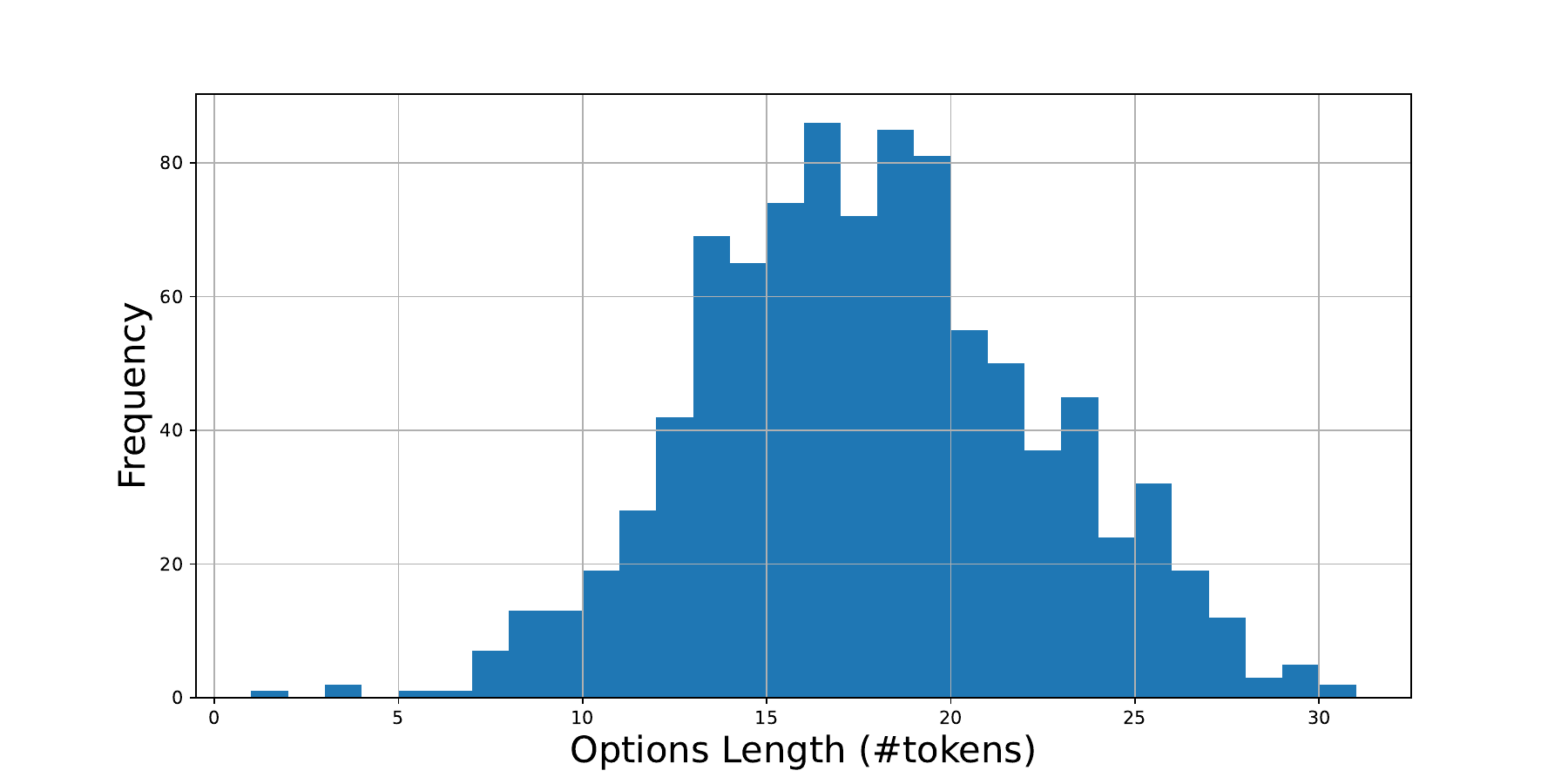}
    \caption{Distribution of the option length (\#tokens) of the main questions.}
    \label{fig:olen-dist-mainq}
\end{figure}

\begin{figure}[!t]
    \centering
    \includegraphics[width=\linewidth]{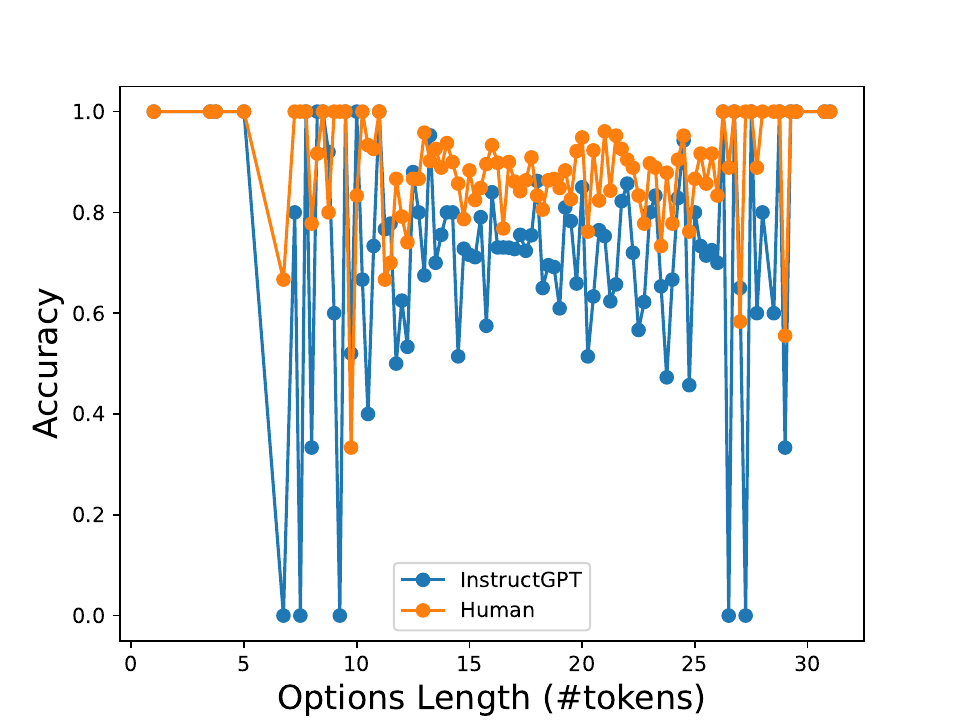}
    \caption{Comparison between the model and human accuracy and option length for the main questions.}
    \label{fig:comp-olen-mainq}
\end{figure}%

\begin{figure}[!t]
    \centering
    \includegraphics[width=\linewidth]{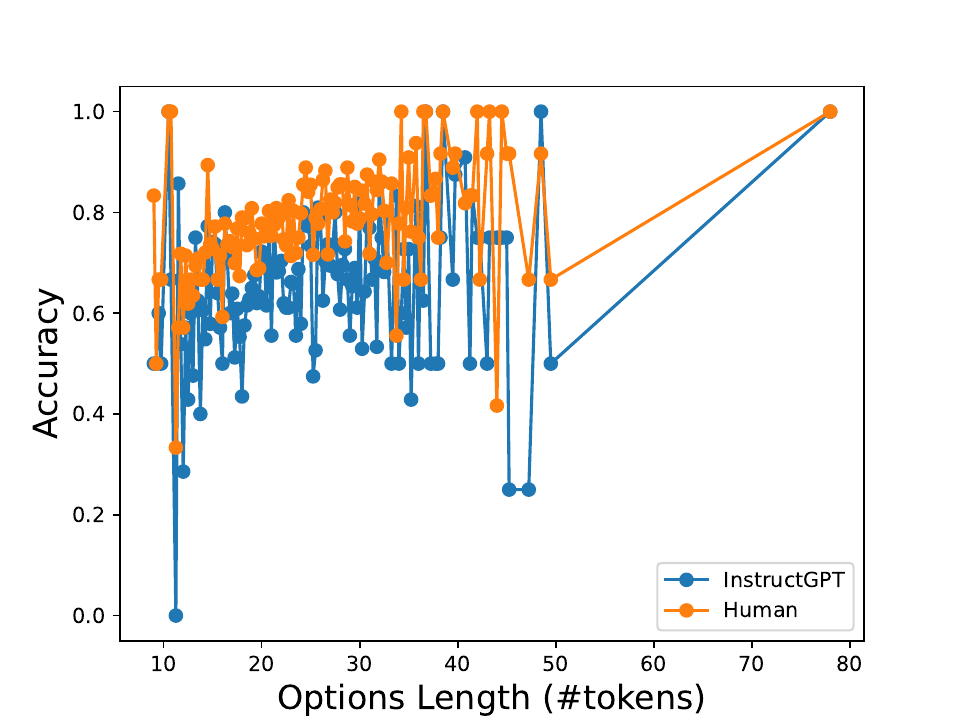}
    \caption{Comparison between the model and human accuracy and option length for the subquestions.}
    \label{fig:comp-olen-subq}
\end{figure}%

\begin{figure}[!t]
    \centering
    \includegraphics[width=\linewidth]{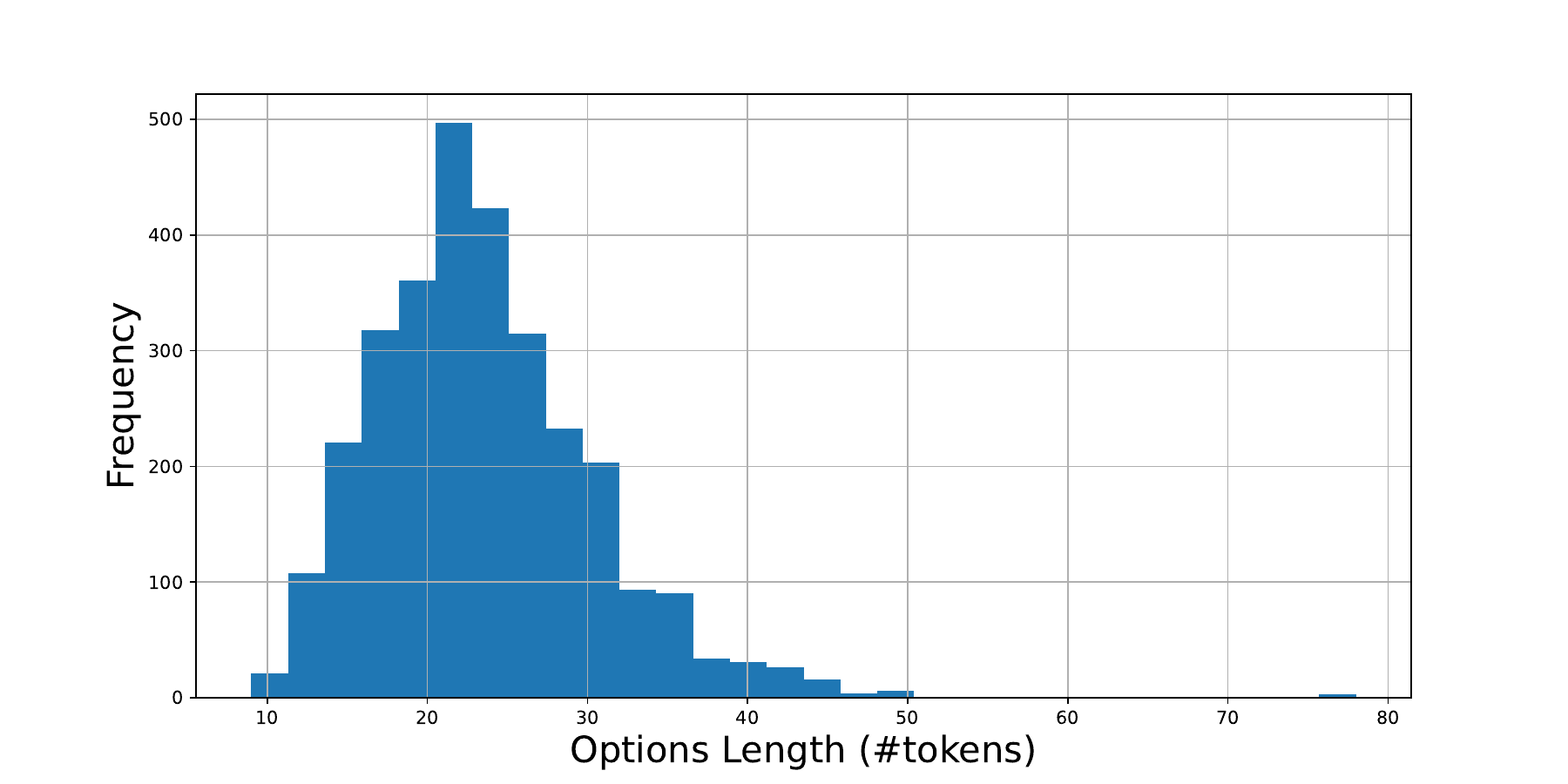}
    \caption{Distribution of the option length (\#tokens) of the subquestions.}
    \label{fig:olen-dist-subq}
\end{figure}%

% Table~\ref{tab:none-of-the-above} shows the InstructGPT performance on the subquestions that have ``None of the above choices'' as the correct answer across the five batches.
% The low accuracy on those subquestions is decomposed into 40.9\% accuracy if the prompt includes the ``None'' option as the correct answer and 13.7\% accuracy otherwise.
% This result demonstrates that using exemplars helps to answer those questions to some extent but not significantly.

\begin{table}[!htb]
    \centering
    \begin{tabular}{lcc} \toprule
    Instruction & Corr. Opt. & Incorr. Opt. \\ \midrule
    Yes & 89.7 (784) & 31.5 (2,218) \\
    No & 88.7 (781) & 28.1 (2,213) \\
    \bottomrule
    \end{tabular}
    \caption{
        Result of the rationale alignment task with and without the task instruction.
    }
    \label{tab:rationale-alignment}
\end{table}

\section{Complementary Few-Shot and Zero-Shot Results}
\label{app:one-shot-result}

In Table~\ref{tab:one-shot-result}, we report the complementary results of few-shot settings, including the models on the one-shot setting. %and models such as LLaMA 7B and 13B across all settings.
We also report the results of LLaMA \cite[7B to 65B;][]{touvron2023llama} for reference.

\section{Main Results of the Subquestions for the Correctly-Answered Main Questions}
\label{app:correct-only-result}

Table~\ref{tab:correct-only-result} shows the main results of the model performance on the subquestions for the main questions that are correctly answered by the model.
Overall, we observe similar trends to the main results with the standard SubQ accuracy.
Interestingly, the models' SubQ accuracies do not significantly improve even when we focus only on the correctly-answered main questions.

\begin{table*}[!h]
\centering \small
\begin{tabular}{lcrrrrrr} \toprule
\textbf{Model} & \textbf{\# Param} & \makecell[c]{\textbf{MainQ}\\\textbf{Acc.}} & \makecell[c]{\textbf{SubQ}\\\textbf{Acc.}} & \makecell[c]{\textbf{Selective}\\\textbf{SubQ Acc.}} & \makecell[c]{\textbf{Eliminative}\\\textbf{SubQ Acc.}} & \textbf{Consist.} &  \makecell[r]{\textbf{MainQ-wise}\\\textbf{SubQ Acc.}} \\ \midrule
\multicolumn{8}{c}{\cellcolor{light-gray} \emph{Fully Finetuned on ReClor}} \\

\textsc{DeBERTa-v3-Large} & 304M & 66.0\hphantom{$_{\pm0.00}$} & 33.1\hphantom{$_{\pm0.00}$} & 60.4\hphantom{$_{\pm0.00}$} & 22.5\hphantom{$_{\pm0.00}$} & 3.7 & 32.2\\
\textsc{UnifiedQA-v2-Base} & 220M & 40.5\hphantom{$_{\pm0.00}$} & 25.8\hphantom{$_{\pm0.00}$} & 19.7\hphantom{$_{\pm0.00}$} & 26.7\hphantom{$_{\pm0.00}$} & 1.8 & 25.0\\
\textsc{UnifiedQA-v2-Large} & 770M & 57.7\hphantom{$_{\pm0.00}$} & 25.0\hphantom{$_{\pm0.00}$} & 17.3\hphantom{$_{\pm0.00}$} & 27.7\hphantom{$_{\pm0.00}$} & 2.4 & 24.6\\
\textsc{UnifiedQA-v2-3B} & 3B & 66.8\hphantom{$_{\pm0.00}$} & 25.3\hphantom{$_{\pm0.00}$} & 19.9\hphantom{$_{\pm0.00}$} & 25.7\hphantom{$_{\pm0.00}$} & 2.1 & 24.1\\
\multicolumn{8}{c}{\cellcolor{light-gray} \emph{Five-Shot on ReClor}} \\
\textsc{Flan-UL2} & 20B & 58.5$_{\pm0.3\hphantom{0}}$ & 66.3$_{\pm6.3\hphantom{0}}$ & 89.6$_{\pm4.6\hphantom{0}}$ & 57.2$_{\pm5.1\hphantom{0}}$ & 28.8 & 65.1\\
\textsc{LLaMA 7B} & 7B & 25.8$_{\pm1.6\hphantom{0}}$ & 28.5$_{\pm7.5\hphantom{0}}$ & 34.3$_{\pm12.2}$ & 24.2$_{\pm5.7\hphantom{0}}$ & 3.4 & 27.2\\
\textsc{LLaMA 13B} & 13B & 38.7$_{\pm2.7\hphantom{0}}$ & 37.0$_{\pm4.0\hphantom{0}}$ & 65.0$_{\pm6.7\hphantom{0}}$ & 25.9$_{\pm3.8\hphantom{0}}$ & 7.4 & 37.2\\
\textsc{LLaMA 33B} & 33B & 58.5$_{\pm1.2\hphantom{0}}$ & 47.7$_{\pm4.2\hphantom{0}}$ & 77.5$_{\pm6.3\hphantom{0}}$ & 38.3$_{\pm5.8\hphantom{0}}$ & 10.6 & 47.3\\
\textsc{LLaMA 65B} & 65B & 69.1$_{\pm0.9\hphantom{0}}$ & 54.4$_{\pm1.7\hphantom{0}}$ & 85.5$_{\pm2.8\hphantom{0}}$ & 46.9$_{\pm5.6\hphantom{0}}$ & 16.0 & 54.1\\
\textsc{Vicuna 7B} & 7B & 33.4$_{\pm2.6\hphantom{0}}$ & 40.0$_{\pm5.1\hphantom{0}}$ & 62.7$_{\pm6.9\hphantom{0}}$ & 30.0$_{\pm3.0\hphantom{0}}$ & 8.5 & 39.7\\
\textsc{Vicuna 13B} & 13B & 46.2$_{\pm0.7\hphantom{0}}$ & 48.5$_{\pm6.1\hphantom{0}}$ & 76.0$_{\pm8.8\hphantom{0}}$ & 42.2$_{\pm5.4\hphantom{0}}$ & 12.1 & 48.1\\
\textsc{InstructGPT} & N/A & 71.8$_{\pm1.0\hphantom{0}}$ & 64.1$_{\pm3.5\hphantom{0}}$ & 89.2$_{\pm3.8\hphantom{0}}$ & 60.7$_{\pm6.1\hphantom{0}}$ & 25.1 & 62.8\\
\textsc{InstructGPT + CoT} & N/A & 67.8$_{\pm0.5\hphantom{0}}$ & 62.9$_{\pm3.1\hphantom{0}}$ & 89.5$_{\pm4.2\hphantom{0}}$ & 55.8$_{\pm3.6\hphantom{0}}$ & 24.9 & 61.5\\
\textsc{LLaMA2 13B} & 13B & 48.5$_{\pm2.5\hphantom{0}}$ & 45.8$_{\pm4.2\hphantom{0}}$ & 79.1$_{\pm4.0\hphantom{0}}$ & 33.1$_{\pm5.2\hphantom{0}}$ & 11.1 & 45.8\\
\textsc{LLaMA2 70B} & 70B & 80.3$_{\pm0.4\hphantom{0}}$ & 59.9$_{\pm2.5\hphantom{0}}$ & 90.7$_{\pm1.9\hphantom{0}}$ & 50.5$_{\pm7.2\hphantom{0}}$ & 21.8 & 59.4\\
\textsc{Mistral 7B} & 7B & 59.9$_{\pm0.9\hphantom{0}}$ & 54.7$_{\pm4.4\hphantom{0}}$ & 84.5$_{\pm2.8\hphantom{0}}$ & 46.8$_{\pm2.8\hphantom{0}}$ & 15.0 & 53.7\\
\multicolumn{8}{c}{\cellcolor{light-gray} \emph{Five-Shot on RULE (for reference)}} \\
\textsc{Flan-UL2} & 20B & 57.9$_{\pm0.2\hphantom{0}}$ & 67.2$_{\pm6.2\hphantom{0}}$ & 89.4$_{\pm4.4\hphantom{0}}$ & 57.1$_{\pm3.6\hphantom{0}}$ & 30.9 & 66.2\\
\textsc{LLaMA 7B} & 7B & 29.1$_{\pm2.3\hphantom{0}}$ & 34.6$_{\pm3.2\hphantom{0}}$ & 66.2$_{\pm8.0\hphantom{0}}$ & 23.7$_{\pm1.3\hphantom{0}}$ & 5.4 & 34.9\\
\textsc{LLaMA 13B} & 13B & 36.8$_{\pm3.5\hphantom{0}}$ & 35.1$_{\pm3.2\hphantom{0}}$ & 67.4$_{\pm5.3\hphantom{0}}$ & 24.4$_{\pm2.6\hphantom{0}}$ & 6.6 & 35.0\\
\textsc{LLaMA 33B} & 33B & 53.6$_{\pm0.4\hphantom{0}}$ & 48.5$_{\pm6.0\hphantom{0}}$ & 78.5$_{\pm5.9\hphantom{0}}$ & 36.8$_{\pm5.4\hphantom{0}}$ & 10.6 & 47.6\\
\textsc{LLaMA 65B} & 65B & 66.2$_{\pm0.7\hphantom{0}}$ & 57.2$_{\pm5.1\hphantom{0}}$ & 86.4$_{\pm3.1\hphantom{0}}$ & 53.7$_{\pm10.3}$ & 18.3 & 56.0\\
\textsc{Vicuna 7B} & 7B & 35.0$_{\pm1.1\hphantom{0}}$ & 40.3$_{\pm3.1\hphantom{0}}$ & 61.4$_{\pm8.8\hphantom{0}}$ & 32.4$_{\pm4.8\hphantom{0}}$ & 9.5 & 40.8\\
\textsc{Vicuna 13B} & 13B & 43.9$_{\pm1.3\hphantom{0}}$ & 43.9$_{\pm2.6\hphantom{0}}$ & 73.0$_{\pm3.1\hphantom{0}}$ & 34.2$_{\pm3.7\hphantom{0}}$ & 9.4 & 43.4\\
\textsc{InstructGPT} & N/A & 70.2$_{\pm0.4\hphantom{0}}$ & 70.2$_{\pm3.5\hphantom{0}}$ & 91.0$_{\pm4.6\hphantom{0}}$ & 64.3$_{\pm2.4\hphantom{0}}$ & 32.7 & 69.3\\
\textsc{InstructGPT + CoT} & N/A & 67.8$_{\pm0.5\hphantom{0}}$ & 62.9$_{\pm3.1\hphantom{0}}$ & 89.5$_{\pm4.2\hphantom{0}}$ & 55.8$_{\pm3.6\hphantom{0}}$ & 24.9 & 61.5\\
\textsc{LLaMA2 13B} & 13B & 47.7$_{\pm3.0\hphantom{0}}$ & 46.9$_{\pm4.7\hphantom{0}}$ & 79.4$_{\pm6.3\hphantom{0}}$ & 33.6$_{\pm4.6\hphantom{0}}$ & 11.0 & 47.0\\
\textsc{LLaMA2 70B} & 70B & 78.9$_{\pm0.6\hphantom{0}}$ & 63.7$_{\pm4.2\hphantom{0}}$ & 90.8$_{\pm2.5\hphantom{0}}$ & 55.2$_{\pm9.2\hphantom{0}}$ & 26.5 & 63.0\\
\textsc{Mistral 7B} & 7B & 58.2$_{\pm1.6\hphantom{0}}$ & 56.2$_{\pm5.1\hphantom{0}}$ & 88.0$_{\pm3.6\hphantom{0}}$ & 49.4$_{\pm7.9\hphantom{0}}$ & 16.3 & 55.6\\
\multicolumn{8}{c}{\cellcolor{light-gray} \emph{Zero-Shot}} \\
\textsc{UnifiedQA-v2-Base} & 220M & 30.4\hphantom{$_{\pm0.00}$} & 43.0$_{\pm2.9\hphantom{0}}$ & 51.0$_{\pm7.5\hphantom{0}}$ & 39.8$_{\pm1.7\hphantom{0}}$ & 8.7 & 43.4\\
\textsc{UnifiedQA-v2-Large} & 770M & 41.4\hphantom{$_{\pm0.00}$} & 43.9$_{\pm4.1\hphantom{0}}$ & 59.9$_{\pm7.6\hphantom{0}}$ & 39.0$_{\pm2.7\hphantom{0}}$ & 7.9 & 42.5\\
\textsc{UnifiedQA-v2-3B} & 3B & 45.5\hphantom{$_{\pm0.00}$} & 49.3$_{\pm2.4\hphantom{0}}$ & 75.4$_{\pm4.1\hphantom{0}}$ & 38.9$_{\pm3.1\hphantom{0}}$ & 12.6 & 49.5\\
\textsc{UnifiedQA-v2-11B} & 11B & 55.2\hphantom{$_{\pm0.00}$} & 56.8$_{\pm2.9\hphantom{0}}$ & 77.6$_{\pm7.1\hphantom{0}}$ & 53.0$_{\pm3.8\hphantom{0}}$ & 17.5 & 55.9\\
\textsc{Flan-T5-XXL} & 11B & 60.0\hphantom{$_{\pm0.00}$} & 63.2$_{\pm3.4\hphantom{0}}$ & 87.3$_{\pm4.8\hphantom{0}}$ & 59.2$_{\pm4.6\hphantom{0}}$ & 24.6 & 62.7\\
\textsc{Flan-UL2} & 20B & 56.2\hphantom{$_{\pm0.00}$} & 65.3$_{\pm5.9\hphantom{0}}$ & 86.0$_{\pm4.8\hphantom{0}}$ & 60.5$_{\pm4.7\hphantom{0}}$ & 26.2 & 63.7\\
\textsc{LLaMA 7B} & 7B & 27.7\hphantom{$_{\pm0.00}$} & 27.0$_{\pm4.5\hphantom{0}}$ & 37.5$_{\pm2.8\hphantom{0}}$ & 23.6$_{\pm3.9\hphantom{0}}$ & 3.1 & 26.9\\
\textsc{LLaMA 13B} & 13B & 31.7\hphantom{$_{\pm0.00}$} & 35.5$_{\pm3.0\hphantom{0}}$ & 55.9$_{\pm4.0\hphantom{0}}$ & 27.5$_{\pm3.8\hphantom{0}}$ & 4.3 & 35.9\\
\textsc{LLaMA 33B} & 33B & 54.5\hphantom{$_{\pm0.00}$} & 50.2$_{\pm4.6\hphantom{0}}$ & 81.8$_{\pm4.3\hphantom{0}}$ & 41.4$_{\pm5.5\hphantom{0}}$ & 12.5 & 50.2\\
\textsc{LLaMA 65B} & 65B & 52.1\hphantom{$_{\pm0.00}$} & 47.5$_{\pm2.2\hphantom{0}}$ & 80.1$_{\pm2.5\hphantom{0}}$ & 38.0$_{\pm5.4\hphantom{0}}$ & 10.4 & 46.6\\
\textsc{Vicuna 7B} & 7B & 36.7\hphantom{$_{\pm0.00}$} & 40.8$_{\pm2.8\hphantom{0}}$ & 77.5$_{\pm2.8\hphantom{0}}$ & 29.4$_{\pm2.8\hphantom{0}}$ & 6.6 & 40.3\\
\textsc{Vicuna 13B} & 13B & 44.2\hphantom{$_{\pm0.00}$} & 48.9$_{\pm4.8\hphantom{0}}$ & 76.9$_{\pm3.4\hphantom{0}}$ & 40.4$_{\pm1.8\hphantom{0}}$ & 13.9 & 48.2\\
\textsc{InstructGPT} & N/A & 64.1\hphantom{$_{\pm0.00}$} & 61.9$_{\pm2.9\hphantom{0}}$ & 89.8$_{\pm2.9\hphantom{0}}$ & 55.8$_{\pm3.6\hphantom{0}}$ & 24.2 & 60.8\\
\textsc{InstructGPT + CoT} & N/A & 63.8\hphantom{$_{\pm0.00}$} & 60.9$_{\pm1.4\hphantom{0}}$ & 89.1$_{\pm2.4\hphantom{0}}$ & 55.9$_{\pm3.2\hphantom{0}}$ & 22.3 & 59.7\\
\textsc{LLaMA2 13B} & 13B & 43.8\hphantom{$_{\pm0.00}$} & 45.5$_{\pm3.7\hphantom{0}}$ & 77.6$_{\pm5.3\hphantom{0}}$ & 33.3$_{\pm3.0\hphantom{0}}$ & 10.7 & 45.4\\
\textsc{LLaMA2 70B} & 70B & 70.8\hphantom{$_{\pm0.00}$} & 57.2$_{\pm4.0\hphantom{0}}$ & 88.2$_{\pm3.2\hphantom{0}}$ & 49.9$_{\pm5.4\hphantom{0}}$ & 19.9 & 56.3\\
\textsc{Mistral 7B} & 7B & 54.0\hphantom{$_{\pm0.00}$} & 54.7$_{\pm3.4\hphantom{0}}$ & 85.1$_{\pm5.5\hphantom{0}}$ & 47.1$_{\pm4.3\hphantom{0}}$ & 15.9 & 53.3\\
\midrule \textsc{Human} & - & 91.5\hphantom{$_{\pm0.00}$} & 82.9\hphantom{$_{\pm0.00}$} & 92.8\hphantom{$_{\pm0.00}$} & 79.3\hphantom{$_{\pm0.00}$} & 57.8 & 81.6 \\   % majority vote-based
\bottomrule
\end{tabular}
\caption{
    Main results of the model performance on our dataset focusing on the subquestions for the main questions that are correctly answered by the model.
    % One-shot model performance on our RULE dataset, as well as the fully supervised and human performance for reference.
}
\label{tab:correct-only-result}
\end{table*}

\section{Relationship between Question and Option Length and Model Performance}
\label{app:length-and-performance}

In Figures~\ref{fig:qlen-dist-mainq} to \ref{fig:comp-olen-subq}, we plot the distribution of the questions and options length and the model performance according to those lengths.
% From these figures, we do not observe any strong trends.

\section{Rationale Alignment Task}
\label{app:rationale-alignment}
% \begin{table}[t]
%     \centering
%     \begin{tabular}{lccc} \toprule
%         Batch & Acc. & \# \textit{None} in shot & \textit{None} Acc. \\ \midrule
%         \#1 & 70.1 & 0 & 10.3 \\
%         \#2 & 69.7 & 0 & 25.9 \\
%         \#3 & 72.9 & 0 & 0.0 \\
%         \#4 & 71.3 & 1 & 43.8 \\
%         \#5 & 66.0 & 1 & 40.6 \\ \midrule
%        Avg. & 70.1 & 0.4 & 32.0 \\ 
%         \bottomrule
%     \end{tabular}
%     \caption{
%         Accuracy of the subquestions that have ``None of the above choices'' as the correct answer (\textit{None Acc}), compared to that of all subquestions (\textit{Acc}).
%         \textit{None in shot} indicates how many ``None'' examples are included in the few-shot prompt for each test split.
%     }
%     \label{tab:none-of-the-above}
% \end{table}

\begin{figure*}[!t]
    \centering
    \includegraphics[width=\textwidth]{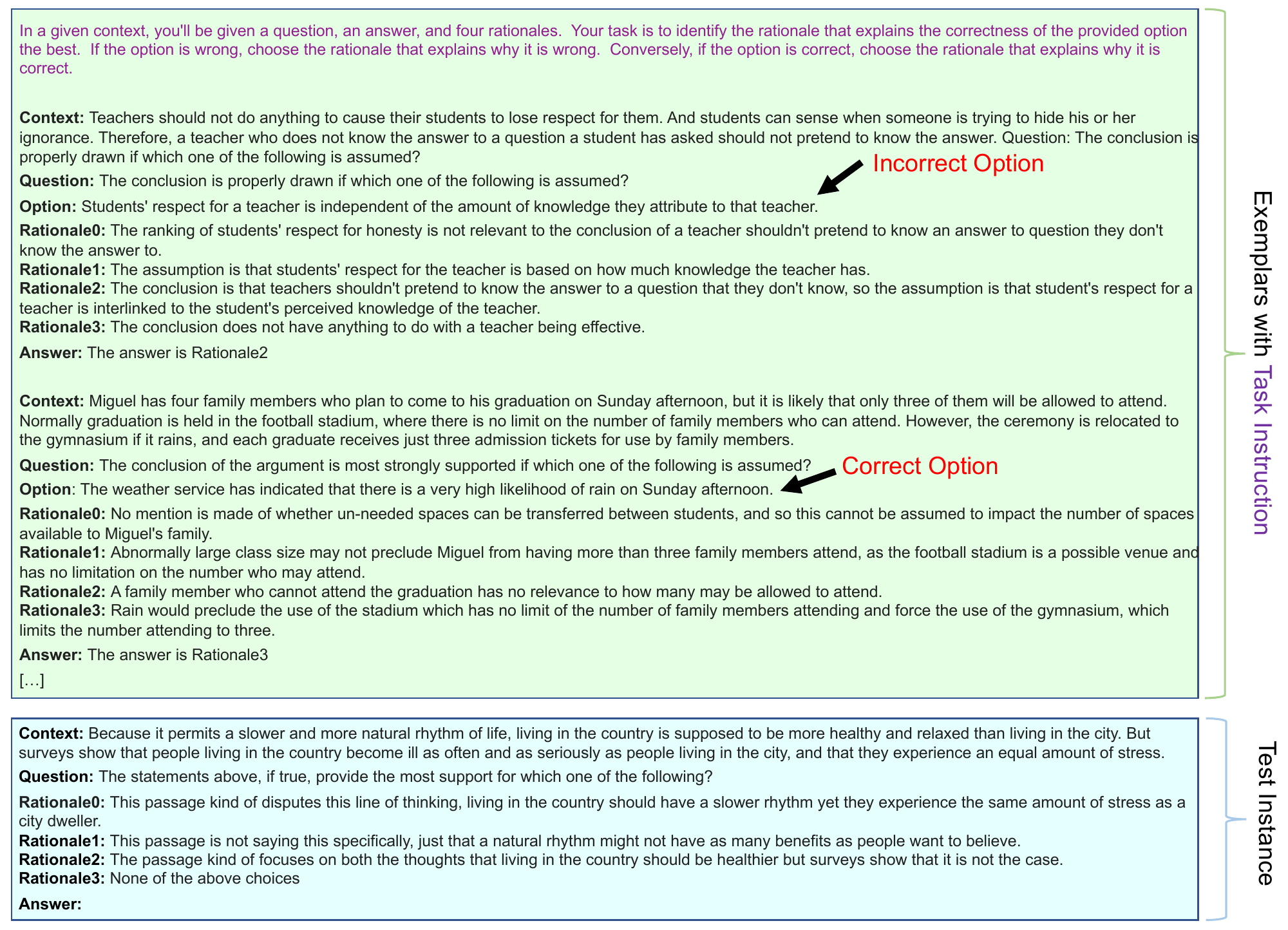}
    \caption{Example of the prompt used for rationale alignment task.}
    \label{fig:prompt-rationale-alignment}
\end{figure*}%

In the rationale alignment task, we test InstructGPT in the five-shot setting.
Similar to the main experiment, we report the average results from five prompts.
Each prompt is composed of five exemplars, with two exemplars presenting the correct option and three exemplars presenting the incorrect option.
Figure~\ref{fig:prompt-rationale-alignment} shows an example of our prompt.
% We also include the task instruction, resulting in the following prompt:

% \begin{itemize}
%     \item[] In a given context, you'll be given a question, an answer, and four rationales. Your task is to identify the rationale that explains the correctness of the provided option the best. If the option is wrong, choose the rationale that explains why it is wrong. Conversely, if the option is correct, choose the rationale that explains why it is correct. \\
%     (\textit{five exemplars}) \\
%     Context: ..., Question: ..., Option: ..., Rationale1: ..., ..., Rationale4: ..., Answer:
% \end{itemize}

The results with and without the task instruction are shown in Table~\ref{tab:rationale-alignment}.
The performance gap between the correct and incorrect options implies that such advanced models may simply infer the correct answer without properly discriminating against incorrect options. 
Such a situation raises two issues: (1) the inability to reason logically like a human, and (2) the limitations of ability measurement using distractors. 
The first issue suggests that the model may not be able to make a clear distinction between what is correct and what is incorrect. %, as humans do. 
% The second issue is that the function expected of distractors in multiple-choice questions, namely, distinguishing between test takers with insufficient knowledge and those with sufficient knowledge \cite{Girel-et-al-distractor}, may not be fulfilled. 
The second issue is that the alternatives in the multiple-choice QA task are generally expected to distinguish between test takers with and without sufficient knowledge \cite{Girel-et-al-distractor}, but such an expectation may not be met in our dataset.

\begin{table}[!h]
    \centering \small
    \begin{tabular}{lcccc} \toprule
    Setting & MainQ & SubQ & \makecell[c]{Selective \\ SubQ} & \makecell[c]{Eliminat.\\ SubQ} \\ \midrule
    0-shot & 41.0$_{-23.0}$ & 58.8$_{-4.2}$ & 86.4$_{-3.7}$ & 53.4$_{-4.3}$ \\
    5-shot & 42.5$_{-29.7}$ & 70.5$_{+0.0}$ & 86.9$_{-3.1}$ & 64.7$_{+1.7}$ \\
    \bottomrule
    \end{tabular}
    \caption{
        Context-ablated accuracy.
        The subscript values indicate the accuracy gap against the full-input setting.
        % Accuracy with the context ablated.
        % The subscript values indicate the performance difference against the full-input setting.
    }
    \label{tab:context-ablation}
\end{table}

\section{Reasoning Type Annotation}
\label{app:reasoning-type}

\begin{table*}[!h]
\centering
\small
\begin{tabular}{p{0.09\textwidth}p{0.25\textwidth}p{0.16\textwidth}p{0.15\textwidth}p{0.05\textwidth}p{0.16\textwidth}} \toprule
\textbf{Reasoning Type}     & \textbf{Passage} & \textbf{Question} & 
\textbf{Option} & \textbf{Correct} & \textbf{Rationale} \\ 
\toprule
\emph{Direct \quad Contextual}       
& 
Trisha: Today' s family is declining in its ability to carry out [...]. There must be a return to the traditional values of commitment and responsibility. Jerod: We ought to leave what is good enough alone. Contemporary families may be less stable than traditionally, but most people do not find that to be bad. [...].
&   
Trisha and Jerod disagree over whether the institution of the family is
& 
no longer traditional.
&
FALSE
&
Both Trisha and Jerod agree that families are no longer traditional, this is not what the argument is about.
 \\ 
\midrule 
\emph{Direct \qquad External}    
&  
A just government never restricts the right of its citizens to act upon their desires except when their acting upon their desires is a direct threat to the health or property of other of its citizens.
&
Which one of the following judgments most closely conforms to the principle cited above? 
&
A just government would not censor writings of Shakespeare, but it could censor magazines and movies that criticize the government.
&
FALSE
&
A just government would not censor magazines and movies that criticize the government because these things do not threaten the health or property of its citizens.
\\ 
\midrule
\emph{Indirect \quad Contextual} 
&
Doctor: The practice of using this therapy to treat the illness cannot be adequately supported by the claim that any therapy for treating the illness is more effective than no therapy at all. What must also be taken into account is that this therapy is expensive and complicated.
&
Which one of the following most accurately expresses the main point of the doctor's argument?
&
The therapy's possible effectiveness in treating the illness is not sufficient justification for using it.
& 
TRUE
&  
Therapy's other costs must be considered before enlisting the treatment as it is not cheap and not simple.
\\ 
\midrule
\emph{Indirect \quad External} 
&     
On average, corporations that encourage frequent social events in the workplace show higher profits than those that rarely do. This suggests that the EZ Corporation could boost its profits by having more staff parties during business hours.
&
Which one of the following, if true, most weakens the argument above?
&
Frequent social events in a corporate workplace leave employees with less time to perform their assigned duties than they would otherwise have.
&
FALSE
&
Frequent social events in a corporate workplace can re-energize employees, like a lunch break does.
\\
\bottomrule    
\end{tabular}
\caption{Examples of the reasoning types with a passage, a question, an option, the correctness of the option, and its human-written rationale.} 
\label{tab:reasoning_types}
\end{table*}
%\matt{in the first column, put the percentage of questions that exhibited this reasoning type in your sample of 100 questions, like ``Cause and effect (23\%)''}\ana{Say what is in blue.}
% Put an icon for each reasoning type
% Disraeli was born on 21 December 1804 at 6 King's Road, Bedford Row, Bloomsbury, London, the second child and eldest son of Isaac D'Israeli, a literary critic and historian, and Maria (Miriam), "née" Basevi

\begin{table*}
    \centering
    \begin{tabular}{p{0.10\textwidth}p{0.85\textwidth}}
        \toprule
        \multicolumn{2}{p{.95\linewidth}}{\textbf{Paragraph:} Trisha: Today' s family is declining in its ability to carry out its functions of child-rearing and providing 
        stability for adult life. There must be a return to the traditional values of commitment and responsibility. Jerod: We ought to leave what is good enough alone. Contemporary families may be less stable than traditionally, but most people do not find that to be bad. Contemporary criticisms of the family are overblown and destructive.} \\
        \midrule
        \multirow{6}{*}{MainQ}    & \textbf{Question:} Trisha and Jerod disagree over whether the institution of the family is \\
             & \textbf{Options:} \\ 
             & 1) valued by most people. \\ 
             & 2) changing over time. \\ 
             & \textbf{3}) adequate as it is. \\ 
             & 4) no longer traditional. \\
        \midrule
        \multirow{8}{1cm}{Selective \qquad SubQ}     & \textbf{Question:} What is the source of the disagreement between Trisha and Jerod regarding whether the institution of the family is adequate as it is? \\
             & \textbf{Options: } \\
             & 1) The argument does not mention value to the people. \\
             & \textbf{2}) Trisha is arguing that things were better with traditional families and Jerod is arguing that they are good now, the argument is about the quality of the relationship now. \\
             & 3) Both Trisha and Jerod agree that families are no longer traditional, this is not what the argument is about. \\
             & 4) None of the above choices. \\
        \midrule
        \multirow{8}{1cm}{Eliminative \qquad SubQ}     & \textbf{Question:} What evidence is there to suggest that Trisha and Jerod's disagreement over whether the institution of the family is no longer traditional is not valid? \\
             & \textbf{Options: } \\
             & \textbf{1}) Both Trisha and Jerod agree that families are no longer traditional, this is not what the argument is about. \\
             & 2) Trisha is arguing that things were better with traditional families and Jerod is arguing that they are good now, the argument is about the quality of the relationship now. \\
             & 3) The argument does not mention value to the people. \\
             & 4) None of the above choices. \\
        \bottomrule
    \end{tabular}
    \caption{Examples of the main questions and subquestions in our dataset.
    The options in bold indicate the correct answer.}
    \label{tab:mainq-subq-example}
\end{table*}

Table~\ref{tab:reasoning_types} shows examples of reasoning types we define in the annotation analysis.
See Table~\ref{tab:mainq-subq-example} for a full example that has the main question and two subquestions.

\section{ReClor Reasoning Types and Subquestion Accuracy}
\label{app:original-reasoning-type}

\begin{figure*}[t]
    \centering
    \includegraphics[width=\linewidth]{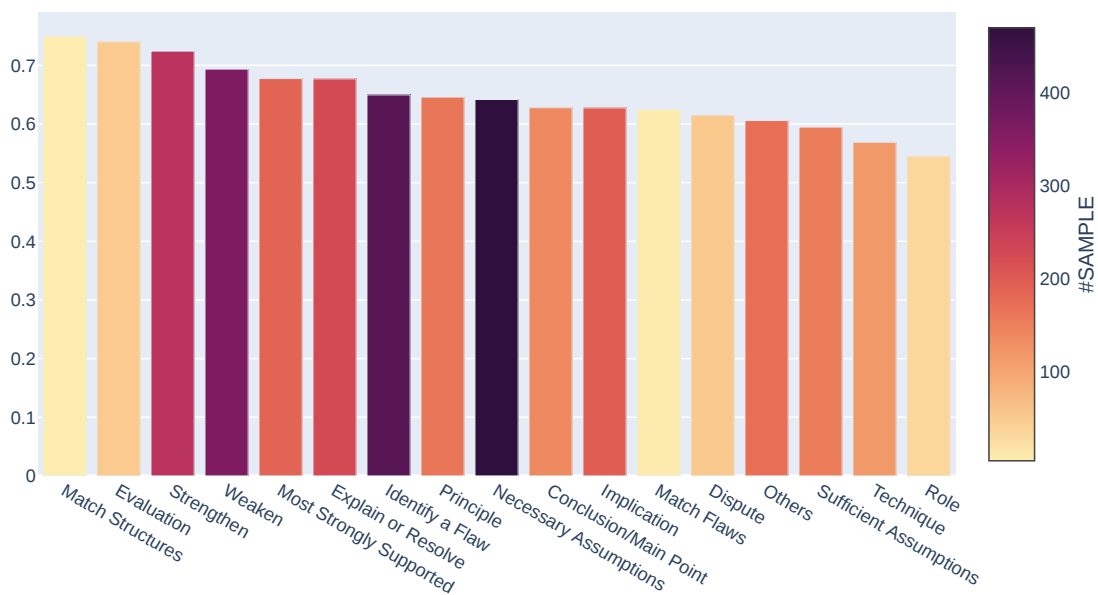}
    \caption{
        Accuracy of InstructGPT and reasoning types originally annotated in the ReClor dataset.
    }
    \label{fig:original-reasoning-type}
\end{figure*}

Figure~\ref{fig:original-reasoning-type} shows the relation between the subquestion accuracy and the reasoning types defined in the original ReClor dataset.
Although we do not observe significant performance differences, we see higher accuracy in Match Structures, Evaluation, Strengthen, and Weaken reasoning, and lower accuracy in Sufficient Assumptions, Technique, and Role reasoning.

\section{Context-Ablation Analysis}
\label{app:context-ablation}

% \paragraph{Does the Context Help in Answering Subquestions?}
% We also conduct input-ablation analysis.
% As the subquestions of a main question share the same set of answer options, we cannot ablate the subquestion texts (chance rate is 31.4\%).
We try to answer the question ``Does the context help in answering subquestions?'' in the context-ablation setting.
By removing the context, we analyze the model performance on the subquestions (and the main questions for reference) to see the dependency between question texts and answer options.
The results in Table~\ref{tab:context-ablation} show the performance reduction by approximately 4 points in the zero-shot setting and no reduction in the five-shot setting.
This result implies question texts depend on answer options to some extent, which potentially makes the subquestions difficult for the models, given the first analysis in this section.

\section{Similarity of Rationale with MainQ Option}
\label{app:sematic-similarity-rationale}
In our process to validate specificity, even if a rationale has the same meaning as the MainQ's option, we can not exclude it.
This implies that some rationales might have the similar meaning as the option and not serve as a valid rationale.
To examine this potential issue, we sample 50 random questions from both the selective SubQ and the eliminative SubQ.
We then count how many of these rationales are semantically similar to the MainQ's option.
We found three such instances in the selective SubQ and one in the eliminative SubQ, which are shown in Table~\ref{tab:similar-option-rationale}.

\begin{table*}
    \renewcommand{\arraystretch}{1.0}
    \centering
    \begin{tabular}{>{\centering\arraybackslash}p{0.15\textwidth}p{0.35\textwidth}p{0.40\textwidth}}
        \toprule
        \textbf{Question Type} & \textbf{Option} & \textbf{Rationale} \\
        \midrule
        \multirow{8}{*}{Selective} & Delays in the communication of discoveries will have a chilling effect on scientific research. & Delays in communicating discoveries would limit the time other scientists have to investigate and contribute. \\
             & Kimmy is a highly compensated and extremely popular television and movie actress. & All the information in the passage indicates that Kimmy is affluent and renowned. \\ 
             & Before new therapeutic agents reach the marketplace, they do not benefit patients. & The passage states that new therapies aid patients only after they are introduced to the marketplace. \\ 
        \midrule[1pt]
        \multirow{3}{*}{Eliminative} & The speed of eye orientation correlates with intelligence, not overall health. & The speed at which one can orient one's eye to a stimulus has been closely associated with overall health. \\
        \bottomrule[1.5pt]
    \end{tabular}
    \caption{Rationales that are semantically similar to the MainQ's option in Selective and Eliminative SubQs.}
    \label{tab:similar-option-rationale}
\end{table*}

% Figure~\ref{fig:prompt-rationale-alignment} shows an example of the prompt used in our analysis of the rationale alignment task.

%\section{Rationale Writing Instructions and Examples}

\end{document}